%% file: p.tex
\pdfoutput=1

\documentclass[sigconf,screen]{acmart}

\copyrightyear{2022}
\acmYear{2022}
\setcopyright{acmcopyright}\acmConference[CCS '22]{Proceedings of the 2022 ACM SIGSAC Conference on Computer and Communications Security}{November 7--11, 2022}{Los Angeles, CA, USA}
\acmBooktitle{Proceedings of the 2022 ACM SIGSAC Conference on Computer and Communications Security (CCS '22), November 7--11, 2022, Los Angeles, CA, USA}
\acmPrice{15.00}
\acmDOI{10.1145/3548606.3560558}
\acmISBN{978-1-4503-9450-5/22/11}

\copyrightyear{2022}
\acmYear{2022}
\setcopyright{acmcopyright}\acmConference[CCS '22]{Proceedings of the 2022 ACM SIGSAC Conference on Computer and Communications Security}{November 7--11, 2022}{Los Angeles, CA, USA}
\acmBooktitle{Proceedings of the 2022 ACM SIGSAC Conference on Computer and Communications Security (CCS '22), November 7--11, 2022, Los Angeles, CA, USA}
\acmPrice{15.00}
\acmDOI{10.1145/3548606.3560558}
\acmISBN{978-1-4503-9450-5/22/11}

\ccsdesc[500]{Security and privacy~Software and application security}
\ccsdesc[500]{Computer systems organization~Embedded and cyber-physical systems}

\include{pkgs}

\newcommand{\sys}{\mbox{\textsc{DriveFuzz}}\xspace}
\newcommand{\avf}{\mbox{\textsc{AV-Fuzzer}}\xspace}
\newcommand{\ads}{\mbox{\textsc{ADS}}\xspace}
\newcommand{\adses}{\mbox{\textsc{ADS}es}\xspace}

\newcommand{\updated}[2]{\textcolor{red}{\xspace{#2}\xspace}}

\input{cmds}

\begin{document}
\input{hdr}
\input{abstract}
\input{keyword}




\maketitle

\input{intro}
\input{background}
\input{threat}
\input{design}
\input{impl}
\input{eval}

\input{relwk}
\input{discussion}
\input{conclusion}
\input{ack}
\input{appendix}

\onecolumn
\begin{multicols}{2}
\bibliographystyle{ACM-Reference-Format}
\bibliography{p,sslab,conf}
\end{multicols}

\end{document}

%% file: pkgs.tex

\makeatletter
\newcommand{\removelatexerror}{\let\@latex@error\@gobble}
\makeatother

\usepackage{amsmath,amsopn,amssymb,amsopn,amsthm}
\usepackage{subfigure}
\usepackage{endnotes,microtype,xspace,graphicx,fancyvrb,multirow}
\usepackage{booktabs}
\usepackage{array,underscore,relsize}
\usepackage[T1]{fontenc}

\usepackage{fancyhdr}
\usepackage{enumitem}
\usepackage[labelfont=bf,font=small,skip=5pt]{caption}
\usepackage{fp}
\usepackage{siunitx}

\usepackage{balance}

\usepackage{listings}

\usepackage{caption}

\usepackage[ruled,vlined,linesnumbered]{algorithm2e}
\SetAlCapNameFnt{\small}
\SetAlCapFnt{\small}
\SetAlgorithmName{Algorithm}{Algorithm}{}
\usepackage{csquotes}

\usepackage{soul}
\usepackage{graphicx}

\usepackage{fontawesome}
\usepackage{bbding, pifont}

\usepackage{etoolbox}
\makeatletter
\patchcmd\algocf@Vline{\vrule}{\vrule \kern-0.4pt}{}{}
\patchcmd\algocf@Vsline{\vrule}{\vrule \kern-0.4pt}{}{}
\makeatother

\usepackage{multicol}

%% file: cmds.tex
\let\URL\relax
\newcommand{\URL}{\url}
\newcommand{\cc}[1]{\mbox{\smaller[0.5]\texttt{#1}}}

\fvset{fontsize=\scriptsize,xleftmargin=8pt,numbers=left,numbersep=5pt}

\def\Snospace~{\S{}}

\newif\ifdraft\drafttrue
\newif\ifnotes\notestrue
\ifdraft\else\notesfalse\fi

\input{glyphtounicode}
\newcolumntype{R}[1]{>{\raggedleft\let\newline\\\arraybackslash\hspace{0pt}}p{#1}}

\newcommand{\squishlist}{
\begin{itemize}[noitemsep,nolistsep]
  \setlength{\itemsep}{-0pt}
}
\newcommand{\squishend}{
  \end{itemize}
}

\usepackage{tikz}
\newcommand*\WC[1]{%
\begin{tikzpicture}[baseline=(C.base)]
\node[draw,circle,inner sep=0.2pt](C) {#1};
\end{tikzpicture}}

\usepackage{xstring}
\newcommand{\PP}[1]{
\vspace{2px}
\noindent{\bf \IfEndWith{#1}{.}{#1}{#1.}}
}

\newcommand{\V}{\checkmark}

\newcommand{\fx}{{\footnotesize \ding{81}}} 
\newcommand{\fo}{\faCube}
\newcommand{\fp}{\faStreetView}
\newcommand{\fc}{\faCar}

\newcommand{\etal}{\textit{et al}.\xspace}
\newcommand{\ie}{\textit{i}.\textit{e}.}
\newcommand{\eg}{\textit{e}.\textit{g}.}

\newcommand{\allbugs}{34\xspace} 
\newcommand{\newbugs}{33\xspace} 
\newcommand{\ackbugs}{10\xspace}  

\newcommand{\adsnewbugs}{30\xspace} 

\newcommand{\anewbugs}{17\xspace}     
\newcommand{\aoldbugse}{one\xspace}   

\newcommand{\ballbugs}{13\xspace}     

\newcommand{\callbugse}{three\xspace} 
\newcommand{\cnewbugs}{3\xspace}      
\newcommand{\cnewbugse}{three\xspace} 

\newcommand{\cackbugse}{two\xspace}   

\newcommand{\boxbeg}{
\vspace{2px}
\noindent\begin{tabular}{|l|}\hline
\begin{minipage}{3.2in}
\vspace{2px}
\noindent
}

\newcommand{\boxend}{
\vspace{2px}
\end{minipage}\\ \hline
\end{tabular}
\vspace{-10pt}
}



%% file: hdr.tex
\title{
    %
    %
    %
    %
    \sys: Discovering Autonomous Driving Bugs
    through Driving Quality-Guided Fuzzing
}

\author{Seulbae Kim}
\affiliation{
    \institution{Georgia Institute of Technology}
    \city{Atlanta}
    \state{Georgia}
    \country{USA}
}
\email{seulbae@gatech.edu}

\author{Major Liu}
\affiliation{
    \institution{University of Texas at Dallas}
    \city{Richardson}
    \state{Texas}
    \country{USA}
}
\email{major.liu@utdallas.edu}

\author{Junghwan ``John'' Rhee}
\affiliation{
    \institution{University of Central Oklahoma}
    \city{Edmond}
    \state{Oklahoma}
    \country{USA}
}
\email{jhree2@uco.edu}

\author{Yuseok Jeon}
\affiliation{
    \institution{UNIST}
    \city{Ulsan}
    \country{Republic of Korea}
}
\email{ysjeon@unist.ac.kr}

\author{Yonghwi Kwon}
\affiliation{
    \institution{University of Virginia}
    \city{Charlottesville}
    \state{Virginia}
    \country{USA}
}
\email{yongkwon@virginia.edu}

\author{Chung Hwan Kim}
\affiliation{
    \institution{University of Texas at Dallas}
    \city{Richardson}
    \state{Texas}
    \country{USA}
}
\email{chungkim@utdallas.edu}

\renewcommand{\shortauthors}{Seulbae Kim et al.}

%% file: abstract.tex
\begin{abstract}

Autonomous driving has become real;
semi-autonomous driving vehicles in an affordable price range
are already on the streets, and
major automotive vendors
are actively developing full self-driving systems
to deploy them in this decade.
Before rolling the products out to the end-users, it is critical to test and ensure the safety of the autonomous driving systems, consisting of multiple layers intertwined in a complicated way.
However, while safety-critical bugs may exist in any layer and even across layers, relatively little attention has been given to testing the entire driving system across all the layers.
Prior work mainly focuses on white-box
testing of individual layers and preventing attacks on each layer.

In this paper, we aim at holistic
testing of autonomous driving systems that have
a whole stack of layers integrated in their entirety.
Instead of looking into the individual layers,
we focus on
the vehicle states
that the system continuously changes
in the driving environment.
This allows us to design \sys,
a new systematic fuzzing framework
that can uncover potential vulnerabilities
regardless of their locations.
\sys automatically generates and mutates driving scenarios
based on diverse factors leveraging a high-fidelity driving simulator.
We build novel driving test oracles based on the real-world traffic rules
to detect safety-critical misbehaviors,
and guide the fuzzer towards such misbehaviors
through driving quality metrics referring to
the physical states of the vehicle.

\sys has discovered \adsnewbugs new bugs
in various layers of two autonomous driving systems
(Autoware and CARLA Behavior Agent)
and \callbugse additional bugs in the CARLA simulator.
We further analyze the impact of these bugs and
how an adversary may exploit them as security vulnerabilities
to cause critical accidents in the real world.
%

\end{abstract}

%% file: keyword.tex
\keywords{Autonomous driving system; Fuzzing}

%% file: intro.tex
\section{Introduction}
\label{s:intro}

Autonomous driving technology
has recently achieved significant breakthroughs,
making self-driving vehicles closer to
practical usages~\cite{chen2015deepdriving, chan2017advancements}.
Modern vehicles in an affordable price range
are already being shipped
with semi-autonomous driving systems
on board.
Major automotive companies are developing
autonomous driving systems (\adses)
to deploy fully autonomous vehicles
that can reliably operate on public roads
within this decade~\cite{mckinsey2019future}.
%
%
However,
despite the notable successes in the autopilot technology,
reports on fatal accidents
caused by erroneous \adses are continuing~\cite{
tesla:accident1, tesla:accident2, tesla:accident3,
waymo:accident, uber:accident}.
Moreover,
recent work has found many unpatched bugs in open-source
\adses~\cite{garcia2020bugs}
and analyzed that comprehensive testing of an \ads
still remains challenging~\cite{koopman2016challenges}.

To ensure the safety of autonomous driving,
existing work has focused on
individual layers of an \ads.
Specifically,
the security research community
has been extensively focusing on
finding adversarial examples
on the perception layer~\cite{sun2020towards, jing122021too, shen2020drift,
nassi2020phantom, cao2019adversarial, song2018adversarial, boloor2020attacking, chernikova2019self},
assuming a threat model in which an attacker
attempts to confuse the machine learning model by
supplying a deceptive driving scene
(\eg, modifying a traffic sign) or
spoofing sensor data
(\eg, injecting falsified LiDAR points).
Some other works test the robustness of the machine learning model
using synthesized and transformed images of driving scenes~\cite{pei2017deepxplore, tian2018deeptest, zhang2018deeproad}.
There are also testing approaches for other layers (e.g., sensing~\cite{jokela2019testing, geiger2012we, broggi2013extensive} and planning~\cite{ohta2016pure, calo2020generating, ndss:2022:ziwen:planfuzz}). However, they still focus on individual layers.

Although these works substantially improve the security of the individual layers,
they are not designed to cover the attacks exploiting
vulnerabilities outside their scopes or specific layers;
for example,
attacks that target bugs irrelevant to the machine learning model
or bugs in the actuation layer.
In addition,
due to the multi-layer architecture of \adses
where different layers work together in a cascading manner,
a bug in one layer may not be detected if tested individually.
For instance,
a bug in the perception layer may not cause a visible
impact when tested alone, but may cause the planning layer
to misbehave.
Moreover, bugs in multiple layers may jointly contribute to
one misbehavior.
Such bugs can only be detected if all the integrated layers
are tested together as a whole.

In this paper,
we introduce a novel approach to enable comprehensive testing of
\adses to uncover critical bugs across all layers.
We aim to design a fully automated testing framework
that generates realistic test input scenarios
on the fly to holistically test \adses
based on the following two key insights:

\begin{itemize}[itemsep=0em,topsep=0em,leftmargin=*]
%
%
%
%
\item
With the recent advances in driving
simulators for autonomous vehicles~\cite{
Dosovitskiy17, airsim2017fsr, lgsvl2020itsc},
it has become feasible to
generate an unlimited number
of \emph{high-fidelity test input scenarios}
with various driving conditions,
including the map, vehicles, pedestrians, and weather conditions
that closely reflect those of the real environments
and offer a highly desirable testing environment
to stress \emph{all layers} of the tested system.
%
\item
Regardless of which layer they belong to,
the impact of bugs ultimately affects
\emph{the physical states of the vehicle}\footnote{
We will use \emph{vehicle states} to refer to
the physical states of the vehicle herein.}
(\eg, position and velocity)
negatively,
for example, causing a collision. 
Thus, we focus on detecting misbehaviors by monitoring
the vehicle states that the \ads continuously alters.
These states
can also be used as \emph{feedback} 
to find bugs more efficiently without relying on the information
specific to individual layers.
\end{itemize}

\smallskip

Based on these insights,
we propose \sys,
a feedback-guided fuzzing framework for
end-to-end testing of \adses
leveraging a driving simulator
(CARLA~\cite{Dosovitskiy17}).
\sys plugs a target \ads into the fuzzing framework
and tests the self-driving system stack as a whole
to facilitate the test coverage to span all layers.
It generates and mutates \emph{driving scenarios}
in which the \ads has to drive from one point to another,
and simulates them in a three-dimensional virtual environment
(similar to a racing video game),
where it has full control over both \emph{spatial and temporal
dimensions of the input spaces as well as multiple actors and entities}
including the roads, pedestrians, and vehicles.

During a test, \sys utilizes 
our new \emph{driving test oracles} 
derived from real-world traffic rules and regulations~\cite{national1972traffic},
and actively monitors the vehicle states
to detect any \emph{misbehavior} that violates the oracles.
We define misbehavior of \ads as
safety-critical and illegal traffic violations,
including collisions, traffic infractions, and immobility,
which have apparent symptoms that wreak havoc on the safety of humans.
If such illegal misbehaviors are not found,
\sys measures the \emph{driving quality score} of
the test input scenario by quantifying the factors that indicate reckless driving,
\eg, accelerating too hard.
The resulting score is then used as
feedback to generate the subsequent test scenario more efficiently
(\ie, towards causing more unsafe driving scenarios),
such that it will trigger corner case bugs more quickly
than completely random fuzzing
(as demonstrated by existing software fuzzers~\cite{manes2019art, zalewski2014american,
bohme2017directed}).

We evaluate \sys by testing
Autoware~\cite{kato2015open, kato2018autoware},
which is a full-fledged (Level 4~\cite{sae2014levels}) \ads
extensively used by car manufacturers and academic
institutions~\cite{autoware2020foundation},
and Behavior Agent, which is a native \ads
integrated into CARLA.
To date,
\sys discovered a total of \allbugs critical bugs;
\newbugs of which are new bugs, including
\anewbugs in Autoware, 
\ballbugs in Behavior Agent,
and \cnewbugs simulation bugs in CARLA.
We have reported all \allbugs bugs to the developers,
and \ackbugs have been confirmed and being patched,
and others are under review.
%
Our discovery of the simulation bugs shows that
\sys is capable of finding bugs in both single and multiple layers of the software stack,
including various components for self-driving and even the simulator itself.
%
%
%
We observe and demonstrate that the bugs we found are \emph{realistic and practical}
to exploit;
that is, attackers can exploit them by controlling the external inputs
in a seemingly legitimate way (\eg, moving nearby objects).
%

Our design is generic and portable to other
\adses (\eg, Baidu Apollo~\cite{baidu2019apollo})
and driving simulators~\cite{airsim2017fsr, lgsvl2020itsc}
as it sits on the interface between the \ads and simulator
(\eg, ROS~\cite{ros2009icra}).
Specifically, \sys does not require the source code and instrumentation nor domain knowledge of the target \ads
since it only controls the input to the
system (input driving scenario)
and monitors the physical output (vehicle states).
\updated{\sys treats \ads as a black box so that other \adses can be supported without significant effort and expertise.}{}
%

This paper makes the following contributions:

\begin{itemize}[itemsep=0em,topsep=0em,leftmargin=*]

\item
We propose
a practical automated testing approach
capable of fuzzing \adses end-to-end,
and revealing safety-critical misbehaviors based on
real-world driving test oracles.
%
\item We design a novel driving quality metric
to estimate the effectivness of test driving scenarios
in exploring the input space of \adses
based on vehicle states,
and leverage the metric to better guide the mutation engine
towards test scenarios that trigger safety-critical misbehaviors.
\item
We implement and evaluate the proposed framework
in a prototype called \sys
to demonstrate how feedback-driven fuzzing can be
applied to the domain of \adses.
We open-sourced \sys at \URL{https://gitlab.com/s3lab-code/public/drivefuzz}.
%
\item
In our evaluation, we find 
\newbugs new bugs, including
30 critical bugs in real \adses and
\callbugse bugs in a full-fledged driving simulator.
We show that these bugs can be readily exploited 
to critically impair the safety of \adses
by causing them to crash, cease to operate,
or violate safety-critical traffic laws.
%
\end{itemize}

%% file: background.tex
\section{Background}
\label{s:background}

\begin{figure}[t]
  \centering
  \includegraphics[width=1.0\columnwidth]{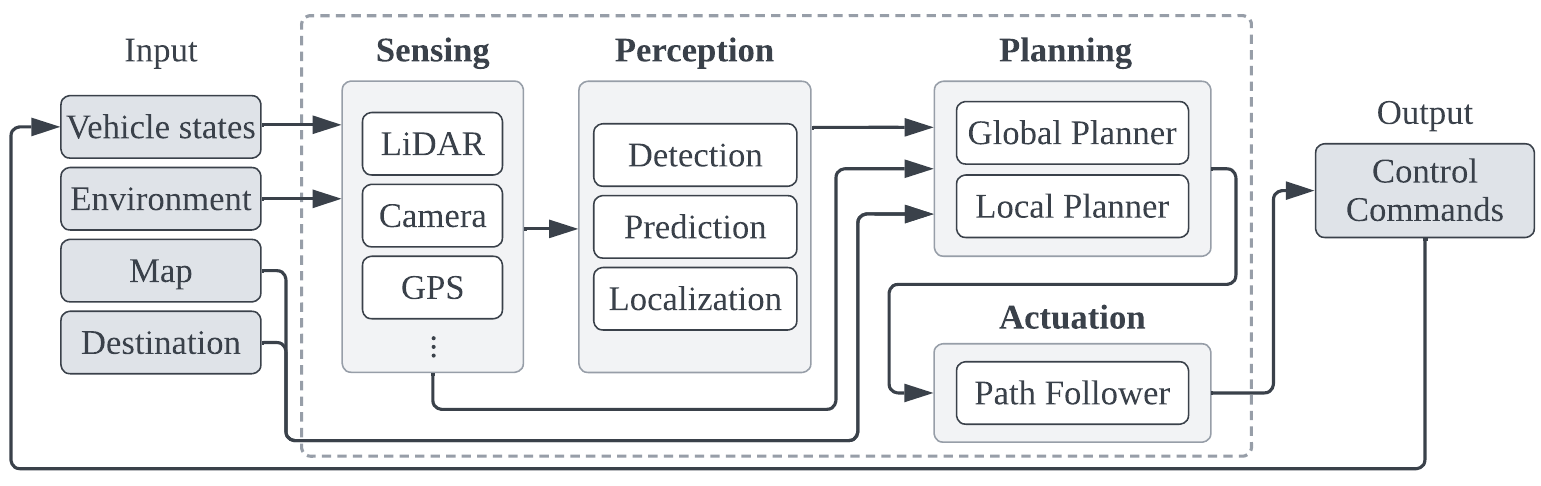}
  \caption{
    A general autonomous driving system (ADS)
    consisting of sensing, perception, planning, and actuation layers.
    %
    By taking the vehicle states and environment perceived by sensors, a 3D map, and a destination as inputs,
    \ads ultimately outputs control commands,
    \ie, steering, throttle and brake controls,
    that in turn update the vehicle states for the next iteration
    of the loop.
  }
  \label{f:ads}
  \vspace*{-1.5em}
\end{figure}

%
\autoref{f:ads} shows a general ADS, which
is the amalgamation of hardware and software layers
responsible for four core tasks:
\textit{sensing},
\textit{perception},
\textit{planning}, and
\textit{actuation}~\cite{
  ceccarelli2004fundamentals, jo2014development, jo2015development},
where each layer aims to substitute its counterpart of human drivers.
Within each layer,
multiple components carry out sub-tasks for autonomous driving.
These layers work together in a cascading manner to drive the vehicle,
\ie,
each layer takes input from the previous layers and
the produced output is consumed by the following layers.

\PP{Sensing}
Autonomous vehicles acquire
raw data of the surrounding environment
using various sensors,
typically including
a LiDAR (Light Detection and Ranging), cameras, a radar, a GPS device, and IMU (Inertial Measurement Unit) sensors
as components.
Any fault in sensors can feed faulty data to the system,
and the error can subsequently propagate to the other layers,
resulting in system-level faults in the worst case.

\PP{Perception}
Perception modules
fuse and interpret the captured sensor data
to comprehend
the current standing
and the environment around a vehicle.
For example,
identifying the traffic signals ahead or
predicting the motion of adjacent vehicles
by assessing their velocities
belongs to the tasks of the perception layer.
Many systems leverage
various computer vision and machine learning techniques
for such tasks.
A perception error can mislead the vehicle
to make faulty decisions,
\eg, estimating the distance to an obstacle
to be farther than the actual distance, ending up hitting it.

\PP{Planning}
With the perceived internal and external states,
the planning layer makes a routing plan 
for the given map and the destination.
Generally,
it first computes a global trajectory consisting of a sequence of waypoints
from the initial position
to the destination the user specifies.
And then,
traffic rules and perceived states (\eg, nearby obstacles)
are taken into account by a local planner,
which updates the trajectory at runtime
to safely drive to the destination.
Errors in this layer can cause
not only inefficient but also unsafe routing
that involves infeasible paths, \eg, crossing a river.

\PP{Actuation}
Given the generated trajectory to follow,
the actuation layer sets up
a concrete motion plan
consisting of
a steering wheel angle,
a target speed at waypoints,
and the amount of throttling or braking,
to seamlessly follow the trajectory.
These commands are sent
to the steering wheel, throttle, and brake controllers
to move an autonomous vehicle as planned.
When the commands move the vehicle in the driving environment,
the vehicle states are changed
and observed by the sensing layer in the following iteration of the loop.
Thus,
an error in the actuation layer can critically impair the vehicle's ability
to properly maneuver in a given situation
and may also affect other layers in the loop
by changing the vehicle states.

\begin{figure*}[t!]
  \centering
    \includegraphics[width=0.9\textwidth]{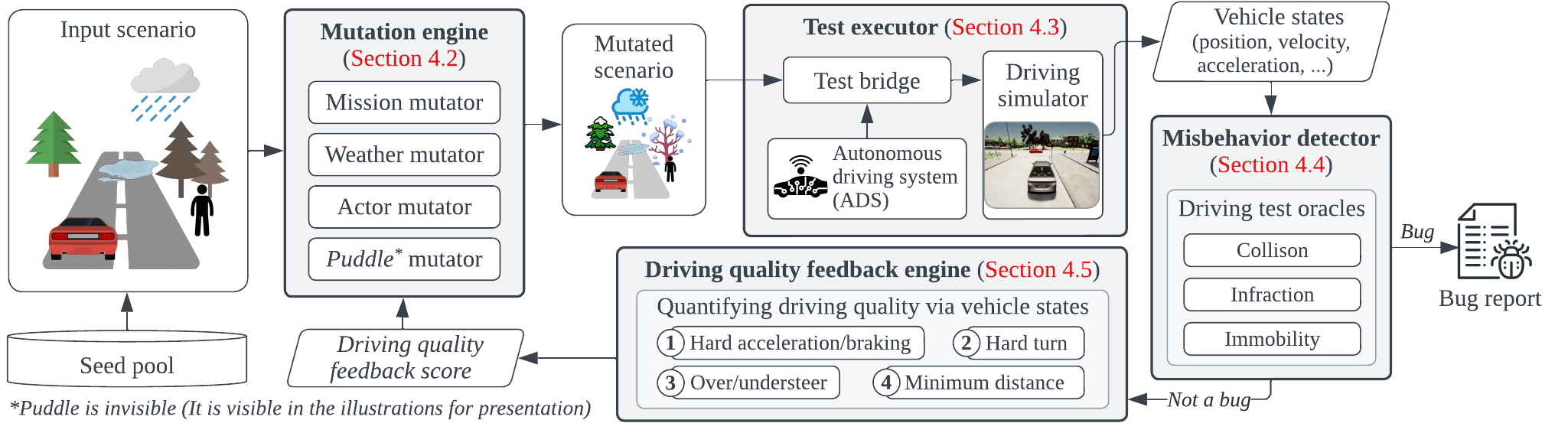}
    \vspace*{-0.5em}
    \caption{
        Overview of the architecture and workflow of \sys.
    }
  \label{f:arch}
  \vspace*{-1em}
\end{figure*}

%% file: threat.tex
\section{Threat Model}
\label{s:model}

\PP{Attack surface}
We assume an attacker who exploits bugs in
\emph{any layer} of an ADS.
Specifically,
the target attack surface 
is \emph{not limited} to a single layer
(\eg, an adversarial example on the perception layer
or injecting falsified data into the sensing layer).

We assume that
the attacker does not only exploit bugs that cause
an immediate failure of a single layer,
but also those that eventually \emph{manifest in other layers}.
For example,
a bug in the perception layer could make an unnoticeable error
in measuring the distance to an object,
and cause the planning
layer to malfunction when the erroneous distance is used 
as an input to find the trajectory
(bug \#15 in \autoref{s:eval}).
Similarly,
a bug in the actuation layer could produce an incorrect,
but seemingly legitimate command to move the vehicle
and cause the perception layer to fail
in the next iteration of the control loop
through the updated vehicle states 
(bug \#17 in \autoref{s:eval}).

We do not assume that the attacker takes control over the
\ads physically
(\eg, attach a device via an OBD-II port)
or remotely
(\eg, perform remote code execution)
to exploit a vulnerability.
%
Instead,
the attacker only has control over the \emph{external} inputs
such as nearby objects or locations
(\eg, moving a nearby vehicle)
with a goal to cause critical misbehavior of the autonomous vehicle
(\eg, a crash, lane invasion, traffic violations,
or becoming immobile).
These external inputs are
\emph{legitimate} and \emph{authentic} inputs to the ADS,
as opposed to
maliciously crafted inputs by adversarial attacks
(\eg, sensor spoofing)
or synthetically generated driving scenes.

\PP{Practical feasibility}
We argue the attacks in our threat model
are realistic and practical.
We further discuss the feasibility of
exploiting the bugs we identified
based on this threat model in \autoref{ss:eval-repro}.

\PP{Extensibility}
The current threat model focuses on the attacks
controlling external inputs
as they are the most imminent and realistic threats
to \adses.
However, since our fuzzer design is generic (\autoref{s:design}),
the threat model can be extended to other attacks,
such as sensor spoofing,
\eg, by introducing additional mutable components.

%% file: design.tex
\section{Design}
\label{s:design}

\subsection{Overview of \sys}
\sys is a feedback-driven mutational fuzzer
that mutates driving scenarios to test an \ads.
It aims to generate physically realistic, yet less-tested corner case driving scenarios
to discover safety-critical misbehaviors in the \ads.
%
\autoref{f:arch} illustrates the workflow of \sys
along with its four main components.

Provided an input driving scenario,
the \textbf{mutation engine} (\autoref{ss:mutation})
generates and
mutates various aspects associated with
the mission (initial and goal positions),
weather, actors (vehicles and pedestrians with their trajectories),
and puddles (areas with substantially low friction)
in the scenario.
The \textbf{test executor} (\autoref{ss:executor})
launches the \ads to be tested,
orchestrates the driving simulator
to prepare for the mutated driving scenario,
and assigns the mission to the ego-vehicle,
\ie, the vehicle
solely controlled by the \ads 
\cite{ogawa2006lane, seo2014tracking}.
%
While the ego-vehicle is carrying out the mission,
the \textbf{misbehavior detector} (\autoref{ss:misbehavior})
utilizes our driving test oracles 
to detect various safety-critical vehicular misbehaviors.
If the ego-vehicle
completes the mission
without any misbehavior,
the \textbf{driving quality feedback engine} (\autoref{ss:feedback})
quantifies the overall driving quality
by analyzing the vehicle states 
to guide further mutations
towards the generation of the scenarios that decrease the quality.
%

\subsection{Mutation Engine}
\label{ss:mutation}

\begin{figure}[t]
  \centering
  \includegraphics[width=0.82\columnwidth]{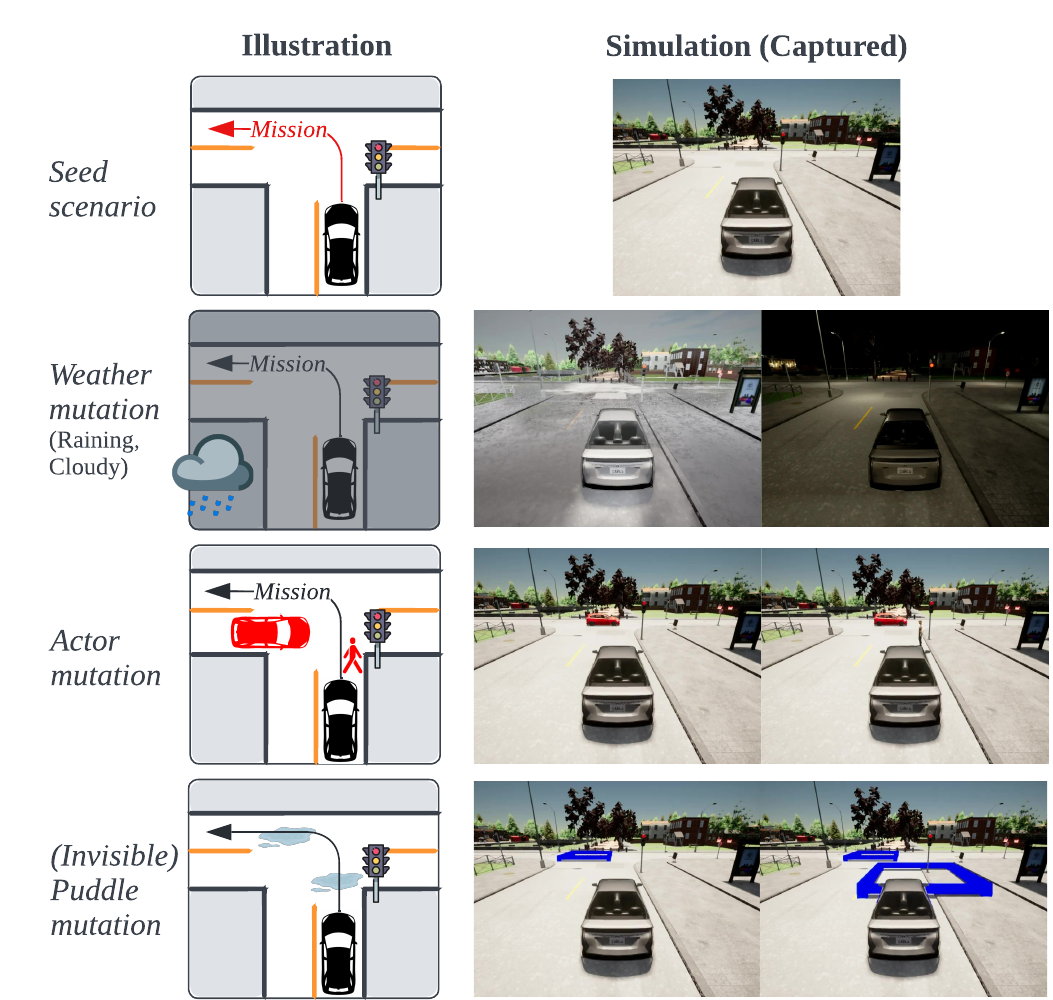}
  \vspace*{-0.1em}
  \caption{
    Examples of how mutations are applied
    to a seed scenario,
    in which the map and the mission are defined.
    Blue boxes indicating puddles are added for visualization
    and invisible during fuzzing.
    %
  }
  \label{f:mutation}
  \vspace*{-1em}
\end{figure}

\subsubsection{Test Input Driving Scenario}
The input space of an \ads
is extremely large and analogous to that of the real world,
along both temporal and spatial domains.
To efficiently explore the input space,
we identify key mutable components in a driving scenario
that can affect various components of an \ads when perturbed.
As illustrated in \autoref{f:mutation},
our driving scenario consists of
(1) a predefined 3D \textbf{map}
that mimics the real world with fair precision
using a standardized road network format~\cite{asam:opendrive},
(2) a \textbf{mission} defined by the initial and goal positions,
(3) \textbf{actors},
\ie, vehicles or pedestrians acting independently to the ego-vehicle,
(4) invisible \textbf{puddles} that affect the frictional force of the road,
and
(5) the \textbf{weather} conditions.


Similar to traditional mutational fuzzers
\cite{
  zalewski2014american, web:libFuzzer},
\sys best performs when seed scenarios to be evolved are given.
It is important to mention that generating input seeds is straightforward and does not require particular expertise in \ads. 
Specifically, for all the experiments in our evaluation, 
we construct input seeds from existing maps provided by the simulator,
where each map includes a set of valid waypoints
(\ie, a mission between two waypoints is guaranteed to be achievable).
We obtain the seeds by randomly selecting two of the predefined waypoints for the initial and goal points.
We further explain the details with example seeds in \autoref{s:seedgen}.

%






\subsubsection{Scenario Mutation}
\label{sss:mutation}
\sys's scenario mutation aims to gradually increase the mutated scenario's impact on the ego-vehicle under the test.
Specifically,
by generating and mutating the components of a scenario
according to the mutation schedule (\autoref{sss:feedbackdriven}),
it affects all four layers of the \ads,
as summarized in \autoref{tbl:mutcoverage}.
%

\begin{table}[t]
  \centering
  \scriptsize
  \caption{
    Layers of an \ads
    directly affected by taking fuzzing actions
    to each component of a driving scenario.
    With all actions combined,
    the coverage of a mutated scenario effectively spans all layers.
  }
  \label{tbl:mutcoverage}
  \vspace*{-0.5em}
  \input{tbl/tbl-mutcoverage}
  \vspace*{-2em}
\end{table}

\PP{Map and mission selection}
A map not only defines static world objects,
such as buildings and trees
that affect the perception module,
but also includes road structures,
such as intersections or curbs
that the planner actively interacts with.
The purpose of diversifying the mission
is to explore different parts of the map
and associated objects along with the road structure,
enabling \sys to thoroughly test diverse issues in the
perception and planning layers.

\PP{Actor generation \& mutation}
The actors in the scenario affect the sensing, perception, and planning layers,
because the behaviors of the actors
may force the \ads
to deviate from the original routing plan,
\eg, by blocking the path.

\sys \textit{generates} an actor by randomly selecting
the type of actor (either a vehicle or a pedestrian),
initial position and destination,
navigation method,
target speed,
and trajectory.
To have the actor generated within the interactable range of the ego-vehicle
(\ie, sensor range spanning the mission),
the initial position of an actor is always selected
from within a configurable range
from the ego-vehicle's initial position.
%
In addition,
to diversify the circumstances actors can render,
we define four kinds of navigation methods:
\begin{itemize}[leftmargin=*,nolistsep,noitemsep]
  \item [\it 1.] {\it Autopilot:}
    an actor performs a safe and lawful autopilot,
    using the ground truth traffic data
    and abiding by all traffic rules
    while heading to the destination.
  \item [\it 2.] {\it Maneuver:}
    an actor executes a sequence of maneuvers
    (\ie, drive forward, switch to left/right lane).
    Each maneuver has a pre-defined amount of time
    describing the duration of the action.
  \item [\it 3.] {\it Linear:}
    an actor blindly travels to the destination
    following a linear trajectory
    without considering the surrounding traffic or objects,
    thereby not complying with any rule.
  \item [\it 4.] {\it Immobile:}
    an actor remains stationary at the initial position.
\end{itemize}

\vspace{0.4em}
The \textit{mutation} of an actor includes a process of
modifying the aspects of a generated actor.
Except for the type and the navigation method,
all other aspects (\eg, the initial position) can be mutated
for an actor to exhibit a variety of behaviors.


%

\PP{Puddle generation \& mutation}
Invisible puddles (\eg, black ice)
reduce the surface friction of the road
and thereby affect the actuation of an \ads.
For example, based on the surface condition of the road,
or the tire condition,
an \ads has to adjust
the control commands accordingly
(\eg, avoid generating large torques on the wheels)
to ensure the ego-vehicle does not lose control.

\sys \textit{generates} a puddle
by randomly selecting
the location, size, and frictional force.
Similarly to the actors,
it \textit{mutates} a puddle
by modifying the location, size, and friction of a puddle.
%

\PP{Weather mutation}
Weather affects the sensing and perception layers,
which act as the eyes of an \ads.
\sys mutates the weather concerning
the following eight aspects:
rain, cloud, wind, fog, wetness, puddle,
solar azimuth angle, and solar altitude.
With a wide variety of available combinations,
a realistic weather condition can be simulated and tested.

\subsubsection{Ensuring Physically Valid Mutation}
\label{sss:semmut}

We aim to test an \ads
under physically feasible circumstances
that can occur in real life.
Thus,
all mutated driving scenarios need to be semantically practicable;
for example,
an actor should not suddenly appear in front of the ego-vehicle
during a simulation.
At the same time,
\sys should not ignore unusual yet possible scenarios
such as running into a person on a highway\footnote{The person could be the driver of a broken car stopped on the side road.}.
%
%
To this end, we ensure the testing always starts after the simulation is fully loaded with the scenario, including the weather condition and all actors/puddles, preventing the abrupt creation of any objects during testing.
In addition,
while allowing the random generation of objects,
we impose a spatial constraint and a temporal constraint
to prevent events defying the physical laws 
and to forestall false positive scenarios
where an ego-vehicle is not at fault of the misbehavior.

\PP{Spatial constraint}
To prevent unrealistic jams resulting in physically impossible scenarios,
such as two distinct vehicles being partially overlapped at an adjacent place,
the initial positions of all actors
are constrained to be at least a few meters away from each other.
The same constraint applies to the static objects (\eg, buildings, traffic lights);
the dimensions of the actors at their initial positions
cannot offend the bounding boxes of the static objects.

%
%
%

%
If the spatial constraint is violated, 
the mutation engine considers it as an infeasible scenario,
and attempts a mutation again
by randomly selecting the location of the actor
that violates the spatial constraint.
Note that
this random process results in
few additional computations (\eg, generating random numbers and checking the spatial constraint),
which only cause negligible overhead
(see \autoref{ss:eval-perf}).

\PP{Temporal constraint}
To prevent unrealistic movements of actors,
\sys imposes temporal constraints
by limiting the maximum speed of actor vehicles and pedestrians
to a conservative value, \eg, 20 and 6 $mph$, respectively.

\smallskip
The spatial and temporal constraints, combined with the navigation methods,
are designed to preclude most unrealistically reckless scenarios
that might lead to false positives.
For example,
a scenario in which
a pedestrian stands still until the ego-vehicle approaches
and then suddenly jumps in at the last moment to cause an unavoidable collision
cannot be generated
because (1) there is an initial distance between the pedestrian and the ego-vehicle (spacial constraint),
(2) the pedestrian cannot walk unrealistically fast (temporal constraint)
and (3) he/she walks either safely (autopilot), linearly to the destination (linear) or does not move (immobile), adhering to the navigation methods.


\subsubsection{Mutation Strategy}
\label{sss:strategy}

Depending on the particular aspect of the target \ads
to be stress-tested,
different mutation strategies specifying the mutable attributes and constraints
can be developed and applied.
The strategies we propose include, but are not limited to the following:
\begin{itemize}[leftmargin=*,nolistsep,noitemsep]
  \item Adversarial maneuver-based:
    only introduces and alters the maneuver of the adjacent actors,
    forcing interactions with the target system,
    \eg, an actor vehicle suddenly cutting the ego-vehicle off
    by switching lanes.
  \item Congestion-based:
    only introduces autopilot actors
    so that the target \ads drives in increasingly congested,
    yet lawful scenarios.
  \item Entropy-based:
    only introduces a linear or immobile actor,
    testing the ability of the target system to safely drive around
    reckless drivers and unlawful pedestrians.
  \item Instability-based:
    only inserts a puddle of different size and friction,
    testing the robustness of the motion controller
    to deal with sudden instabilities triggered by external forces.
\end{itemize}
Each strategy can be independently applied to a fuzzing campaign,
or orchestrated to be jointly applied under a probabilistic scheduling
(\eg, randomly selecting the next strategy to apply after each round).

\begin{figure}[t]
\removelatexerror
\begin{algorithm}[H]
  \scriptsize
  \caption{Driving quality feedback-driven fuzzing}
  \label{algo:fuzzing}
  \SetKwBlock{Begin}{procedure}{end procedure}
  \SetKwInOut{Input}{Input}
  \SetKwInOut{Output}{Output}

  \Input{$S$ - a set of seed scenarios (seed pool), $strategy$ - mutation strategy, \\
  $N_c$ - Maximum \# cycles, $N_p$ - Size of population}
  \Output{$bug$ - a detailed bug report, $s'$ - the buggy scenario}
  
  \ForEach{$seed \in S$}{
    $fuzz\_one(seed)$ \label{algo:line:repeat}
  }
  
  \Begin($fuzz\_one{(}seed{)}$) {
    $s \gets seed$ \\
    
    \For{$cycles \gets 1$ to $N_c$}{
      $s \gets mutator.generate(s, strategy)$ \label{algo:line:generate} // \autoref{sss:mutation}, \autoref{sss:strategy} \\
      $last\_worst\_score \gets 0$ \\
      
      \For{$rounds \gets 1$ to $N_p$}{
        $s' \gets mutator.mutate(s)$ \label{algo:line:mutate} // \autoref{sss:mutation}, \autoref{sss:semmut} \\
        $states \gets executor.simulate(s')$ \label{algo:line:simulate} // \autoref{ss:executor} \\
        
        \uIf{$detector.check\_misbehavior(states)$ /*\autoref{ss:misbehavior}*/ $== True$\label{algo:line:check}}{
          $save\_bug\_report(states, s')$ \\
          \Return \\
        }
        \Else{
          $score \gets feedback.check\_driving\_score(states)$ \label{algo:line:score} // \autoref{ss:feedback} \\
    
          \If{$score \leq last\_worst\_score$}{
            $last\_worst\_score \gets score$ \\
            $successor \gets s'$ \label{algo:line:worst} \\
          }
        }
      }
      $s \gets successor$ \label{algo:line:successor}
    }
  }
\end{algorithm}
\vspace*{-1em}
\end{figure}

\subsubsection{Feedback-driven Mutation Scheduling}
\label{sss:feedbackdriven}
To efficiently explore the input space,
\sys leverages a feedback mechanism to generate and mutate the components of a scenario as presented in \autoref{algo:fuzzing}.
At each cycle,
\sys first generates and
introduces an actor or puddle to the scenario
(\autoref{algo:line:generate}).
Then,
the generated component or the weather is mutated $N_p$ times to create
a population of size $N_p$ (\autoref{algo:line:mutate}).
Each mutated scenario is executed,
and its quality is evaluated
by the feedback engine (\autoref{ss:feedback}),
which measures the driving quality score (\autoref{algo:line:score}).
At the end of the cycle,
if none of the population triggers a misbehavior (\autoref{algo:line:check}),
they are ranked by the driving quality score,
and the one that scored the least among the population
is selected (\autoref{algo:line:worst})
and passed on to the next cycle (\autoref{algo:line:successor}).
\sys repeats the process of
adding a new component into this chosen scenario
and searching for the most ``harmful'' mutation
that disrupts the driving behavior of an autonomous vehicle most 
significantly.

As the fuzzing cycle repeats,
the scenario gradually gains intensity
as more actors and puddles are inserted.
%
However, more mutations may not always lead to a critical misbehavior. 
To prevent exploring less-promising directions of the mutation,
\sys aborts and starts a new campaign with a new seed (\autoref{algo:line:repeat})
when it reaches the maximum cycles ($N_c$)
without finding a misbehavior.


%

\subsection{Test Executor}
\label{ss:executor}
The test executor runs an \ads under the given driving scenario in a driving simulator,
collecting various vehicle states for the fuzzing process.
For the simulator,
we choose to use CARLA~\cite{Dosovitskiy17},
a high-fidelity driving simulator implemented using Unreal Engine.
CARLA is known for
its active development status and usage,
professionally designed realistic maps,
a wide range of supported sensors,
flexibility in controlling various aspects of a driving scenario,
and the ability to integrate various \adses with ease
by supporting Robot Operating System (ROS)~\cite{ros2009icra},
a universal middleware,
which many robotic systems are built on top of.

\subsubsection{Test Bridge}
The test bridge connects the mutation engine
to the \ads and the simulator, testing the mutated driving scenarios.

\PP{Loading the input driving scenario}
The test bridge first orchestrates the CARLA simulator
to set up the input scenario in the simulated world.
It connects to the simulation server,
opens the map, configures the weather,
spawns actors, puddles, and the ego-vehicle
as specified by the mutated input scenario.
When the loading is finished, 
the \ads is launched.

\PP{Initializing the target ADS for testing} 
The test bridge launches the \ads stack and waits until it is completely initialized.
Then, it attaches the autopilot functionality to the ego-vehicle
spawned in the simulated world.
Once the system is online and the autopilot agent is loaded,
a test is ready to be simulated.

\subsubsection{Driving Simulator}
\label{sss:simulator}
The driving simulator plays a key role in
synthesizing real-time sensor data
as well as computing vehicle states.
The simulator in the loop has multiple benefits 
compared to an alternative option of using a real vehicle~\cite{ohta2016pure}
equipped with
appropriate sensors and a companion computer
to bridge the \ads software
with the vehicular controllers.
We employ the simulator in \sys
to fully leverage the following benefits:
(1) test vehicles of different physical specifications
and self-driving software stacks without altering the testing scheme,
(2) test vehicles with significantly lower cost
compared to the testing of physical vehicles,
(3) test vehicles under various circumstances
including but not limited to unlikely situations
without physical constraints,
and
(4) fully automate a testing sequence.

In a loop,
CARLA is responsible for simulating each frame
by applying the control commands issued by the \ads
to the ego-vehicle,
and updating the states of in-simulation actors,
\eg, the position of a pedestrian moving at 1.5 $m/s$ towards the North.
The \ads
combines the updated states of the ego-vehicle
with the new sensory data read from the simulator
to decide the subsequent control command.
The loop terminates
when the vehicle reaches the destination,
or any issue is found by the misbehavior detector.

\subsection{Misbehavior Detector}
\label{ss:misbehavior}
When the \ads fails to handle the input scenario,
it can lead to a wide spectrum of undesirable consequences
from software-oriented errors (\eg, memory error in a component)
to vehicular misbehaviors (\eg, collision).
Our misbehavior detector intends to point out
\emph{obvious illegal acts} in the driving behaviors of \ads
by applying definitive standards.
%
%
Inspired by the fact that
the \adses are designed
to drive in the real world
complying with traffic rules and regulations~\cite{national1972traffic},
we build the following three driving test oracles
that check for the events that are closely related to human safety:
\emph{collisions}, \emph{infractions}, and \emph{immobility}
of the ego-vehicle.

\begin{itemize}[leftmargin=*,nolistsep,noitemsep]
    \item \PP{Collision}
    Collision is one of the most destructive events
    that can cause significant damage to human drivers.
    By attaching a collision sensor to the ego-vehicle
    (that would be corresponding to multiple sensors around a real vehicle),
    a collision to any object is captured and reported.
    
    \item \PP{Infraction}
    Infractions are traffic violations
    including
    (1) speeding,
    (2) invading lanes, and
    (3) running on red lights,
    which are directly involved in
    approximately 30\%, 8.5\%, and 4\%, respectively,
    of the annual fatal accidents
    in the United States in 2018~\cite{nhtsa:trafficsafety2018}.
    %
    As \sys has full access to the simulated space,
    it 
    %
    compares
    the states of the vehicle (\eg, current speed)
    with the defined traffic rule (\eg, speed limit)
    to check for any violation.

    \item \PP{Immobility}
    A vehicle that is not moving at a particular location
    would become a cause of subsequent undesirable events
    such as collisions
    (\eg, a car stopped in the middle of an intersection
    would cause other cars to crash into it).
    The immobility monitor measures the time duration
    when the vehicle is not moving,
    excluding legitimate stops (\eg, at traffic lights).
    If it exceeds a threshold ($60\;sec$ in this paper),
    that is considered a misbehavior.
\end{itemize}

\smallskip
The misbehavior detector monitors every frame of the simulation
and refers to these oracles to check for any violation
(\autoref{algo:line:check} of \autoref{algo:fuzzing}).
Upon detecting a misbehavior,
the incident is reported, and
the simulation is terminated immediately after
logging all vehicle states for a later inspection.


\subsection{Driving Quality Feedback Engine}
\label{ss:feedback}
We propose a new \emph{driving quality metric}
that abstracts the performance of \ads under a testing scenario.
In particular,
the metric is measured by evaluating various events in the driving maneuvers during testing
that does not immediately trigger safety-critical misbehaviors,
but are likely to lead to those.
The metric is later used to guide the input scenario mutation towards buggy conditions. 

Note that we develop this new metric
because existing metrics such as code coverage
are not suitable for our context.
Specifically,
while the code coverage-guided mutation has proven effective in many modern grey-box fuzzers
to approximate the amount of the explored input space
for \emph{sequential} programs,
it is not effective for distributed and \emph{stateful} systems
such as an \ads.
%
%
In particular,
\ads runs smaller nodes changing states driven by data,
consisting of loops running state machines.
Their code coverage quickly saturates regardless of the testing progress,
hence inadequate to approximate the test coverage
(see \autoref{s:afl}).

\subsubsection{Driving quality measurement}
\label{ss:qualitymeasure}
When no safety-critical misbehavior is detected,
\sys analyzes the driving data to guide the input mutator
so that it can effectively mutate the input scenario
towards the scenario \emph{likely to} trigger safety-critical misbehaviors.
To quantitatively measure how close a vehicle is to the
safety-critical misbehaviors,
we refer to the official reports~\cite{najm2007pre, national2008national}
from the U.S. Department of Transportation,
National Highway Traffic Safety Administration (NHTSA), which
investigates the causes of traffic accidents.
According to the reports, 52\% of the fatal accidents of known causes
are attributed to either \emph{reckless} or \emph{clumsy} driving behaviors,
such as hard acceleration or oversteer.
%
Many major car insurance companies
(\eg, Allstate, Progressive, and State Farm)
also support this idea by
having their programs evaluate
the driving quality
based on the number of
hard braking, hard acceleration, and hard turning events
to determine the insurance rate~\cite{ins:allstate, ins:progressive, ins:statefarm, ins:root}.

%

%
Inspired by the real-world usage,
\sys measures the driving quality
based on the number of
hard accelerations, hard brakings, hard turns,
oversteers and understeers,
and the minimum distance to other actors.
%
The following paragraphs \WC{1}--\WC{4} present
how we measure each factor constituting the driving quality
by leveraging the vehicle states on the driving data of the simulation.

\noindent \WC{1} \textbf{Hard acceleration and hard braking detection.}
The ratio of longitudinal acceleration of a vehicle $A_x$
to the gravitational constant $g$
(approximately 9.8 $m/s^2$) 
is a generally accepted way of representing
the harshness of acceleration or braking events~\cite{botzer2019relationship, hill2019computer}.
%
The hard acceleration/braking indicator $K_{ab}$ is given by:
\setlength{\abovedisplayskip}{2.0pt}
\setlength{\belowdisplayskip}{2.0pt}
\begin{equation}
  K_{ab} = { A_x } / { g }
\end{equation}
If $K_{ab}$ exceeds a threshold, \sys counts the frame
as either a hard acceleration or hard braking event.
For the threshold,
%
%
NHTSA used $0.4-0.6$
to identify hard acceleration or hard braking~\cite{dingus2006100, klauer2009comparing}.
Other studies claim that
$0.5$ is the threshold
people typically agree on~\cite{botzer2019relationship, hill2019computer}.
Taking the upper bound,
we use $0.6$ as a decision boundary for $K_{ab}$,
which is the force
that a vehicle can reach or stop from 60 $mph$
in less than five seconds.
%
%
%
%
%
\begin{equation}
  \#ha = count(K_{ab} \geq 0.6), \; \#hb = count(K_{ab} \leq -0.6)
\end{equation}

\noindent \WC{2} \textbf{Hard turn detection.}
A hard turn occurs
when a driver tries to turn the vehicle
at an excessive speed.
As a hard turn is related to
the lateral force applied to the vehicle,
we leverage a detection algorithm
that uses a hard turn indicator $K_t$, such that
\begin{equation}
  K_t = { V_y } / { SWA }
\end{equation}
where $V_y$ and $SWA$ denote
the lateral speed, and the steering wheel angle,
respectively.
If
(1) $SWA$ is greater than a steering threshold, and
(2) $K_t$ is above a hard turn threshold,
\sys counts the frame as a hard turn.
Both thresholds are configurable,
and we empirically determined them as 20 and 0.18, respectively,
such that
\begin{equation}
  \#ht = count(SWA \geq 20 \wedge K_t \geq 0.18)
\end{equation}

\noindent \WC{3} \textbf{Oversteer and understeer detection.}
Oversteer and understeer
represent the reaction of a vehicle
to the steering effort.
Oversteer occurs when the rear tires lose grip
and the vehicle turns more than the amount the driver steers,
and understeer occurs
when the front tires lose grip,
so the vehicle turns less than the steering amount.
Both frequently occur in competitive racing sports
where aggressive controls are required,
and they often lead to accidents
as a vehicle loses control and slips while turning.
In normal driving conditions,
oversteer or understeer can take place
as a result of imprecise control,
or because of low friction on the road
caused by natural events, such as black ice.
No matter what the cause is,
both are 
deemed very dangerous~\cite{gysen2009design, jang2003control},
being ranked in the 7th in ``top 12 causes of fatal car accidents in the USA''
by NHTSA.

Broadly,
there are two approaches
that attempt to detect oversteer and understeer events:
model-based detection and
fuzzy logic-based detection.
Model-based detection tends to be accurate
but requires precise models of the vehicles, tires, and friction.
%
On the other hand,
fuzzy logic~\cite{zadeh1988fuzzy, novak2012mathematical} 
approximates the ``truthiness'' of a linguistic statement
on a continuum as a fuzzy value
rather than a boolean value
and aggregates multiple values with rules
to infer the final level of output.
To grasp the overall safety
and generate feedback,
\sys does not require the detection to be meticulously accurate.
Moreover,
model-specific detection is ill-suited to the purpose of \sys
to serve as a generic framework
for testing \ads
planted on various vehicle models.
Thus,
we adopted the fuzzy logic-based detection
proposed by~\cite{anderson2011fuzzy, pandit2013model}
that works reasonably well across different vehicle models.

In summary,
four indicators,
$SWA$ (steering angle in $deg$),
$V_x$ (longitudinal velocity in $km/h$),
$AV_z$ (yaw rate in $deg/s$), and
$A_y$ (lateral acceleration in $gs$),
which can be obtained from the driving data
are used for fuzzy logic
to compute the degree of oversteer $K_{os}$ and understeer $K_{us}$.
%
%
With the inferred oversteer and understeer levels,
which are floating-point numbers in $\{0, 1\}$,
we tuned the threshold to determine the final results
as follows:
\begin{equation}
  \#os = count(K_{os} \geq 0.4),
  \#us = count(K_{us} \geq 0.4)
\end{equation}

\noindent \WC{4} \textbf{Minimum distance}
Any failure to maintain a safe distance
from other vehicles or pedestrians 
implies that the system is close to potential misbehaviors.
For example,
if a minimum distance to a pedestrian is one foot,
we can interpret that
as the ego-vehicle near-missed hitting the pedestrian,
and with a slight mutation,
the scenario could cause a collision.
To take such events into account,
\sys measures the distances from the ego-vehicle
to all other actors per frame
and keeps track of the minimum distance, $md$.
The smaller $md$ is,
the more deduction is applied to the driving score.

\PP{Overall driving quality score}
With all the ingredients ready,
\sys computes the overall driving quality score
by multiplexing the number of events. 
The driving quality score starts from zero,
and the number of the events captured above is deducted,
and then the inverse of the minimum distance (\ie, $1/md$)
multiplied by a configurable coefficient is deducted,
resulting in the final feedback score.
In summary,
the driving quality $score$ is given by:
\begin{equation}
\label{eq:score}
    score = -(\#ha + \#hb + \#ht + \#os + \#us + c/md)
\end{equation}
%
The final $score$ is delivered to the input mutator
for the decision of the scenario that is worth further mutating
(\autoref{algo:line:score} of \autoref{algo:fuzzing}).

\PP{Tuning metrics}
Weights for the driving quality factors can be configured to prioritize certain misbehavior,
depending on the users' needs and the characteristics of the target system.
In our experiments, we treat all factors equally (\ie, \autoref{eq:score})
to prevent \sys from being biased toward any particular misbehavior.

\subsubsection{Key contribution of driving quality feedback}
\label{ss:targetedfeedback}
In the context of testing \adses,
our design of physical vehicular states-based driving quality feedback
is highly pertinent for two reasons.
First,
using the physical states of a vehicle,
it allows \sys to pragmatically quantify the recklessness of the driving
without requiring code-level analysis nor examining internal states,
of which the availability is not always guaranteed.
Second,
unlike the feedback suggested by the related work~\cite{li2020av}
that may lead \ads away from the bugs we detected
(see \autoref{ss:comp-avfuzzer}),
our fine-grained feedback mechanism
provides proper guidance towards unsafe driving scenarios,
resulting in the detection of actual, safety-critical misbehaviors.

%% file: tbl/tbl-mutcoverage.tex
\begin{tabular}{p{8em} p{10em} p{14em}}
  \toprule
  \textbf{Component} & \textbf{Action} & \textbf{Affected layers} \\
  \midrule
  Map and mission & Seed selection & Sensing, Perception, Planning \\
  Actor           & Generation \& Mutation & Sensing, Perception, Planning \\
  Puddle          & Generation \& Mutation & Planning, Actuation \\
  Weather         & Mutation & Sensing, Perception \\
  \bottomrule
\end{tabular}

%% file: impl.tex
\begin{table}[t]
  \scriptsize
  \centering
  \caption{Implementation complexity of \sys.}
  \vspace*{-0.5em}
  \input{tbl/tbl-impl}
  \label{tbl:impl-loc}
  \vspace*{-2em}
\end{table}

\begin{table*}[ht!]
  \centering
  \scriptsize
  \caption{New bugs \sys revealed
  in multiple layers of Autoware,
  Behavior Agent,
  and CARLA simulator.
  Impact indicates which system-level misbehaviors
  were captured by the driving test oracles during testing,
  strategy shows the mutation strategy used,
  and the root cause is determined by our manual analysis afterward.
  ACK indicates whether bugs are confirmed by the developers.
  %
  }
  \vspace*{-0.5em}
  \input{tbl/tbl-bugs}
  \label{tbl:bugs}
  \vspace*{-0.5em}
\end{table*}

\section{Implementation}
\label{s:impl}

\sys is prototyped in approximately 2.3K lines of Python 3 code, as shown in \autoref{tbl:impl-loc}.
%

\PP{ROS and portability}
ROS~\cite{ros2009icra} is a de facto middleware that provides
a means of message passing between distributed nodes,
hardware abstraction,
and a toolset 
for the easier development of robotic systems.
%
\sys incorporates ROS in the design of the test executor
and makes any ROS-based ADSes~\cite{kato2015open, kato2018autoware}
and simulators~\cite{Dosovitskiy17, airsim2017fsr, lgsvl2020itsc}
pluggable into the system.
%


\PP{Bug reproduction}
ROS leverages a publisher-subscriber message passing scheme;
nodes publish messages to a topic,
and other nodes subscribe to the topic to receive the messages.
Thus,
all flows
including sensory inputs and control commands
are summarized in the messages.
\sys records all underlying ROS messages,
essentially capturing all data flows that
happened during fuzzing,
and later replays them 
to reproduce and debug the buggy scenarios.

\PP{Clock synchronization}
Depending on the hardware,
the simulation could run slower than a wall clock
and stall \adses from obtaining real-time sensor data.
%
%
\sys synchronizes \adses with the simulator's time, not the wall clock,
so that
if the simulation runs behind the wall clock
while computing and rendering each frame,
\adses can wait for the data and react upon correctly.

%% file: tbl/tbl-impl.tex
\begin{tabular}{llrl}
\toprule
\textbf{} & \textbf{Component} & \textbf{LoC} & \textbf{Language} \\
\midrule
\multirow{5}{*}{\rotatebox{90}{\parbox{5em}{\centering \sys\\framework}}}
& Mutation engine                               & 440  & Python \\
& Misbehavior detector and driving test oracles & 119  & Python \\
& Feedback engine and driving quality metrics   & 636  & Python \\
& Test executor                                 & 1125 & Python \\
& Additional bridge for Autoware                & 48   & Shell script \\
\bottomrule
\end{tabular}

%% file: tbl/tbl-bugs.tex
\resizebox{0.98\textwidth}{!}{
\setlength{\tabcolsep}{3pt}
\begin{tabular}{p{0.1em} c l l l c c l c}
\toprule
  \textbf{} &
  \textbf{Bug \#} &
  \textbf{Layer} &
  \textbf{Component} &
  \textbf{Description} &
  \textbf{Impact} &
  \textbf{Strategy} &
  \textbf{Root cause} &
  \textbf{ACK} \\

\midrule
  \multirow{17}{*}{\rotatebox{90}{Autoware}}
  & 01 & Sensing    & Fusion    & LiDAR \& camera fusion misses small objects on road & C & all & Logic err & \\
  & 02 & Perception & Detection & Perceives the road ahead as an obstacle at a steep downhill & I & all & Logic err & \V \\
  & 03 & Perception & Detection & Fails to semantically tag detected traffic lights and cannot take corresponding actions & C, V & all & Logic err & \\
  & 04 & Perception & Detection & Fails to semantically tag detected stop signs and cannot take corresponding actions & C, V & all & Logic err & \\
  & 05 & Perception & Detection & Fails to semantically tag detected speed signs and cannot take corresponding actions & V & all & Logic err & \\
  & 06 & Perception & Localization & Faulty localization of the base frame while turning & C, L & all & Logic err & \V \\
  & 07 & Perception & Localization & Localization error when moving underneath bridges and intersections & C, L & all & Logic err & \V \\
  & 08 & Planning   & Global planner & Generates infeasible path if the given goal is unreachable & C, L & all & Logic err & \V \\
  & 09 & Planning   & Global planner & Generates infeasible path if the goal's orientation is not aligned with lane direction & C, I, L  & all & Logic err & \V \\
  & 10 & Planning   & Global planner & Global path starts too far from the vehicle's current location & C, I, L  & all & Logic err & \V \\
  & 11 & Planning   & Local planner & Target speed keeps increasing at certain roads, overriding the speed configuration & S, C & all & Logic err & \V \\
  & 12 & Planning   & Local planner & Fails to avoid forward collision with a moving object & C & all & Logic err & \\
  & 13 & Planning   & Local planner & Fails to avoid lateral collision (\ads perceives the approaching actor before collision) & C & ent & Not impl &  \\
  & 14 & Planning   & Local planner & Fails to avoid rear-end collision (\ads perceives the approaching actor before collision) & C & ent & Not impl & \\
  & 15 & Planning   & Local planner & While turning, ego-vehicle hits an immobile actor partially blocking the intersection & C & ent & Logic err & \\
  & 16 & Actuation  & Pure pursuit & Ego-vehicle keeps moving after reaching the destination & C, L & all & Logic err & \V \\
  & 17 & Actuation  & Pure pursuit & Fails to handle sharp right turns, driving over curbs & C, L & all & Faulty conf & \\
\midrule
  \multirow{13}{*}{\rotatebox{90}{Behavior Agent}}
  & 18 & Perception & Detection & Indefinitely stops if an actor vehicle is stopped on a sidewalk & I & ent & Logic err & \\
  & 19 & Perception & Detection & Flawed obstacle detection logic; lateral movement of an object is ignored & C & con & Logic err & \\
  & 20 & Planning & Global planner & Generates inappropriate trajectory when initial position is given within an intersection & C, L, V & all & Logic err & \\
  & 21 & Planning & Local planner & Improper lane changing, cutting off and hitting an actor vehicle & C & man & Logic err & \\
  & 22 & Planning & Local planner & Vehicle indefinitely stops at stop signs as planner treats stop signs as red lights and waits for green & I & all & Logic err & \\
  & 23 & Planning & Local planner & Vehicle does not preemptively slow down when the speed limit is reduced & S & all & Logic err & \\
  & 24 & Planning & Local planner & Always stops too far (> 10 m) from the goal due to improper checking of waypoint queue & F & all & Logic err & \\
  & 25 & Planning & Local planner & Collision prevention does not work at intersections (only checks if actors are on the same lane) & C & all & Logic err & \\
  & 26 & Planning & Local planner & Fails to avoid lateral collision (\ads perceives the approaching actor before collision) & C & man & Not impl & \\
  & 27 & Planning & Local planner & Fails to avoid rear-end collision (\ads perceives the approaching actor before collision) & C & man & Not impl & \\
  & 28 & Planning & Local planner & No dynamic replanning; the vehicle does infeasible maneuvers to go back to missed waypoints & C, L & ins & Not impl & \\
  & 29 & Actuation  & Controller & Keeps over-accelerating to achieve the target speed while slipping, creating jolt back on dry surface & C, L & ins & Not impl & \\
  & 30 & Actuation  & Controller & Motion controller parameters (PID) are poorly tuned, making the vehicle overshoot at turns & C, L & all & Faulty conf & \\
\midrule
  \multirow{3}{*}{\rotatebox{90}{CARLA}}
  & 31 & \multicolumn{2}{c}{Simulator} & Simulation does not properly apply control commands & C, L, V & all & Logic err & \V \\
  & 32 & \multicolumn{2}{c}{Simulator} & Vector map contains a dead end blocked by objects as a valid lane & I, V & all & Data err & \\
  & 33 & \multicolumn{2}{c}{Simulator} & Occasionally inconsistent simulation result & I, V & all & Logic err & \V \\

\midrule
  \multicolumn{9}{r}{
    [Impact]
    \textbf{C}: \textbf{C}ollision /
    \textbf{F}: \textbf{F}ails to complete a mission /
    \textbf{I}: Vehicle becomes \textbf{I}mmobile /
    \textbf{L}: \textbf{L}ane invasion /
    \textbf{S}: \textbf{S}peeding /
    \textbf{V}: Miscellaneous traffic \textbf{V}iolation
  } \\
  \multicolumn{9}{r}{
    [Strategy]
    \textbf{all}: all strategies /
    \textbf{man}: Adversarial \textbf{man}euver-based /
    \textbf{con}: \textbf{con}gestion-based /
    \textbf{ent}: \textbf{ent}ropy-based /
    \textbf{ins}: \textbf{ins}tability-based
  } \\

\bottomrule
\end{tabular}
}

%% file: eval.tex
\section{Evaluation}
\label{s:eval}

We evaluate the effectiveness of \sys
as a fuzzer for \adses
by assessing
the number of bugs detected by \sys (\autoref{ss:eval-numbugs})
with their analyses (\autoref{ss:casestudy}),
the feasibility of exploiting the discovered bugs in the real world
(\autoref{ss:eval-repro}),
how \sys fares against a state-of-the-art approach (\autoref{ss:comp-avfuzzer}),
the correctness of the driving test oracles (\autoref{ss:eval-oracle}),
the correctness of driving quality measurement (\autoref{ss:eval-quality}),
the effectiveness of feedbacks (\autoref{ss:eval-feedback}),
and the fuzzing performance (\autoref{ss:eval-perf}).

\PP{Experimental setup}
We ran \sys on a server machine running Ubuntu 18.04,
powered by
16-core Intel Xeon Gold 5218 CPU,
192-GB main memory,
and 8 GeForce RTX 2080 Ti graphics cards.
To allow parallel execution of testing workloads
and increase the testing performance,
we used Docker containers.
We simultaneously ran four pairs of CARLA and \ads containers
connected via a ROS bridge,
and assigned a dedicated GPU to each container.
For the effectiveness and performance evaluations
(\autoref{ss:eval-feedback} and \autoref{ss:eval-perf})
where randomness can skew the results,
we report the average of repeated runs,
following the suggestions in~\cite{klees2018evaluating}. 

\PP{Test targets}
We tested the following \adses:
\begin{itemize}[leftmargin=*,nolistsep,noitemsep,topsep=1pt]
  \item Autoware:
    A full-fledged \ads
    with active development status.
    Started in 2015,
    it has been internationally adopted
    by many well-known automobile manufacturers,
    \eg, BMW~\cite{aeberhard2015automated},
    and qualified to run driverless vehicles on public roads
    in Japan since 2017~\cite{autoware:vision}.
    %
  \item Behavior Agent:
    An \ads developed by CARLA,
    implementing path planning, feedback-based PID control,
    compliance with traffic laws, and collision avoidance.
\end{itemize}

%
%
%

\PP{Seed scenarios}
For the experiments,
we used 40 valid seed scenarios
to test target systems in various environments and conditions,
obtained by the procedure described in \autoref{s:seedgen}.

\subsection{Detected Misbehaviors}
\label{ss:eval-numbugs}
%
%
\sys found multiple scenarios that trigger various safety-critical misbehaviors,
stemming from
a total of \allbugs bugs in Autoware, Behavior Agent, and CARLA;
\anewbugs previously unknown bugs
and \aoldbugse known issue in Autoware,
of which 8 bugs are already confirmed,
and \ballbugs bugs in Behavior Agent,
which are awaiting confirmation.
%
In CARLA,
\callbugse critical simulation bugs are detected,
and \cackbugse of them have been acknowledged
so far. 
All bugs have been responsibly reported to and discussed with the developers.

\autoref{tbl:bugs} summarizes
all the new bugs we found, the component they are located in, the impact and the root cause of each bug.
By comprehensively testing the entire \ads end-to-end
with high-fidelity driving scenarios,
\sys identifies misbehaviors
from all components of the system,
including
sensing, perception, planning, and actuation.
The videos of the bugs are available at
\textbf{\url{https://youtube.com/channel/UCpCrUiGanDKX-qxj8jcUVGQ}}.
%

\PP{Root cause identification}
To identify the bug and the root cause of observed misbehaviors,
we replay the recorded simulation data and analyze critical events.
This procedure involves a component-wise data- and control-flow analysis
of answering the following diagnostic questions:
\begin{itemize}[leftmargin=*,nolistsep,noitemsep,topsep=1pt]
  \item Did sensors accurately read the environment?
  \item Did the perception layer correctly interpret the sensor data?
  \item Did the planning layer find a feasible path?
  \item Did the actuation layer emit appropriate control commands?
\end{itemize}

\PP{Contribution of test oracles}
The ``Impact'' column in \autoref{tbl:bugs}
shows the accumulation of all misbehaviors triggered in multiple scenarios,
which stem from the same root cause.
For example, bug \#28 (impact C, L) caused a collision in some scenarios,
and a lane invasion in other scenarios,
depending on the location on the map and the nearby objects
at the moment the bug was triggered.
%
%
Overall, the collision oracle contributed to the detection of the most (76\%) bugs,
because \ads bugs usually make the vehicle lose control
and susceptible to a collision.
Traffic infractions (V, L, and S) were triggered by 60\% of the bugs.
We can also observe that all bugs that caused a lane invasion caused a collision as well.
This does not imply that the collision oracle can replace the lane invasion oracle;
these oracles were individually triggered in different scenarios.
For versatility under any circumstances, both oracles should be utilized.

\input{casestudy}

\subsection{Feasibility of Bug Exploitation}
\label{ss:eval-repro}

It is feasible to reliably exploit all \anewbugs Autoware bugs (except for two) 
and all \ballbugs Behavior Agent bugs,
adhering to the threat model presented in \autoref{s:model};
controlling external inputs. 
Specifically,
we evaluate the viability of launching
object-based attacks or location-based attacks
targeting the discovered bugs.

\begin{figure}[t]
  \centering
  \includegraphics[width=\columnwidth]{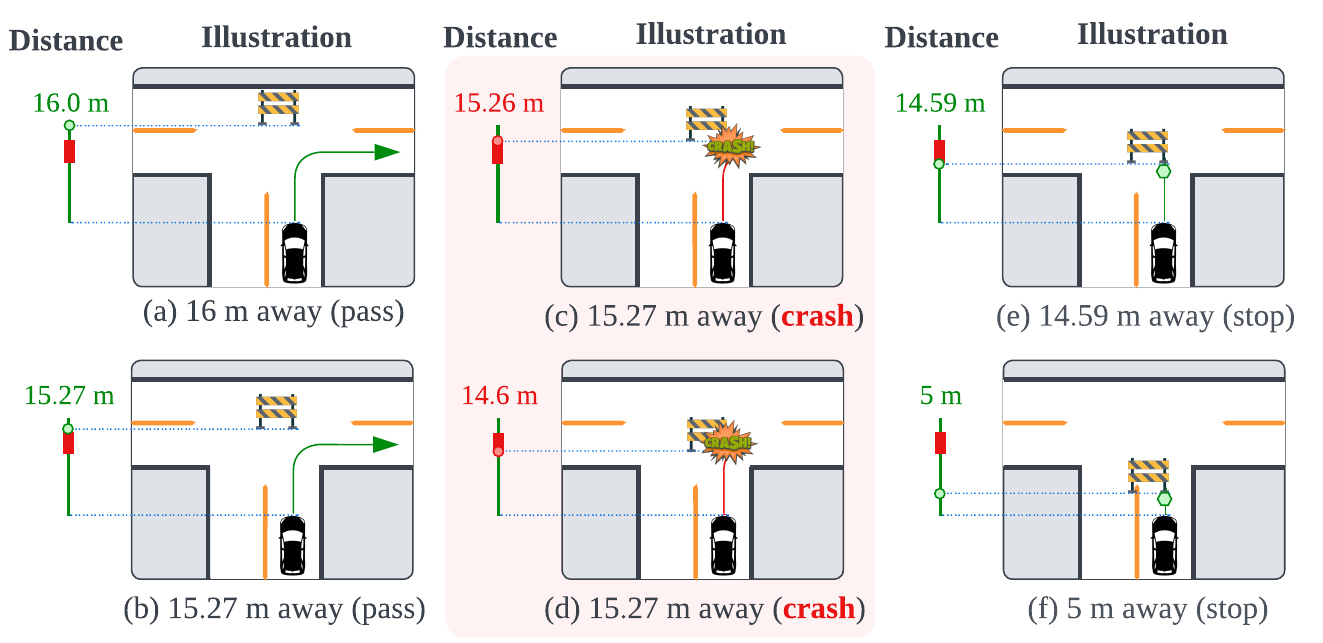}
  \vspace*{-1em}
  \caption{
    Testing the variants of bug \#15
    by changing the distance of the obstacle
    from the initial position of Autoware ego-vehicle.
    %
    %
    When the obstacle is at a moderate distance (14.6--15.26 m),
    \ie, (c) and (d),
    the ego-vehicle initiates a turn and hits the object,
    even though it senses and perceives the existence of the object correctly.
  }
  \label{fig:bug15}
  \vspace*{-1em}
\end{figure}

\PP{Object-based attacks}
There are 11 bugs (\#1, 12--15, 18, 19, 21, 25--27)
that enable object-based attacks.
To understand how easy it is to exploit the bugs in the real world,
we run experiments on how sensitive each bug is 
to multiple variables (\ie, potential requirements of the exploitation),
including the color and shape (model) of the vehicle,
location/trajectory of the controlled object, and the weather.
If a bug requires many attributes for its exploitation,
it essentially means that its exploitability is low.
As shown in \autoref{tbl:attack-obj},
only one attribute needs to be controlled for all bugs,
except for bug \#1,
which requires two attributes to be controlled.
%
Specifically,
to exploit bug \#1, an attacker can use an object
of any color
at any location near the path of the ego-vehicle
as long as its height is lower than approximately 50 cm,
which is commonplace.
Bugs \#13, 14, 26, and 27 are even easier to exploit;
having any vehicle of any model and color, or pedestrian
approach towards the ego-vehicle from behind or side
is sufficient.
%
The same set of attributes does not affect bug \#12.
However, after moving into the ego-vehicle's path,
the object (vehicle or pedestrian) has to be located
within one meter of the ego-vehicle's front bumper.
For bug \#15,
the only relevant attribute is the location of the object.
As illustrated in \autoref{fig:bug15},
the adversarial object has to be placed within a certain range of distance
from the ego-vehicle, which allows a window of 66 cm.
While it seems tight, it is obviously viable,
considering that the window is
approximately a third of the width of a mid-sized sedan.

\begin{table}[t]
  \centering
  \footnotesize
  \caption{
    Enumeration of the object types, 
    the attributes that are irrelevant to the successful attack (\fx, meaning that they can have arbitrary values),
    and the attributes that should be controlled.
  }
  \vspace*{-0.5em}
  \resizebox{1.0\columnwidth}{!}{%
  \setlength{\tabcolsep}{3pt}
  \begin{tabular}{c c c c l c c c c c l}
    \toprule
    \multirow{2}[1]{*}{\textbf{Bug \#}} & \multicolumn{3}{c}{\textbf{Object type}} & & \multicolumn{5}{c}{\textbf{Irrelevant}} & \multirow{2}[1]{*}{\textbf{Need to control}} \\
    \cmidrule{2-4} \cmidrule{6-10}
    & {\bf O} & {\bf P} & {\bf V} & & {\bf C} & {\bf L} & {\bf S} & {\bf T} & {\bf W} & \\ \midrule
    
    01     & \fo &     &     &    & \fx &     &     & \fx & \fx & L: be close to the path, S: height < 50cm \\
    12, 19 &     & \fp & \fc &    & \fx & \fx & \fx &     & \fx & T: cut in from side to dist < 1m \\
    13, 26 &     & \fp & \fc &    & \fx & \fx & \fx &     & \fx & T: approach from behind \\
    14, 27 &     & \fp & \fc &    & \fx & \fx & \fx &     & \fx & T: approach from side \\
    15     & \fo & \fp & \fc &    & \fx &     & \fx & \fx & \fx & L: located within 66cm range \\
    18     &     &     & \fc &    & \fx &     & \fx & \fx & \fx & L: located on sidewalk \\
    21     &     &     & \fc &    & \fx & \fx & \fx &     & \fx & T: drive at a similar speed alongside \\
    25     & \fo & \fp & \fc &    & \fx &     & \fx & \fx & \fx & L: located on the cross lane of intersection \\
    \midrule
    \multicolumn{11}{r}{
    [Object types] O: object / P: pedestrian / V: vehicle
    } \\
    \multicolumn{11}{r}{
    [Attributes] C: color / L: location / S: shape / T: trajectory / W: weather
    }\\
    \bottomrule
  \end{tabular}
  }
  \label{tbl:attack-obj}
  \vspace*{-1em}
\end{table}

\PP{Location-based attacks}
Bugs \#2--11, 16, 17, 20, 22--24, 28--30 can be exploited
by taking advantage of location-based attacks.
%
Bugs \#3--5 can be triggered at any location
with traffic lights, stop signs, or speed limit signs,
regardless of the weather condition,
as they stem from software errors
not being able to find matching tags for detected objects.
Bugs \#8 and 9 are triggered immediately when
the planning layer receives the mission,
if the goal position is unreachable 
or not aligned with the lane. 
An attacker may provide such adversarial destinations to the system
through social engineering,
\eg, sharing a Google Maps link that sets the destination through navigation API.
Bugs \#2, 7, 16, 17, 28, and 29 require an attacker
to lure the ego-vehicle to any location that has a certain property;
any downward slope of an angle greater than 30 degrees
that abruptly flattens at the end (bug \#2),
any location under a bridge
that the vehicle has already passed over,
\eg, an underpass of an interchange (bug \#7),
any destination at a location the ego-vehicle
can sufficiently accelerate before reaching it,
\eg, the end of a long straight road (bug \#16),
any 90-degree curve connecting the rightmost lanes,
which can be observed at most three-way or four-way intersections,
(bug \#17),
and puddles covering an area vehicle turns (bug \#28, 29).
Bug \#6 and \#30 happen at arbitrary curves.
Notably,
bugs \#10 and \#11 are the only bugs that require a specific location of the map,
and thus can be harder to exploit in the real world.

\PP{Summary}
All 30 bugs except for two (\#10 and \#11)
have a wide window of exploitation in the input space
that an adversary can easily control in the real world to cause safety-critical misbehaviors.
%

\subsection{Comparison with \avf}
\label{ss:comp-avfuzzer}

\avf~\cite{li2020av} is a state-of-the-art \ads testing approach
that mutates the trajectory of two actor vehicles
driving nearby, aiming to detect vehicle-to-vehicle collisions.
It uses the longitudinal distance from the ego-vehicle
to actor vehicles as a fitness function for the mutation
to create scenarios with smaller distances.
It detected five buggy scenarios:
(1) hitting an overtaking vehicle,
(2) hitting another vehicle while trying to cut in,
(3) hitting a vehicle that cuts in,
(4) rear-ending a suddenly braking vehicle,
and (5) interpreting two adjacent vehicles as one and hitting one.

\PP{Quantitative comparison}
\sys was able to automatically generate all five crash scenarios \avf found
and successfully detected misbehaviors (bugs \#12-14, 19, 21, 25-27).
On the other hand,
\avf is bound to miss 26 out of \allbugs (76\%) bugs \sys found
due to fundamental limitations in the design.
We discuss the reasons in the following,
referring to the latest source code of \avf\footnote{
\url{https://github.com/cclinus/AV-Fuzzer/tree/4f67868/freeway}}.
%
First,
the input space of \avf is a subset of \sys's driving scenarios.
\avf divides a scenario into five time-slices
(line 10 in \cc{drive_experiment.py} and lines 11-34 in \cc{Chromosome.py}),
and randomly mutates the target speed and the maneuver
(\eg, go straight, change to the left lane, or change to the right lane)
(lines 203-234 in \cc{GeneticAlgorithm.py})
of \emph{two} hardcoded actor vehicles
(line 9 in \cc{drive_experiment.py}),
which always start driving at fixed positions (lines 180-181 in \cc{simulation.py}).
In contrast,
\sys explores a multifaceted input space
including
the mission, 
weather, 
locations, and trajectories
of an unbounded number of actor vehicles and/or pedestrians, and puddles.
%
%
%
Second,
\sys detects not only collisions (to vehicles, people, and objects),
but also safety-critical traffic violations (\eg, running red lights) with the driving test oracles.
However, \avf only considers vehicle-to-vehicle collisions
(lines 190-215 in \cc{simulation.py}),
which is a subset of the misbehaviors \sys detects.

\PP{Qualitative comparison}
In addition to the size of the input space
and the types of errors a fuzzer handles,
the quality of fuzzing feedback is tightly coupled with the quality of bugs
a fuzzer can detect.
As we discussed in \autoref{ss:targetedfeedback},
the feedback engine of \sys generates a fine-grained feedback of the recklessness
by referring to the physical vehicular states
that is highly relevant to the targeted misbehavior.
For example, when mutating the input scenario
that triggered bug \#29 in \autoref{tbl:bugs},
\sys placed puddles at the locations that decreased the driving quality the most
due to oversteering and hard acceleration,
and could eventually cause a misbehavior.
In the case of \avf,
it only favors scenarios in which the ego-vehicle
gets closer to the actor vehicles,
without considering the physical states of the \ads.
Unfortunately,
merely reducing the vehicular distance is not sufficient
to find scenarios (such as the one for bug \#29)
where no other vehicles are involved.

\subsection{Correctness of Driving Test Oracles}
\label{ss:eval-oracle}

The accuracy of misbehavior detection
depends on the correctness of the driving test oracles
\sys leverages.
We evaluate it
by injecting errors that cause the misbehavior that each oracle targets.
\autoref{tbl:oracle} shows
each misbehavior and corresponding errors
that are injected to synthesize scenarios
where each misbehavior must be observed.
For example, to test the collision oracle,
the input mutator is set to create a high-speed vehicle
driving directly towards the ego-vehicle
at 100 different locations.
After injecting each error,
we run \sys to check whether the intended misbehavior is detected or not
from each mutated scenario.
Except for the four rare false negatives
caused by a known issue in CARLA's lane invasion sensor~\cite{carla:laneinvasion},
the oracles never missed any misbehavior.

\subsection{Correctness of Driving Quality Metrics}
\label{ss:eval-quality}

\sys analyzes the vehicle states
to generate driving quality feedback
by detecting vehicular events. 
To ensure that \sys correctly implements the detection of each event,
we tested the feedback engine under a few synthesized experiments
that are designed to trigger the events.
Due to space constraints,
we show the correctness of detecting the two most complicated events:
understeer and oversteer,
which require correct implementation of fuzzy logic,
and present the figures in \autoref{s:qualityfig}.

\PP{Understeer experiment}
When understeer is triggered,
a vehicle cannot turn in the direction it desires,
as the frontal grip is lost.
The situation can be contrived
by placing a puddle at an intersection
where the vehicle has to make a turn
because the steering will not have any effect
on turning the vehicle once it starts slipping
(see \autoref{f:usfig}).
The feedback engine
successfully detected such events
as shown in \autoref{f:usgraph},
spotting the moments of understeer.

\PP{Oversteer experiment}
If a vehicle with a non-zero yaw speed
enters a section of a road with reduced friction,
tires easily lose grip and cause the vehicle to oversteer.
By synthesizing a scenario
where the ego-vehicle diagonally enters a puddle
as shown in \autoref{f:osfig},
we triggered oversteer
and tested the feedback engine.
As shown in \autoref{f:osgraph},
the feedback engine reliably detected the oversteers.


\subsection{Effectiveness of Driving Quality Feedback}
\label{ss:eval-feedback}

\begin{figure}[t!]
  \input{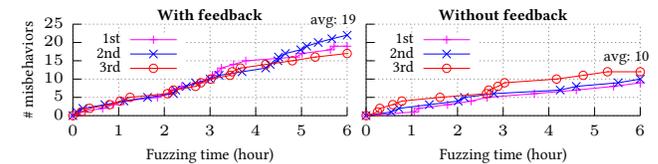}
  \caption{
    Number of misbehaviors observed while fuzzing Autoware
    for six hours with (left) and without (right)
    the driving quality feedback.
    Each configuration is repeated three times.
  }
  \label{f:feedback}
  \vspace*{-1em}
\end{figure}

By associating the likelihood of observing misbehaviors
with low quality (\eg, reckless or clumsy) driving,
the driving quality feedback
prevents the mutation engine from over-exploring
less interesting (\ie, hardly buggy) driving scenarios.
%
As a result,
it contributes to the effectiveness of \sys
in revealing more bugs within a given time frame.
%
%
To demonstrate this,
we run two configurations of fuzzers;
one with the driving quality feedback (\ie, the proposed setting)
and the other without the feedback
to fuzz Autoware starting with the same seed scenario.
%
%
%
%
As shown in \autoref{f:feedback},
\sys with the feedback found an average of 19 misbehaviors,
which are caused by bugs \#12--14 (\autoref{tbl:bugs}) in different situations
(note that the initial seed scenario was the one that revealed bug \#13).
Meanwhile,
without any guidance,
\sys blindly mutated scenarios and only discovered an average of 10 misbehaviors,
showing a significant decline (-47\%).
The result substantiates the design choice of \sys
that favoring the scenarios with lower driving quality
results in a better chance of finding bugs.

\subsection{Fuzzing Overhead}
\label{ss:eval-perf}

\begin{figure}[t!]
  \input{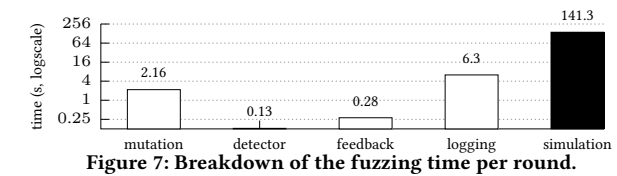}
  \vspace*{-1em}
  \caption{
    Breakdown of the fuzzing time per round.
  }
  \label{f:perf}
  \vspace*{-1em}
\end{figure}

%
The total duration of one fuzzing round varies significantly
depending on the length of a scenario
and the existence of the bug since buggy scenarios would terminate early.
In our experiments,
the average throughput of \sys was 150 seconds per end-to-end execution.
\autoref{f:perf} presents the breakdown
of the average time spent by each module per fuzzing execution.
The time required by
\sys-specific modules (white boxes) 
including mutation engine, misbehavior detector,
driving quality feedback engine, and logger,
only accounted for \textbf{6\%} of the total fuzzing time.
The mutation time
includes time spent for retries to ensure
semantically correct scenarios
(\autoref{sss:semmut}),
where the number of retries
ranged from zero to 2K times (in an extreme case)
with an average of 300 retries.

This is negligible compared to the simulation overhead (black box),
which dominates the overall fuzzing time
(\textbf{94\%}).
Thus,
employing a GPU with more computing power, or
parallelizing the simulations to multiple GPUs
can contribute to resolving the inevitable bottleneck.
Moreover,
although
the throughput of \sys may seem low
compared to traditional fuzzing approaches
that feature a high fuzzing speed,
the use of a driving simulator
enables \sys to
scale testing
with significantly lower cost
than physically testing autonomous vehicles
(\autoref{sss:simulator});
we could detect all \allbugs bugs
by running \sys for a week.



\begin{table}[t!]
  \centering
  \scriptsize
  \caption{Driving test oracles
  and the injected errors that trigger each misbehavior.
  100 different scenarios are created and tested for each error
  (\# TP: misbehavior was detected,
  \# FN: oracle missed the misbehavior).
  We manually confirmed that there was no false alarm. 
  }
  \vspace*{-0.5em}
  \input{tbl/tbl-oracle}
  \label{tbl:oracle}
  \vspace*{-1em}
\end{table}

%% file: casestudy.tex
\begin{figure}[t!]
  \centering
  \subfigure[Bug \#2]{
    \label{fig:case2}
    \includegraphics[clip,trim=1em 2em 1em 2em,width=0.16\textwidth]{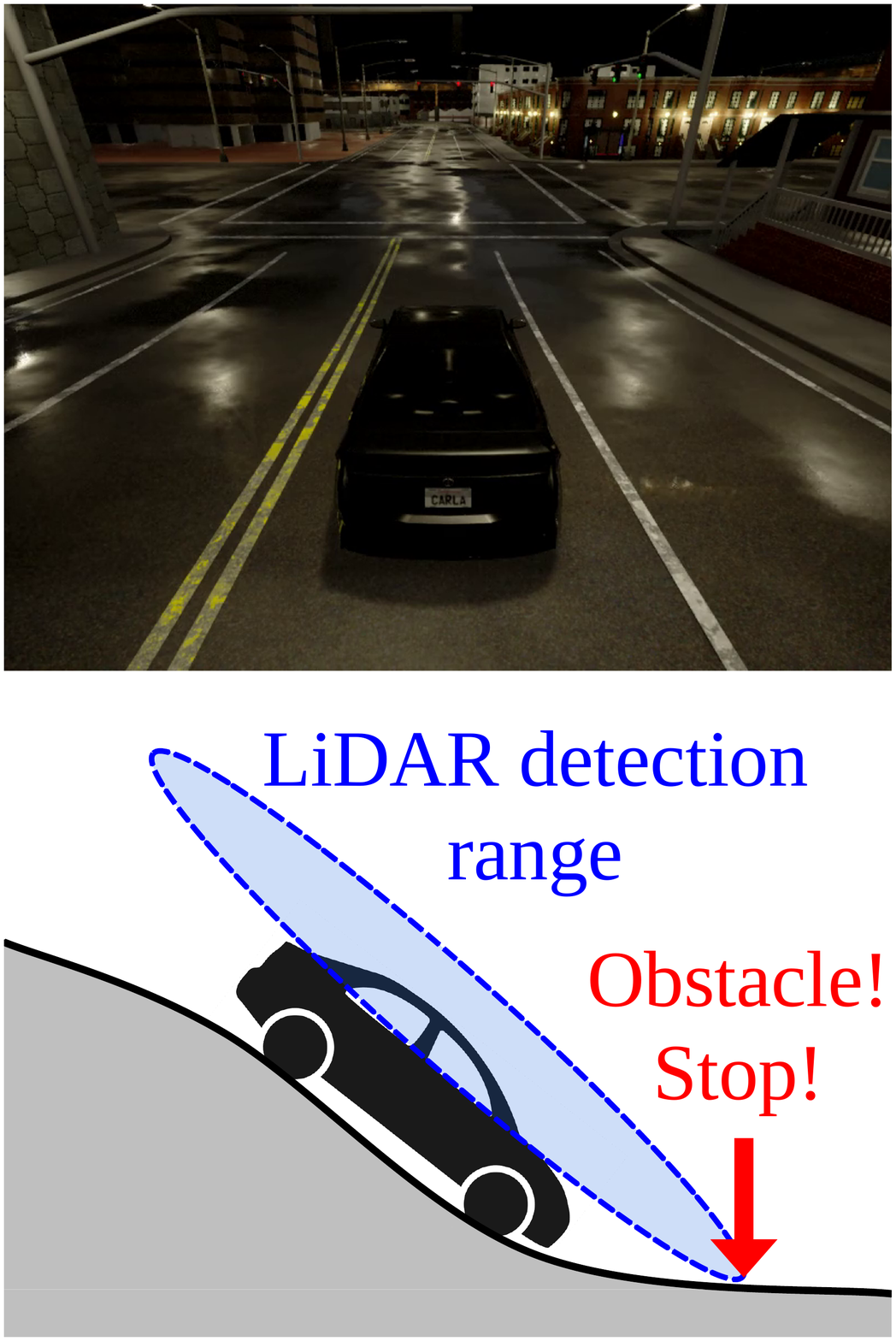}
  }
  \hspace{-1.6em}
  \subfigure[Bug \#7]{
    \label{fig:case7}
    \includegraphics[clip,trim=1em 2em 1em 2em,width=0.16\textwidth]{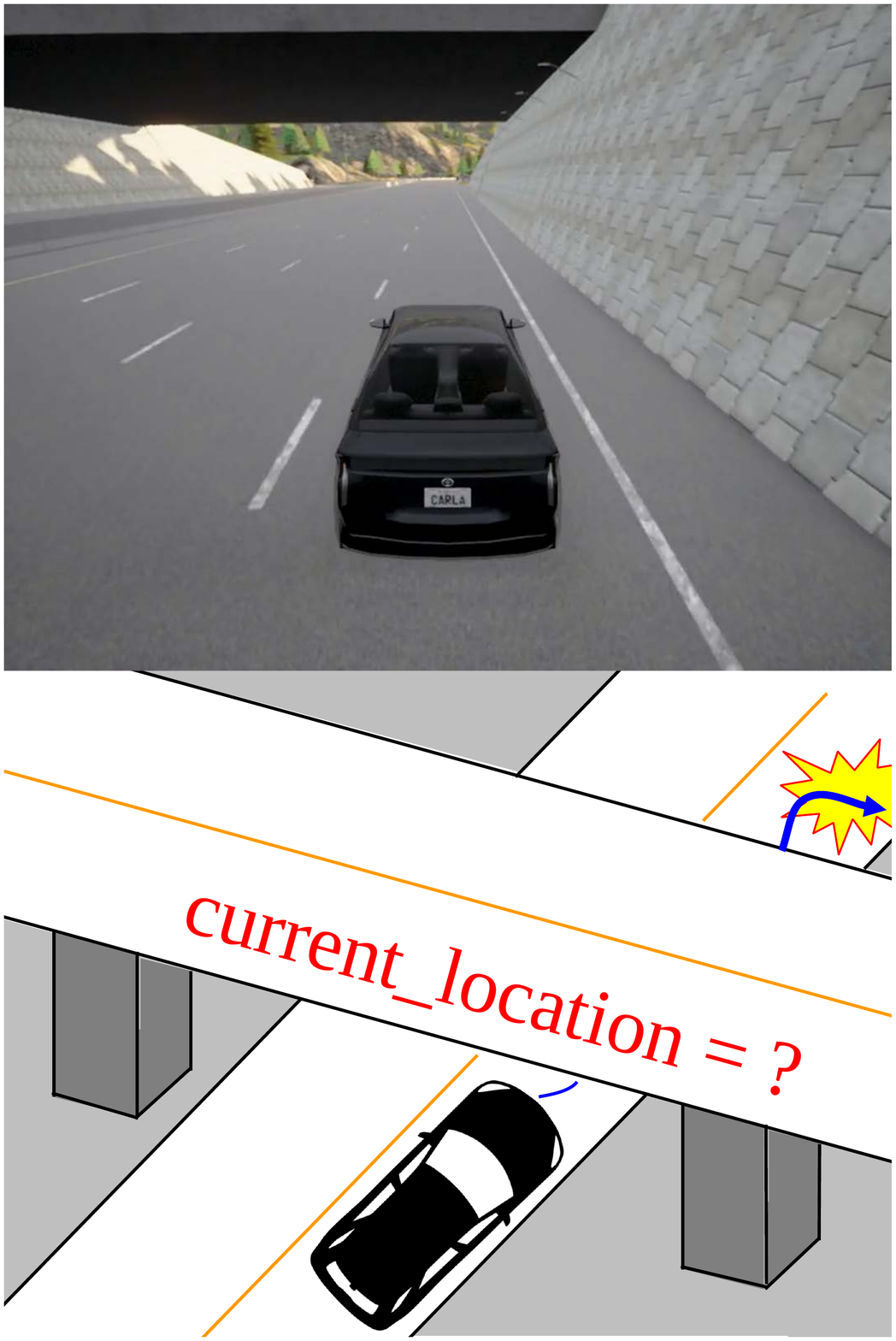}
  }
  \hspace{-1.6em}
  \subfigure[Bug \#12]{
    \label{fig:case12}
    \includegraphics[clip,trim=1em 2em 1em 2em,width=0.16\textwidth]{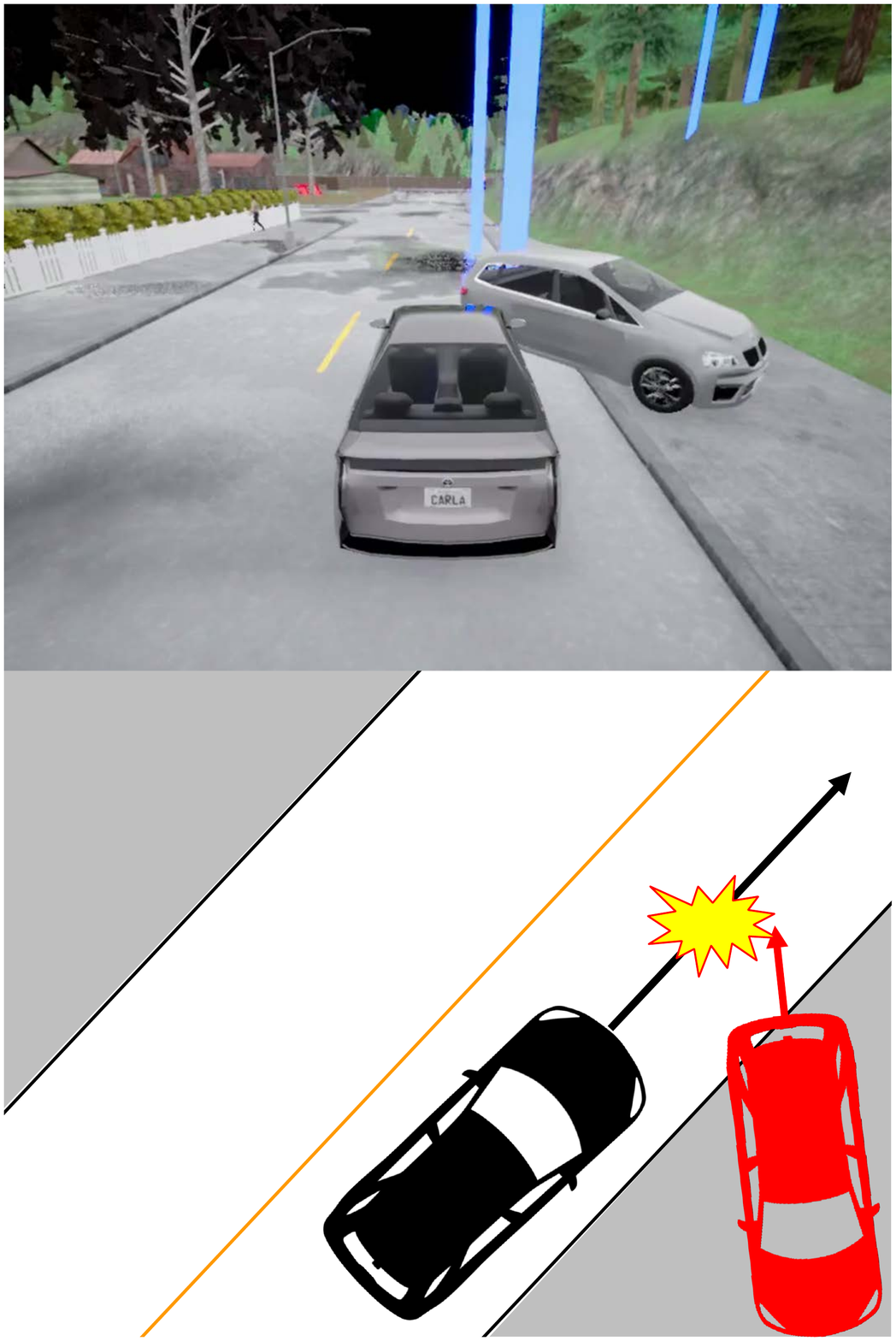}
  }
  \hspace{-1.6em}
  \subfigure[Bug \#15]{
    \label{fig:case15}
    \includegraphics[clip,trim=1em 2em 1em 2em,width=0.16\textwidth]{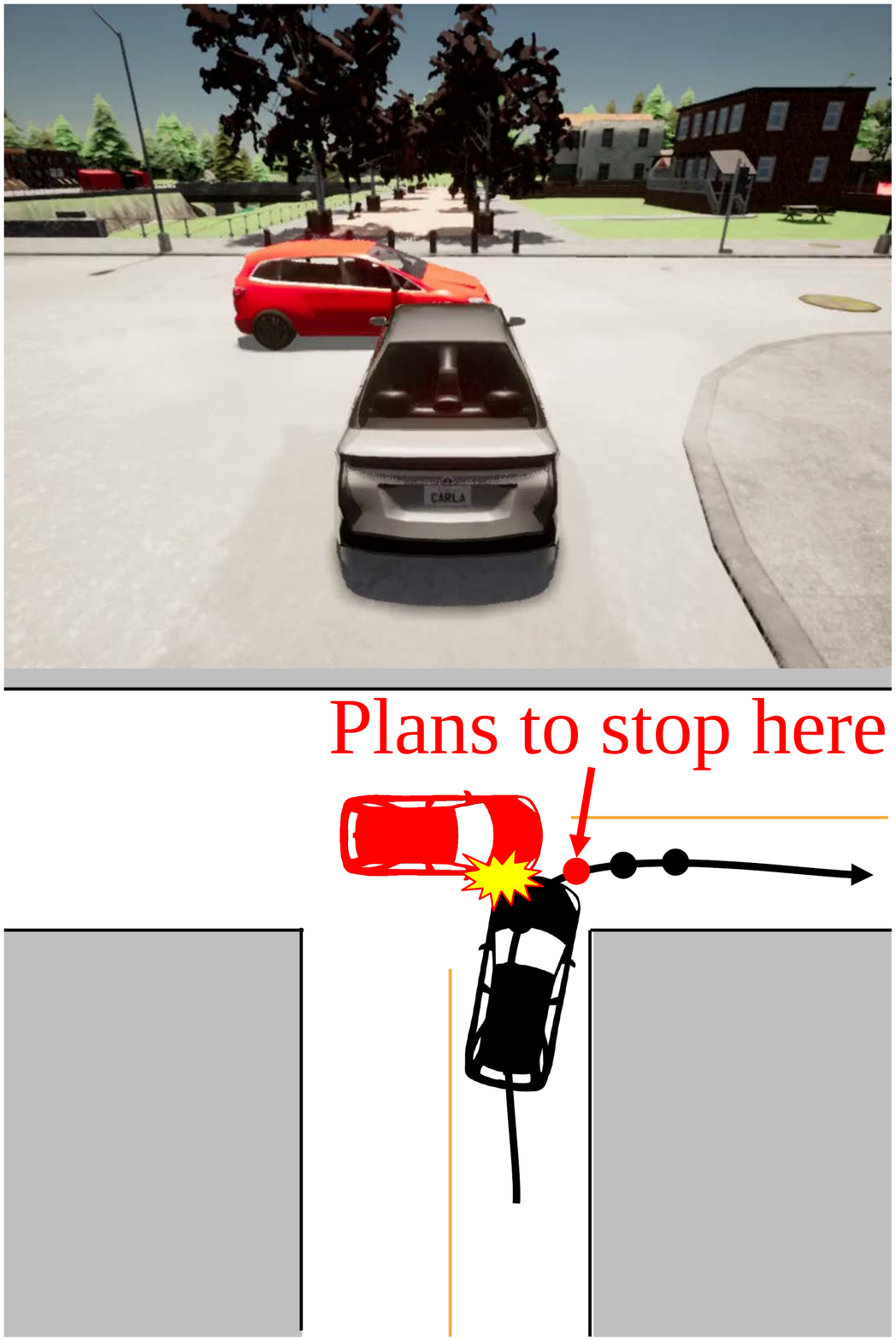}
  }
  \hspace{-1.6em}
  \subfigure[Bug \#17]{
    \label{fig:case17}
    \includegraphics[clip,trim=1em 2em 1em 2em,width=0.16\textwidth]{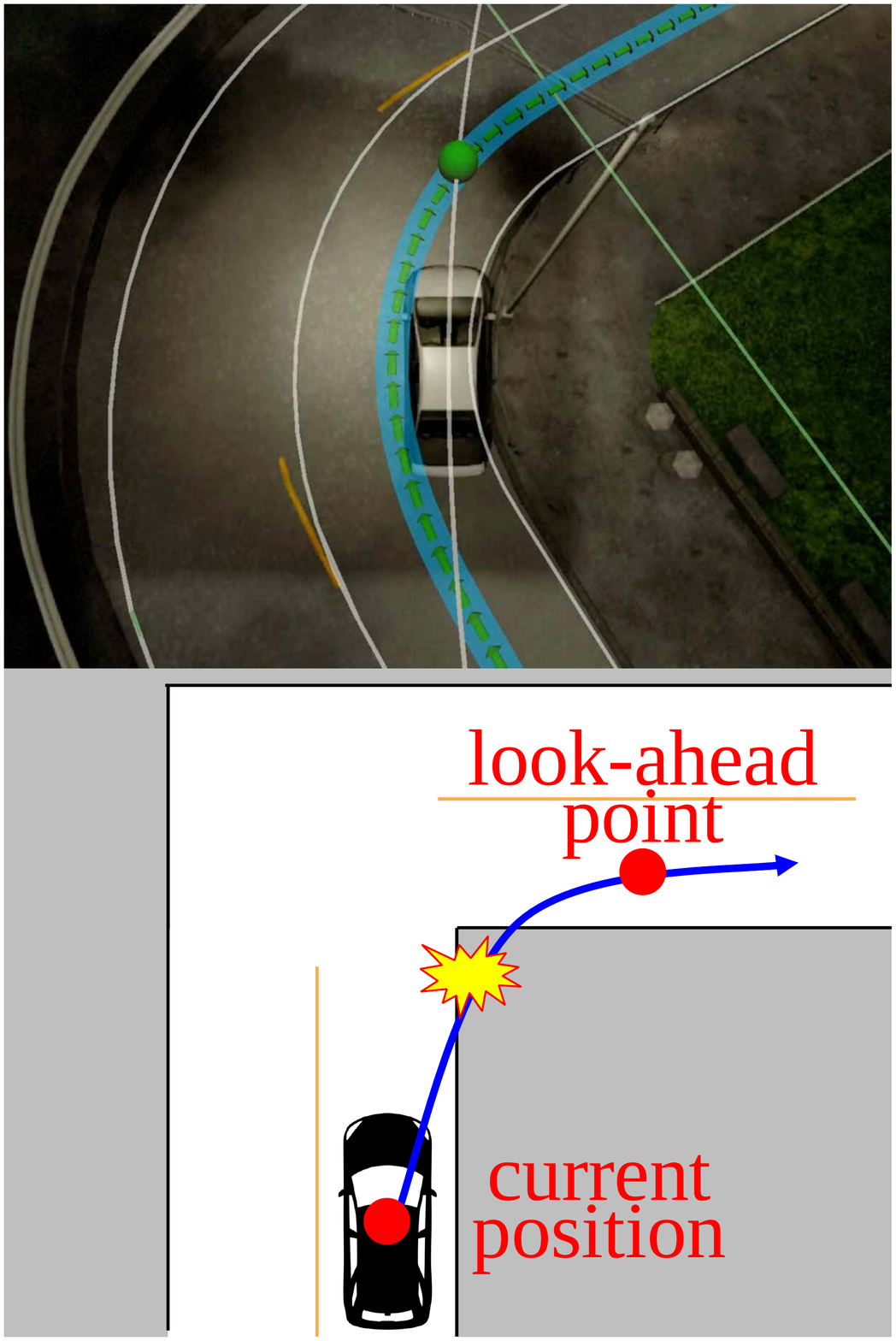}
  }
  \hspace{-1.6em}
  \subfigure[Bug \#25]{
    \label{fig:case25}
    \includegraphics[clip,trim=1em 2em 1em 2em,width=0.16\textwidth]{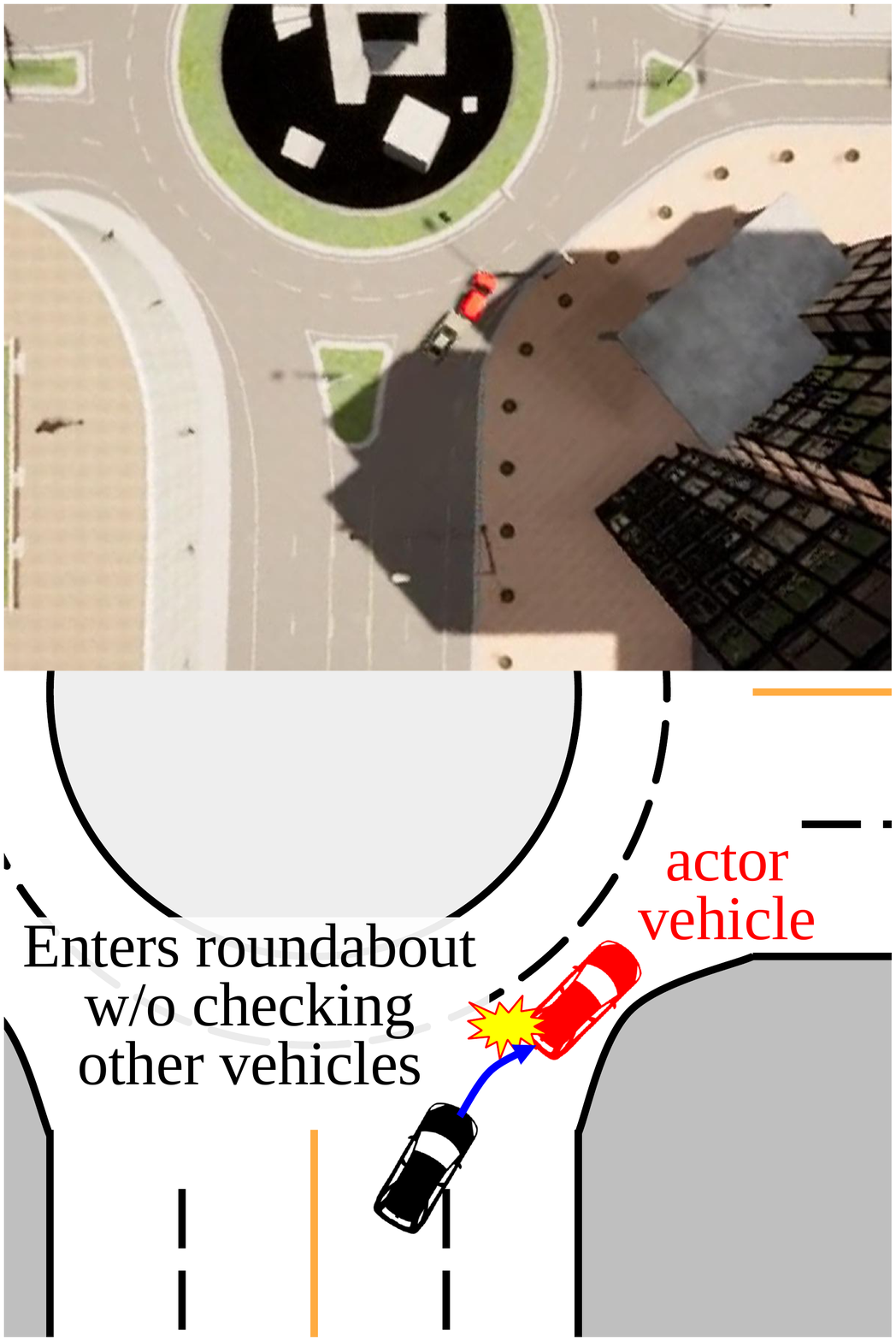}
  }
  \vspace{-0.5em}
  \caption{
    Notable cases of detected bugs.
    The images (top) show the snapshots
    of the camera feed at the moments when the bugs
    were detected.
    The diagrams (bottom) briefly describe the situation.
  }
  \label{f:cases}
  \vspace*{-2em}
\end{figure}

\subsection{Case Study}
\label{ss:casestudy}

We present an in-depth analysis of the selected bugs \sys found.
The bugs are categorized into four different types as follows.

\PP{Cross-layer bugs}
\sys identifies bugs that are caused by multiple layers,
requiring comprehensive testing of the \ads with all layers.
Cross-layer bugs are difficult to detect by testing
the individual layers because their symptoms often become
visible in a different layer from the buggy layer.

\noindent $\bullet$ Bug \#15 (see \autoref{fig:case15})
stems from two subtle problems in the perception and planning layers.
%
%
First, the perception layer measures the distance to the obstacle
from the center of the ego-vehicle, treating the vehicle as a point.
At this layer, this error does not necessarily cause a visible impact.
However, when the planner checks if the obstacle is blocking the path
based on this invalid distance,
it does not consider the dimension of the ego-vehicle, as well.
%
When the trajectory is not linear
(\eg, facing an obstacle while turning),
the dimension of the ego-vehicle is considered as zero by both layers,
making the edge of the bumper hit the obstacle.


%
\noindent $\bullet$ Bug \#17 (see \autoref{fig:case17})
is due to a faulty configuration in the actuation layer,
which is aggravated by faulty perception.
Autoware's controller smooths a trajectory
by constantly following the virtual curve
from the vehicle's position to the look-ahead point,
rather than strictly shooting for every point of a path.
%
%
The default configuration of the controller
sets the minimum look-ahead distance parameter too large,
which causes the ego-vehicle to cut through a curb
when it enters a sharp right curve at a low speed.
When this fault manifests,
the perception layer fails to identify the objects
on the unexpected trajectory,
and the vehicle ends up colliding with the static objects on the curb,
\eg, fences or street lights.


\PP{Logic errors}
A majority of the bugs \sys discovered
turned out to be logic errors,
where
the logic behind the implementation of a component
is the cause of misbehavior.

\noindent $\bullet$ Bug \#2 (see \autoref{fig:case2})
is caused by both sensor and perception layers;
at the end of a steep downhill
where the ego-vehicle faces down
while the road ahead flattens,
the LiDAR of Autoware senses the road ahead as an obstacle.
%
Without any verification of the point cloud data
published by the LiDAR,
the perception layer 
concludes that there is a massive object blocking the way,
and the local planner subsequently decides to stop immediately.
As the entire path is seemingly blocked,
the ego-vehicle becomes immobile thereafter.

\noindent $\bullet$ Bug \#7 (see \autoref{fig:case7})
is a critical bug that causes a localization error.
Autoware utilizes the Normal Distributions Transform (NDT) matching algorithm,
which estimates the current position of the ego-vehicle on the map
by combining the data from
the LiDAR, Inertial Measurement Unit,
Global Navigation Satellite System sensors,
and vehicle odometer data.
The localization plays a pivotal role
in the correctness of an \ads,
as all driving decisions are made
based on the estimated current position.
Unfortunately,
the NDT matching fails to correctly estimate the position
when the ego-vehicle is under a bridge,
presumably because its estimation
relies solely on the latitude and longitude, but not the altitude.

\noindent $\bullet$ Bug \#12 (see \autoref{fig:case12})
is notable as
it is directly related to the safety of passengers.
When a vehicle 
cuts in from either side to the front of the ego-vehicle,
the LiDAR sensor detects the vehicle,
and the perception layer perceives it as a vehicle.
However,
the local planner ignores the perceived vehicle
and fails to command a stop.


\noindent $\bullet$ Bug \#25 (see \autoref{fig:case25})
presents a devastating logic error in the planner of Behavior Agent.
The planner should slow down and stop if an obstacle is ahead.
However, as a part of optimization,
it only checks if anything is on the \textit{same lane} as the ego-vehicle.
As a result, when the ego-vehicle is switching lanes
or turning at an intersection/roundabout to enter another lane,
the planner fails to notice the obvious objects, causing collisions.



\PP{Missing features}
\sys found that
some of the misbehaviors
stem from not implementing essential features.

\noindent $\bullet$ Bugs \#13, 14, 26, and 27 demonstrate
that none of the components of Autoware and Behavior Agent
handles lateral and rear-end collision avoidance.
Even though the LiDAR sensor covers all 360 degrees
and \textit{perceives} approaching vehicles from all directions,
the local planner only considers
the objects lying in front of the vehicle
when revising the path plan,
\eg, taking a detour.
%
Thus,
when reckless vehicles approached the ego-vehicle
from behind or side in some scenarios,
the ego-vehicle did not try
to avoid them, (\eg, by accelerating or steering),
being subject to collisions.

\noindent $\bullet$ Bug \#29:
Electronic Stability Control (ESC)~\cite{liebemann2004safety}
is one of the essential and common in-vehicle safety features
that prevents and helps recover from oversteer and understeer
by automatically braking individual wheels and limiting engine powers.
Unfortunately, this essential feature is missing in Behavior Agent,
being vulnerable to bug \#29.
When the vehicle starts to slip due to a puddle,
the rotation of the wheels is not converted to vehicular speed.
Not considering the slipping state,
the controller keeps generating greater torques on the wheels
to achieve the target velocity,
which creates an excessive burst of acceleration
when the vehicle finally gets out of the puddle
and makes the vehicle lose control.

\PP{Simulation errors}
\sys also identifies errors within the simulator,
showing its end-to-end testing strategy's effectiveness.
%
Bugs \#31--33 manifested themselves
as one of the misbehaviors the detector examined
and later turned out to be the faults of the CARLA simulator
while debugging them.

\noindent $\bullet$ Bug \#31:
the ego-vehicle deviated from the planned path
while turning left at an intersection.
It did not turn as much as it was required
to follow the curved path,
but still throttled, and crashed into a building.
By analyzing the control commands Autoware issued,
we found that the ego-vehicle tried to steer
more and more towards the left
as it deviated from the path.
The culprit was CARLA,
which did not properly simulate the vehicle states
by applying the control commands it received from Autoware.
In real vehicles,
mechanical errors can cause similar problems
if it does not apply physical controls,
(\eg, steering),
as requested by the software stack.

\noindent $\bullet$ Bug \#32 is a data-related error.
In one of the CARLA maps,
the vector map
mistakenly listed a dead end
blocked by gas tanks
as a valid lane.
The lane was included in the path
found by the global path planner
in some scenarios,
and the ego-vehicle ended up getting stuck behind the gas tanks
blocking the path.
Similarly in the real world,
an autonomous vehicle could make inadequate path plans
if the ground truth data, such as a map, 
is not up to date.


%% file: tbl/tbl-oracle.tex
\footnotesize
\begin{tabular}{l l c c}
\toprule
\textbf{Misbehavior} &
\textbf{Injected error} &
  \textbf{\# TP} &
\textbf{\# FN} \\
\midrule
Collision & Have a vehicle rear-end the ego-vehicle & 100 & 0 \\
Speeding  & Set target speed to above limit & 100 & 0 \\
Running red lights & Disable traffic light detection & 100 & 0 \\
Immobility & Disable control module & 100 & 0 \\
Lane invasion & Force steer left & 96 & 4 \\

\bottomrule
\end{tabular}

%% file: data/feedback.tex
\begingroup
\scriptsize
  \makeatletter
  \providecommand\color[2][]{%
    \GenericError{(gnuplot) \space\space\space\@spaces}{%
      Package color not loaded in conjunction with
      terminal option `colourtext'%
    }{See the gnuplot documentation for explanation.%
    }{Either use 'blacktext' in gnuplot or load the package
      color.sty in LaTeX.}%
    \renewcommand\color[2][]{}%
  }%
  \providecommand\includegraphics[2][]{%
    \GenericError{(gnuplot) \space\space\space\@spaces}{%
      Package graphicx or graphics not loaded%
    }{See the gnuplot documentation for explanation.%
    }{The gnuplot epslatex terminal needs graphicx.sty or graphics.sty.}%
    \renewcommand\includegraphics[2][]{}%
  }%
  \providecommand\rotatebox[2]{#2}%
  \@ifundefined{ifGPcolor}{%
    \newif\ifGPcolor
    \GPcolortrue
  }{}%
  \@ifundefined{ifGPblacktext}{%
    \newif\ifGPblacktext
    \GPblacktextfalse
  }{}%
  \let\gplgaddtomacro\g@addto@macro
  \gdef\gplbacktext{}%
  \gdef\gplfronttext{}%
  \makeatother
  \ifGPblacktext
    \def\colorrgb#1{}%
    \def\colorgray#1{}%
  \else
    \ifGPcolor
      \def\colorrgb#1{\color[rgb]{#1}}%
      \def\colorgray#1{\color[gray]{#1}}%
      \expandafter\def\csname LTw\endcsname{\color{white}}%
      \expandafter\def\csname LTb\endcsname{\color{black}}%
      \expandafter\def\csname LTa\endcsname{\color{black}}%
      \expandafter\def\csname LT0\endcsname{\color[rgb]{1,0,0}}%
      \expandafter\def\csname LT1\endcsname{\color[rgb]{0,1,0}}%
      \expandafter\def\csname LT2\endcsname{\color[rgb]{0,0,1}}%
      \expandafter\def\csname LT3\endcsname{\color[rgb]{1,0,1}}%
      \expandafter\def\csname LT4\endcsname{\color[rgb]{0,1,1}}%
      \expandafter\def\csname LT5\endcsname{\color[rgb]{1,1,0}}%
      \expandafter\def\csname LT6\endcsname{\color[rgb]{0,0,0}}%
      \expandafter\def\csname LT7\endcsname{\color[rgb]{1,0.3,0}}%
      \expandafter\def\csname LT8\endcsname{\color[rgb]{0.5,0.5,0.5}}%
    \else
      \def\colorrgb#1{\color{black}}%
      \def\colorgray#1{\color[gray]{#1}}%
      \expandafter\def\csname LTw\endcsname{\color{white}}%
      \expandafter\def\csname LTb\endcsname{\color{black}}%
      \expandafter\def\csname LTa\endcsname{\color{black}}%
      \expandafter\def\csname LT0\endcsname{\color{black}}%
      \expandafter\def\csname LT1\endcsname{\color{black}}%
      \expandafter\def\csname LT2\endcsname{\color{black}}%
      \expandafter\def\csname LT3\endcsname{\color{black}}%
      \expandafter\def\csname LT4\endcsname{\color{black}}%
      \expandafter\def\csname LT5\endcsname{\color{black}}%
      \expandafter\def\csname LT6\endcsname{\color{black}}%
      \expandafter\def\csname LT7\endcsname{\color{black}}%
      \expandafter\def\csname LT8\endcsname{\color{black}}%
    \fi
  \fi
    \setlength{\unitlength}{0.0500bp}%
    \ifx\gptboxheight\undefined%
      \newlength{\gptboxheight}%
      \newlength{\gptboxwidth}%
      \newsavebox{\gptboxtext}%
    \fi%
    \setlength{\fboxrule}{0.5pt}%
    \setlength{\fboxsep}{1pt}%
\begin{picture}(4752.00,1152.00)%
    \gplgaddtomacro\gplbacktext{%
      \csname LTb\endcsname
      \put(355,345){\makebox(0,0)[r]{\strut{}$0$}}%
      \csname LTb\endcsname
      \put(355,483){\makebox(0,0)[r]{\strut{}$5$}}%
      \csname LTb\endcsname
      \put(355,621){\makebox(0,0)[r]{\strut{}$10$}}%
      \csname LTb\endcsname
      \put(355,759){\makebox(0,0)[r]{\strut{}$15$}}%
      \csname LTb\endcsname
      \put(355,897){\makebox(0,0)[r]{\strut{}$20$}}%
      \csname LTb\endcsname
      \put(355,1035){\makebox(0,0)[r]{\strut{}$25$}}%
      \csname LTb\endcsname
      \put(427,225){\makebox(0,0){\strut{}$0$}}%
      \csname LTb\endcsname
      \put(772,225){\makebox(0,0){\strut{}$1$}}%
      \csname LTb\endcsname
      \put(1116,225){\makebox(0,0){\strut{}$2$}}%
      \csname LTb\endcsname
      \put(1461,225){\makebox(0,0){\strut{}$3$}}%
      \csname LTb\endcsname
      \put(1805,225){\makebox(0,0){\strut{}$4$}}%
      \csname LTb\endcsname
      \put(2150,225){\makebox(0,0){\strut{}$5$}}%
      \csname LTb\endcsname
      \put(2494,225){\makebox(0,0){\strut{}$6$}}%
      \put(2218,1063){\makebox(0,0)[l]{\strut{}avg: 19}}%
    }%
    \gplgaddtomacro\gplfronttext{%
      \csname LTb\endcsname
      \put(97,690){\rotatebox{-270}{\makebox(0,0){\strut{}\# misbehaviors}}}%
      \put(1460,45){\makebox(0,0){\strut{}Fuzzing time (hour)}}%
      \put(1460,1095){\makebox(0,0){\strut{}\textbf{With feedback}}}%
      \csname LTb\endcsname
      \put(787,918){\makebox(0,0)[r]{\strut{}1st}}%
      \csname LTb\endcsname
      \put(787,810){\makebox(0,0)[r]{\strut{}2nd}}%
      \csname LTb\endcsname
      \put(787,702){\makebox(0,0)[r]{\strut{}3rd}}%
    }%
    \gplgaddtomacro\gplbacktext{%
      \csname LTb\endcsname
      \put(2565,345){\makebox(0,0)[r]{\strut{}}}%
      \csname LTb\endcsname
      \put(2565,483){\makebox(0,0)[r]{\strut{}}}%
      \csname LTb\endcsname
      \put(2565,621){\makebox(0,0)[r]{\strut{}}}%
      \csname LTb\endcsname
      \put(2565,759){\makebox(0,0)[r]{\strut{}}}%
      \csname LTb\endcsname
      \put(2565,897){\makebox(0,0)[r]{\strut{}}}%
      \csname LTb\endcsname
      \put(2565,1035){\makebox(0,0)[r]{\strut{}}}%
      \csname LTb\endcsname
      \put(2637,225){\makebox(0,0){\strut{}$0$}}%
      \csname LTb\endcsname
      \put(2981,225){\makebox(0,0){\strut{}$1$}}%
      \csname LTb\endcsname
      \put(3326,225){\makebox(0,0){\strut{}$2$}}%
      \csname LTb\endcsname
      \put(3670,225){\makebox(0,0){\strut{}$3$}}%
      \csname LTb\endcsname
      \put(4014,225){\makebox(0,0){\strut{}$4$}}%
      \csname LTb\endcsname
      \put(4359,225){\makebox(0,0){\strut{}$5$}}%
      \csname LTb\endcsname
      \put(4703,225){\makebox(0,0){\strut{}$6$}}%
      \put(4428,787){\makebox(0,0)[l]{\strut{}avg: 10}}%
    }%
    \gplgaddtomacro\gplfronttext{%
      \csname LTb\endcsname
      \put(2511,690){\rotatebox{-270}{\makebox(0,0){\strut{}}}}%
      \put(3670,45){\makebox(0,0){\strut{}Fuzzing time (hour)}}%
      \put(3670,1095){\makebox(0,0){\strut{}\textbf{Without feedback}}}%
      \csname LTb\endcsname
      \put(2997,918){\makebox(0,0)[r]{\strut{}1st}}%
      \csname LTb\endcsname
      \put(2997,810){\makebox(0,0)[r]{\strut{}2nd}}%
      \csname LTb\endcsname
      \put(2997,702){\makebox(0,0)[r]{\strut{}3rd}}%
    }%
    \gplbacktext
    \put(0,0){\includegraphics{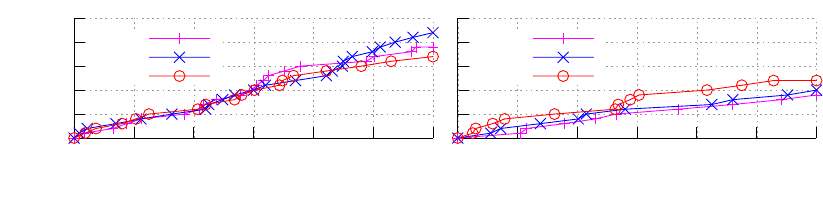}}%
    \gplfronttext
  \end{picture}%
\endgroup

%% file: data/perf.tex
\begingroup
\scriptsize
  \makeatletter
  \providecommand\color[2][]{%
    \GenericError{(gnuplot) \space\space\space\@spaces}{%
      Package color not loaded in conjunction with
      terminal option `colourtext'%
    }{See the gnuplot documentation for explanation.%
    }{Either use 'blacktext' in gnuplot or load the package
      color.sty in LaTeX.}%
    \renewcommand\color[2][]{}%
  }%
  \providecommand\includegraphics[2][]{%
    \GenericError{(gnuplot) \space\space\space\@spaces}{%
      Package graphicx or graphics not loaded%
    }{See the gnuplot documentation for explanation.%
    }{The gnuplot epslatex terminal needs graphicx.sty or graphics.sty.}%
    \renewcommand\includegraphics[2][]{}%
  }%
  \providecommand\rotatebox[2]{#2}%
  \@ifundefined{ifGPcolor}{%
    \newif\ifGPcolor
    \GPcolortrue
  }{}%
  \@ifundefined{ifGPblacktext}{%
    \newif\ifGPblacktext
    \GPblacktextfalse
  }{}%
  \let\gplgaddtomacro\g@addto@macro
  \gdef\gplbacktext{}%
  \gdef\gplfronttext{}%
  \makeatother
  \ifGPblacktext
    \def\colorrgb#1{}%
    \def\colorgray#1{}%
  \else
    \ifGPcolor
      \def\colorrgb#1{\color[rgb]{#1}}%
      \def\colorgray#1{\color[gray]{#1}}%
      \expandafter\def\csname LTw\endcsname{\color{white}}%
      \expandafter\def\csname LTb\endcsname{\color{black}}%
      \expandafter\def\csname LTa\endcsname{\color{black}}%
      \expandafter\def\csname LT0\endcsname{\color[rgb]{1,0,0}}%
      \expandafter\def\csname LT1\endcsname{\color[rgb]{0,1,0}}%
      \expandafter\def\csname LT2\endcsname{\color[rgb]{0,0,1}}%
      \expandafter\def\csname LT3\endcsname{\color[rgb]{1,0,1}}%
      \expandafter\def\csname LT4\endcsname{\color[rgb]{0,1,1}}%
      \expandafter\def\csname LT5\endcsname{\color[rgb]{1,1,0}}%
      \expandafter\def\csname LT6\endcsname{\color[rgb]{0,0,0}}%
      \expandafter\def\csname LT7\endcsname{\color[rgb]{1,0.3,0}}%
      \expandafter\def\csname LT8\endcsname{\color[rgb]{0.5,0.5,0.5}}%
    \else
      \def\colorrgb#1{\color{black}}%
      \def\colorgray#1{\color[gray]{#1}}%
      \expandafter\def\csname LTw\endcsname{\color{white}}%
      \expandafter\def\csname LTb\endcsname{\color{black}}%
      \expandafter\def\csname LTa\endcsname{\color{black}}%
      \expandafter\def\csname LT0\endcsname{\color{black}}%
      \expandafter\def\csname LT1\endcsname{\color{black}}%
      \expandafter\def\csname LT2\endcsname{\color{black}}%
      \expandafter\def\csname LT3\endcsname{\color{black}}%
      \expandafter\def\csname LT4\endcsname{\color{black}}%
      \expandafter\def\csname LT5\endcsname{\color{black}}%
      \expandafter\def\csname LT6\endcsname{\color{black}}%
      \expandafter\def\csname LT7\endcsname{\color{black}}%
      \expandafter\def\csname LT8\endcsname{\color{black}}%
    \fi
  \fi
    \setlength{\unitlength}{0.0500bp}%
    \ifx\gptboxheight\undefined%
      \newlength{\gptboxheight}%
      \newlength{\gptboxwidth}%
      \newsavebox{\gptboxtext}%
    \fi%
    \setlength{\fboxrule}{0.5pt}%
    \setlength{\fboxsep}{1pt}%
\begin{picture}(4608.00,1152.00)%
    \gplgaddtomacro\gplbacktext{%
      \csname LTb\endcsname
      \put(516,312){\makebox(0,0)[r]{\strut{}$0.25$}}%
      \csname LTb\endcsname
      \put(516,456){\makebox(0,0)[r]{\strut{}$1$}}%
      \csname LTb\endcsname
      \put(516,600){\makebox(0,0)[r]{\strut{}$4$}}%
      \csname LTb\endcsname
      \put(516,743){\makebox(0,0)[r]{\strut{}$16$}}%
      \csname LTb\endcsname
      \put(516,887){\makebox(0,0)[r]{\strut{}$64$}}%
      \csname LTb\endcsname
      \put(516,1031){\makebox(0,0)[r]{\strut{}$256$}}%
      \put(987,120){\makebox(0,0){\strut{}mutation}}%
      \put(1785,120){\makebox(0,0){\strut{}detector}}%
      \put(2584,120){\makebox(0,0){\strut{}feedback}}%
      \put(3382,120){\makebox(0,0){\strut{}logging}}%
      \put(4180,120){\makebox(0,0){\strut{}simulation}}%
    }%
    \gplgaddtomacro\gplfronttext{%
      \csname LTb\endcsname
      \put(114,635){\rotatebox{-270}{\makebox(0,0){\strut{}time (s, logscale)}}}%
      \put(987,656){\makebox(0,0){\strut{}2.16}}%
      \put(1785,364){\makebox(0,0){\strut{}0.13}}%
      \put(2584,444){\makebox(0,0){\strut{}0.28}}%
      \put(3382,767){\makebox(0,0){\strut{}6.3}}%
      \put(4180,1089){\makebox(0,0){\strut{}141.3}}%
    }%
    \gplbacktext
    \put(0,0){\includegraphics{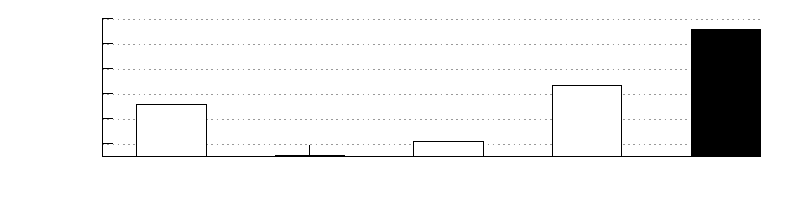}}%
    \gplfronttext
  \end{picture}%
\endgroup

%% file: relwk.tex
\section{Related Work}
\label{s:relwk}

\PP{Testing autonomous driving systems}
%
%
Most existing approaches 
focus on the white-box testing of individual layers:
sensing~\cite{jokela2019testing, geiger2012we, broggi2013extensive},
perception ~\cite{pei2017deepxplore, tian2018deeptest, zhang2018deeproad}, and
planning~\cite{ohta2016pure, calo2020generating, ndss:2022:ziwen:planfuzz}.
For example,
%
%
the series of works on the perception layer~\cite{
pei2017deepxplore, tian2018deeptest, zhang2018deeproad}
tests the robustness of the neural network model
with synthetically transformed camera images based on the model's
activation patterns.
%
%
PlanFuzz~\cite{ndss:2022:ziwen:planfuzz}
tries to find denial of service vulnerabilities
in the planning layer
by introducing physical objects into driving scenes
and guiding the input scenario generation based on the code execution
that it monitors through instrumentation.
%
%
%
%
Unlike these works, 
\sys considers a target \ads as a whole system
rather than focusing on a specific layer or problem.
This holistic approach allows us to find
not only those bugs that individual layer testing
covers, but also other types of bugs
with \emph{propagating} impacts across multiple layers
that cause critical accidents.
%
%
Besides, our approach does not require the source code,
instrumentation, or domain knowledge of the target \ads
in contrast to those white-box approaches.

A few testing works take a holistic approach similar to
ours~\cite{li2020av, han2020metamorphic}.
The closest to our work is \avf~\cite{li2020av}.
\avf mutates the trajectory of nearby vehicles
with an objective to find scenarios where the ego-vehicle gets too
close to them.
Although conceptually similar,
its input dimension and the scope of safety violations
are a small subset of what \sys considers,
as evaluated in \autoref{ss:comp-avfuzzer}.
Han \etal~\cite{han2020metamorphic} propose an adversarial
testing approach,
which tests autonomous vehicles under rather unrealistic test cases
(\eg, a static obstacle suddenly appearing and disappearing).
This approach
does not necessarily focus on
the feasibility of exploiting the bugs
from the attacker's perspective.
In contrast,
our approach focuses on generating semantically valid test cases
that attackers can exploit.

Fremont \etal~\cite{fremont2020formal}
tackle the testing problem from a different
but complementary angle
by applying a formal methods-based approach.
They focus on generating test cases that will run on a real track
based on the formal verification of driving scenarios,
rather than finding bugs in \adses.


\PP{Adversarial example attacks}
Many existing works focus on finding
adversarial attacks that deceive the machine learning model
of the perception layer 
~\cite{
sun2020towards, jing122021too, shen2020drift,
nassi2020phantom, cao2019adversarial, song2018adversarial,
boloor2020attacking, chernikova2019self}.
These attacks input sensor data with carefully crafted perturbations
to cause misclassification,
such as camera images with a modified traffic sign,
or spoofed LiDAR data.
Similar to the testing approaches on individual layers,
these works target a specific layer and problem;
\ie, the lack of robustness of
machine learning model in the perception layer.
Complementary to these works,
the goal of \sys is finding vulnerabilities
in any layer of an \ads regardless of their location.

\PP{Coverage-guided fuzzing}
Many existing fuzzers are geared towards
improving bug detection abilities across various domains. 
In previous studies,
some focus on
improving the code coverage feedback~\cite{web:syzkaller, web:libFuzzer, zalewski2014american},
while others
retrieve more advanced information (\eg, data flow)
to handle systems in new domains or platforms
(\eg, drone control)~\cite{sialveras2015introducing,
  rawat2017vuzzer, osterlund2020parmesan,
  chen2018iotfuzzer, fiterau2020analysis, kim2019touching, karim2020atfuzzer,
  dinh2021favocado, chen2019ptrix, kim2019rvfuzzer, kim2019finding}. 
%
%
Unfortunately,
none of these approaches can be directly applied to \adses
as they are designed to find
typical software bugs only (\eg, memory safety violation),
relying on obvious symptoms of program failures (\eg, segmentation faults)
and general code coverage to guide the input mutation.
%
To address this limitation,
\sys is designed specifically for holistically fuzzing \adses
leveraging new test oracles and quality metrics that focus on
driving semantics and vehicle states.



%
%
%
%

%% file: discussion.tex
\section{Discussion and Future Work}
\label{s:discussion}

\PP{Fidelity of simulation}
Despite a potential gap between the simulated
and real environments,
the use of high-fidelity simulation
brings the quality of test cases in close
proximity to that of physical testing
and significantly enhances the quality of
automated \ads testing over existing methods.
This is also demonstrated by the fact that \sys discovers
\newbugs new \ads and simulator
bugs in the corner case driving scenarios
that existing testing methods could not attempt to
generate.
%
\ackbugs (out of \newbugs reported) bugs have been acknowledged by the
\ads developers,
and most are readily exploitable with concrete attacks
by an adversary as we demonstrate in \autoref{ss:eval-repro}.
More importantly,
it not only enables a full degree of automation,
but also provides other practical benefits,
such as low cost and safety of testing, 
in comparison with physical testing with real vehicles.
It is also supported by the fact that
major \ads vendors
rely heavily on simulators to develop and
test their systems before physical testing~\cite{
  waymo:simulation, huang2016autonomous}.
In our future work,
we plan to reproduce the \ads issues in this
paper with a real autonomous vehicle. 



\PP{Definition of good/bad behaviors}
\label{pp:judgingmisbehaviors}
Defining good and bad behaviors is challenging
as it is a subjective matter that depends on the circumstance and the intent of the behaviors.
For example,
Crossing a yellow line at a two-lane expressway is considered an infraction,
while it is circumstantially benign
if it is to avoid a collision with an object, \eg, a vehicle blocking the road.
In light of this,
we made the misbehavior oracles configurable
so that they can be adjusted per target.
In addition,
when misbehaviors are detected,
\sys generates detailed reports
with all sensor data including the camera feed,
so that users can further reconfigure and fine-tune the oracles.
%

\PP{Extensibility of DriveFuzz}
\label{pp:eval-extensibility}
\sys is designed with an extensibility in mind;
the mutation engine, 
misbehavior detector, 
and driving quality feedback engine 
are \emph{generic},
operating independently of the \ads under test.
In addition,
the test executor, 
which bridges the \ads with the simulator and \sys,
supports ROS to maximize the compatibility with
the ROS-based systems.
This is showcased by testing a ROS-based system,
Autoware in \autoref{s:eval}.

\PP{Limitation}
Our driving quality-based feedback
directs \sys to scenarios where an \ads performs unsafe maneuvers.
We have proven that
such feedback is effective in triggering misbehavior that we target.
However,
similar to most feedback-driven fuzzers
that register a specific fitness function as a feedback,
\sys can have a local optima problem~\cite{manes2020ankou},
\ie, reaching a local optimum in the search space as a result of feedback guidance,
and ends up missing other potential bugs
that are less related to the feedback.

In addition,
there may exist attacks that do not affect the driving quality score
but still cause misbehaviors.
For example, if an adversary draws a fake curved lane on a straight road
and misleads an \ads to invade a sidewalk,
the driving quality score can still be good
if an ego-vehicle seamlessly follows the fake lane,
while the resulting circumstance is a lane invasion.
As a mitigation,
we can extend and diversify the driving quality score metrics,
\eg, considering the adherence to the original plan,
to deal with the bugs that are not necessarily
coupled with clumsy driving behaviors.

%% file: conclusion.tex
\section{Conclusion}
\label{s:conclusion}

This paper presents \sys,
an end-to-end fuzzing framework
designed to find bugs in all layers of \adses
that are readily exploitable by attackers. 
%
\sys detects bugs by
(1) automatically generating and mutating high-fidelity driving scenarios,
(2) checking for safety-critical misbehaviors
using driving test oracles contrived by
studying real-world traffic rules,
and
(3) measuring our novel driving quality score
by inspecting the vehicle states
and using it as feedback to guide the mutation engine
towards buggy scenarios.
\sys has found \anewbugs new bugs in Autoware, 
\ballbugs bugs in Behavior Agent, 
and \cnewbugse bugs in the simulator, 
showing that it can discover bugs in all layers of the tested system.
Our study shows that the bugs we found
can be triggered by only controlling legitimate inputs
and cause devastating vehicle accidents.

%% file: ack.tex
\section*{Acknowledgment}
\label{s:ack}

We thank the anonymous reviewers,
and our shepherd, Ziming Zhao, for their insightful feedback.
This work was supported in part by
the University of Texas at Dallas Office of Research through the NFRS program,
Texas A\&M Engineering Experiment Station
on behalf of its SecureAmerica Institute, Institute of Information \& Communications Technology Planning \& Evaluation (IITP) grant funded by the Korea government (MSIT) (No.2022-0-00745, The Development of Ransomware Attack Source Identification and Analysis Technology),
and a gift from Cisco Systems.
%

%% file: appendix.tex
\appendix

\section{Seed generation and verification}
\label{s:seedgen}
We present our seed generation and verification process in \autoref{algo:seedgen},
which can be performed without substantial domain knowledge of a particular \ads.

\PP{Seed construction.}
Our seed scenarios are created on top of the maps CARLA provides
(\url{https://carla.readthedocs.io/en/0.9.10/core_map/}),
which have various road components (\eg, lanes, junctions, traffic lights)
along with static objects such as trees and buildings.
Specifically,
we first selected five common components of the road system;
urban street, highway, interchange, intersection, and roundabout.
%
For each of the five road components,
we select $n_s = 8$ missions,
in which the mission (\autoref{algo:line:mission})
consisting of the initial position ($p_i$) and the goal position ($p_g$)
requires either driving within or around the component.
Other than having the pre-selected map and mission assigned,
each seed driving scenario is in a clean slate,
having no other actors (\autoref{algo:line:actor})
or puddles (\autoref{algo:line:puddle})
with sunny weather (\autoref{algo:line:weather}).

\PP{Checking the validity of seeds.}
The legitimacy of all 40 seeds is verified (\autoref{algo:line:verify})
prior to fuzzing
by dry-running the target \ads with the seed scenario (\autoref{algo:line:dryrun})
and confirming that it successfully completes the mission (\autoref{algo:line:success}).

\begin{figure}[h!]
  \centering
  \vspace*{-0.5em}
  \subfigure[Example seed scenario at urban streets]{
    \label{fig:seed1}
    \includegraphics[width=0.35\textwidth]{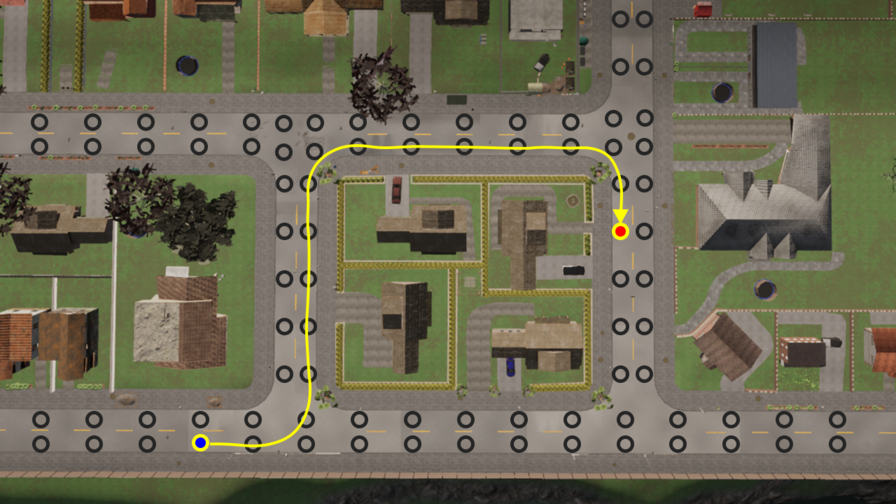}
  }
  \vspace*{-0.5em}
  \subfigure[Example seed scenario at a highway]{
    \label{fig:seed2}
    \includegraphics[width=0.35\textwidth]{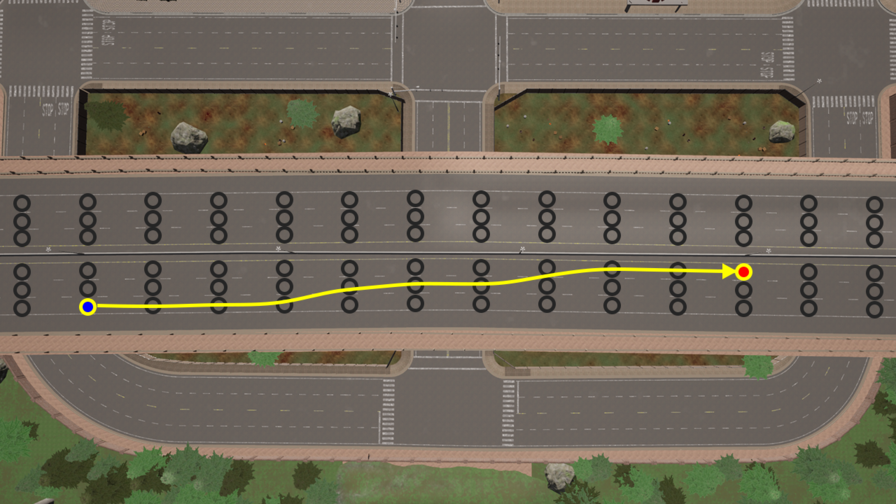}
  }
  \vspace*{-0.5em}
  \subfigure[Example seed scenario at a cloverleaf interchange]{
    \label{fig:seed3}
    \includegraphics[width=0.35\textwidth]{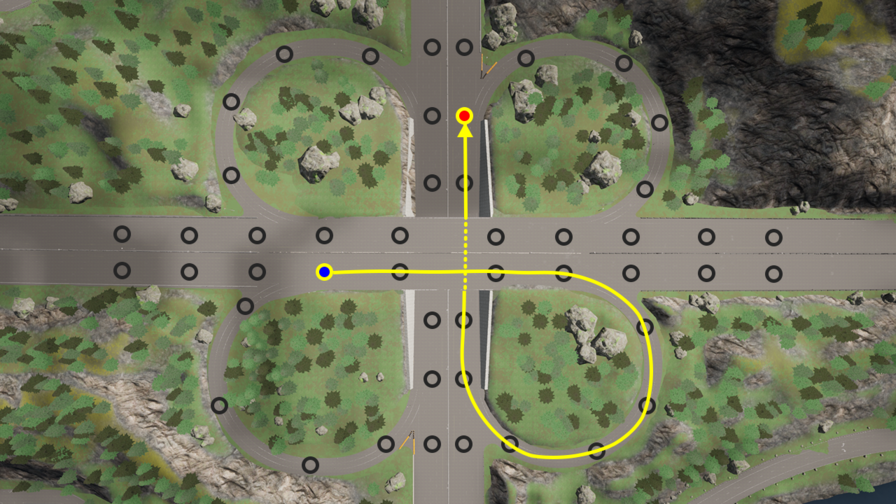}
  }
  \vspace*{-0.5em}
  \subfigure[Example seed scenario at an intersection]{
    \label{fig:seed4}
    \includegraphics[width=0.35\textwidth]{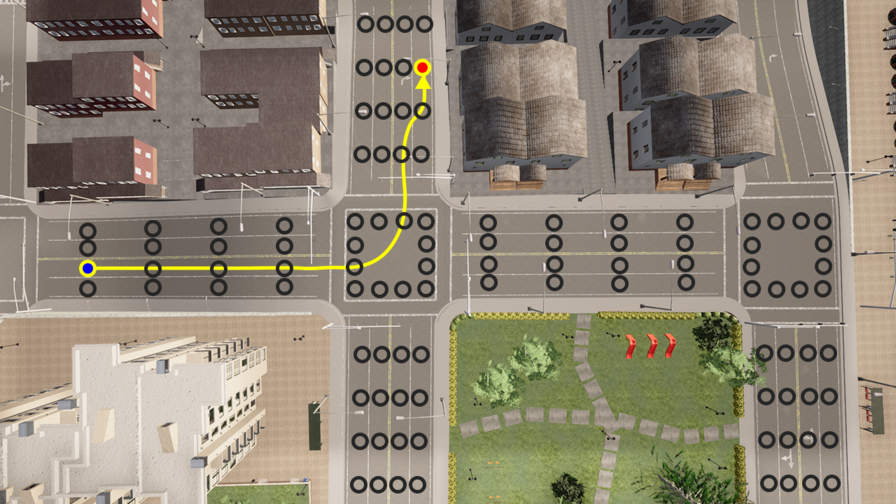}
  }
  \vspace*{-0.5em}
  \subfigure[Example seed scenario at a roundabout]{
    \label{fig:seed5}
    \includegraphics[width=0.35\textwidth]{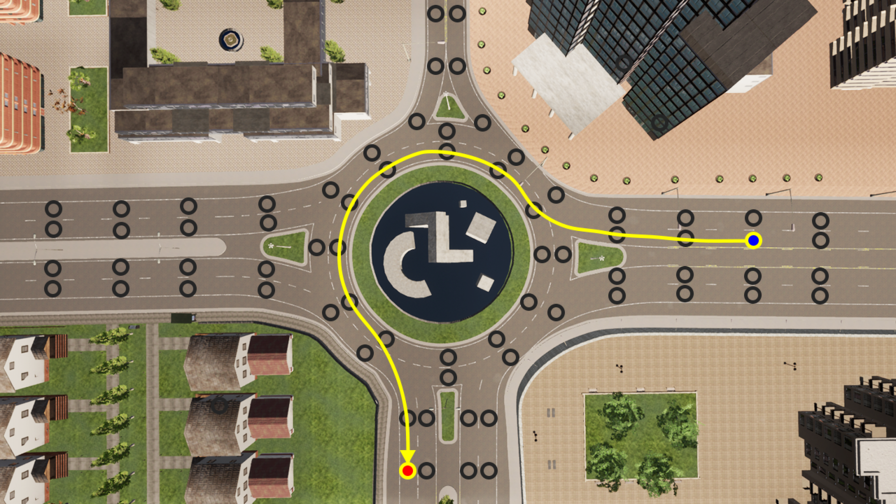}
  }
  \caption{
    Example seed scenario at each road component.
  }
  \label{f:seed}
\end{figure}

\PP{Examples.}
\autoref{f:seed} shows concrete examples of the seed scenarios
generated and verified through the aforementioned procedure.
The black circles indicate the predefined valid waypoints in the CARLA map
(please note that the number and density of the waypoints shown
are greatly reduced for an illustration purpose).
A random waypoint can be retrieved by
\cc{get_random_position_near()} function in \autoref{algo:seedgen}.
The yellow arrow connecting the blue (initial position, $p_i$)
with the red circles (goal position, $p_g$)
indicates the mission assigned to each seed scenario.
%

\begin{figure}[h!]
\removelatexerror
\begin{algorithm}[H]
  \scriptsize
  \caption{Seed generation and verification}
  \label{algo:seedgen}
  \SetKwBlock{Begin}{procedure}{end procedure}
  \SetKwInOut{Input}{Input}
  \SetKwInOut{Output}{Output}

  \Input{$C$ - a set of road components $= \{urban, highway, interchange, intersection, roundabout\}$,\\
  $n_s$ - \# scenarios per road component}
  \Output{$S$ - a set of seed scenarios}
  
  \Begin($generate\_seed{(}{)}$) {
    \ForEach{$comp \in C$}{
      \For{$i \gets 1$ to $n_s$}{
        \While{$true$}{
          $p_i \gets 0$ // initial position \\
          $p_g \gets 0$ // goal position \\
          \While{$p_i == p_g$}{
            $p_i \gets get\_random\_position\_near(comp)$ \\
            $p_g \gets get\_random\_position\_near(comp)$ \\
          }
          $mission \gets \{p_i, p_g\}$ \label{algo:line:mission} \\
          $actors \gets \varnothing$  \label{algo:line:actor}\\
          $puddles \gets \varnothing$ \label{algo:line:puddle} \\
          $weather \gets sunny$ \label{algo:line:weather} \\
          $seed \gets init\_scenario(mission, actors, puddles, weather)$ \\
          
          \If{$verify\_scenario\{seed\} == success$}{
            \textbf{break}
          }
        }
        $S \gets S \cup seed$
      }
    }
  }
  \Begin($verify\_scenario{(}s{)}$ \label{algo:line:verify}) {
    $states \gets executor.simulate(s)$ \label{algo:line:dryrun}
    
    \uIf{$detector.check\_misbehavior(states) == false$ \label{algo:line:success}}{
      \Return $success$ // seed mission successfully completed \\
    }
    \Else{
      \Return $retry$ // seed mission failed \\
    }
  }
\end{algorithm}
\end{figure}

\section{On the effectiveness of code coverage-based metrics}
\label{s:afl}

As we discuss in \autoref{ss:feedback},
the coverage-based feedback of
traditional grey-box fuzzers (\eg, AFL~\cite{zalewski2014american})
is less effective in estimating the effectiveness of test driving scenarios
when testing a distributed and stateful system,
because
the behavior of such system is \textit{dominantly driven by the data}
being communicated between distributed nodes,
rather than the control flow.

\begin{figure}[h!]
  \input{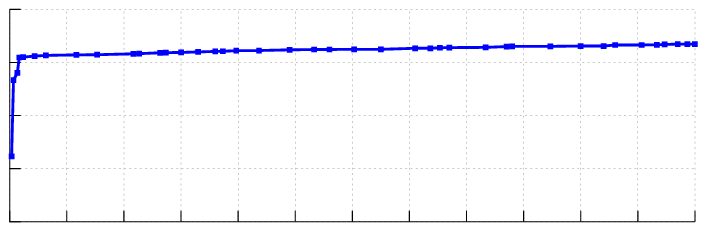}
  \caption{Change of code coverage while fuzzing Autoware for six hours with \sys.}
  \label{f:coverage}
\end{figure}

We demonstrate this in \autoref{f:coverage},
which shows the code coverage measured using the \texttt{gcov} coverage test program
while fuzzing Autoware for six hours.
Since AFL cannot be directly applied to testing Autoware,
we measured the code coverage while fuzzing with \sys.
Even though the code coverage was quickly saturated at approximately 32\%,
the ego-vehicle \textit{showed a wide variety of behaviors}
driving in the mutated scenarios after it reached the saturation point,
\eg, navigating through different parts of the map,
and exhibited multiple misbehaviors
including collisions and speed limit violations.
The uncovered code included unused and irrelevant portions of Autoware,
such as visualization, graphic user interface (GUI),
and unused modules (\eg, alternative controllers).
This substantiates our claim that
code coverage is not an effective metric
to approximate the test coverage
in finding bugs of \adses.



\section{Figures for Driving Quality Metrics Correctness Evaluation}
\label{s:qualityfig}

\begin{figure}[b]
\centering
  \subfigure[Frame 1]{
    \label{fig:us1}
    \includegraphics[width=0.11\textwidth]{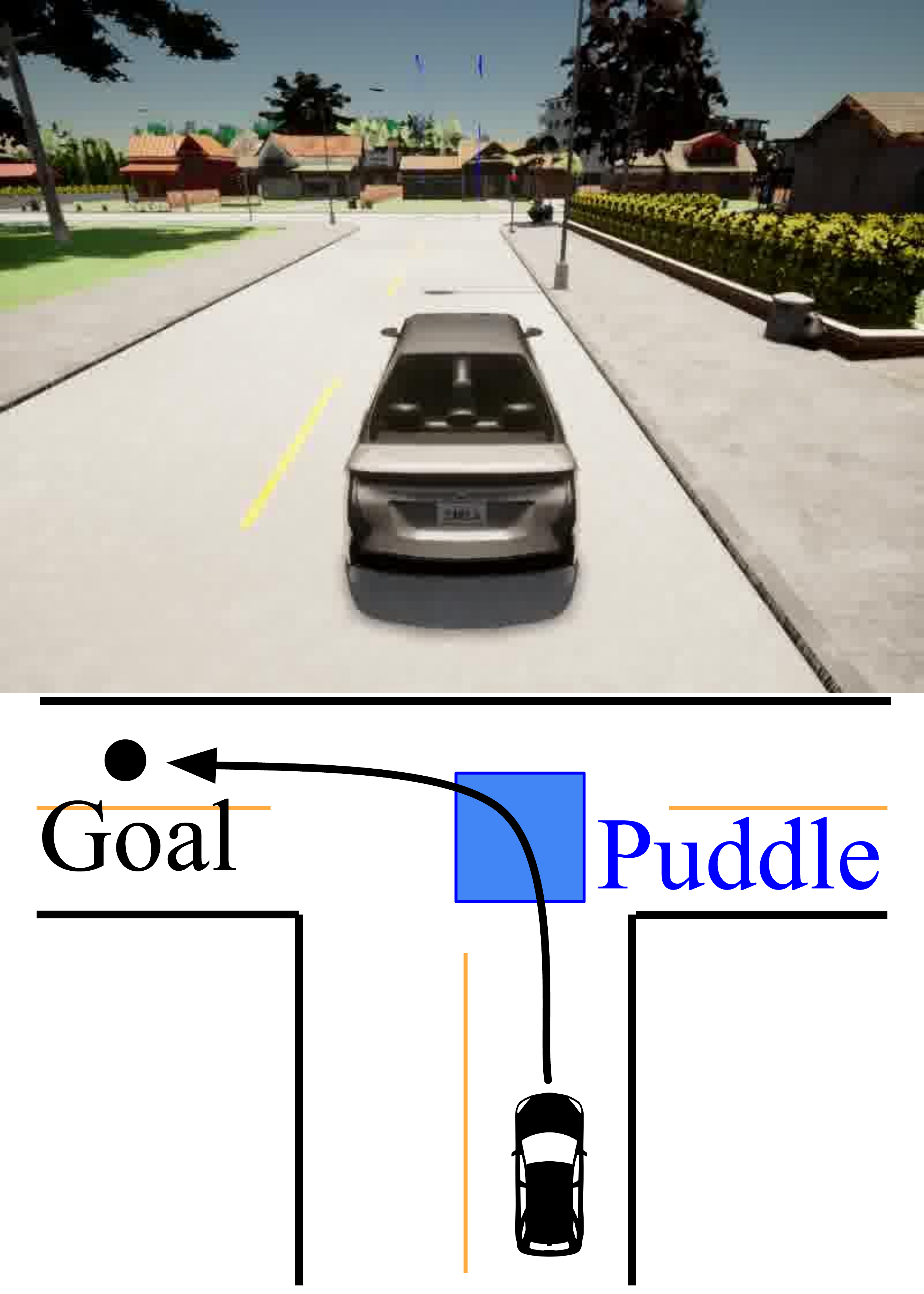}
  }
  \hspace{-0.8em}
  \hfill
  \subfigure[Frame 150]{
    \label{fig:us2}
    \includegraphics[width=0.11\textwidth]{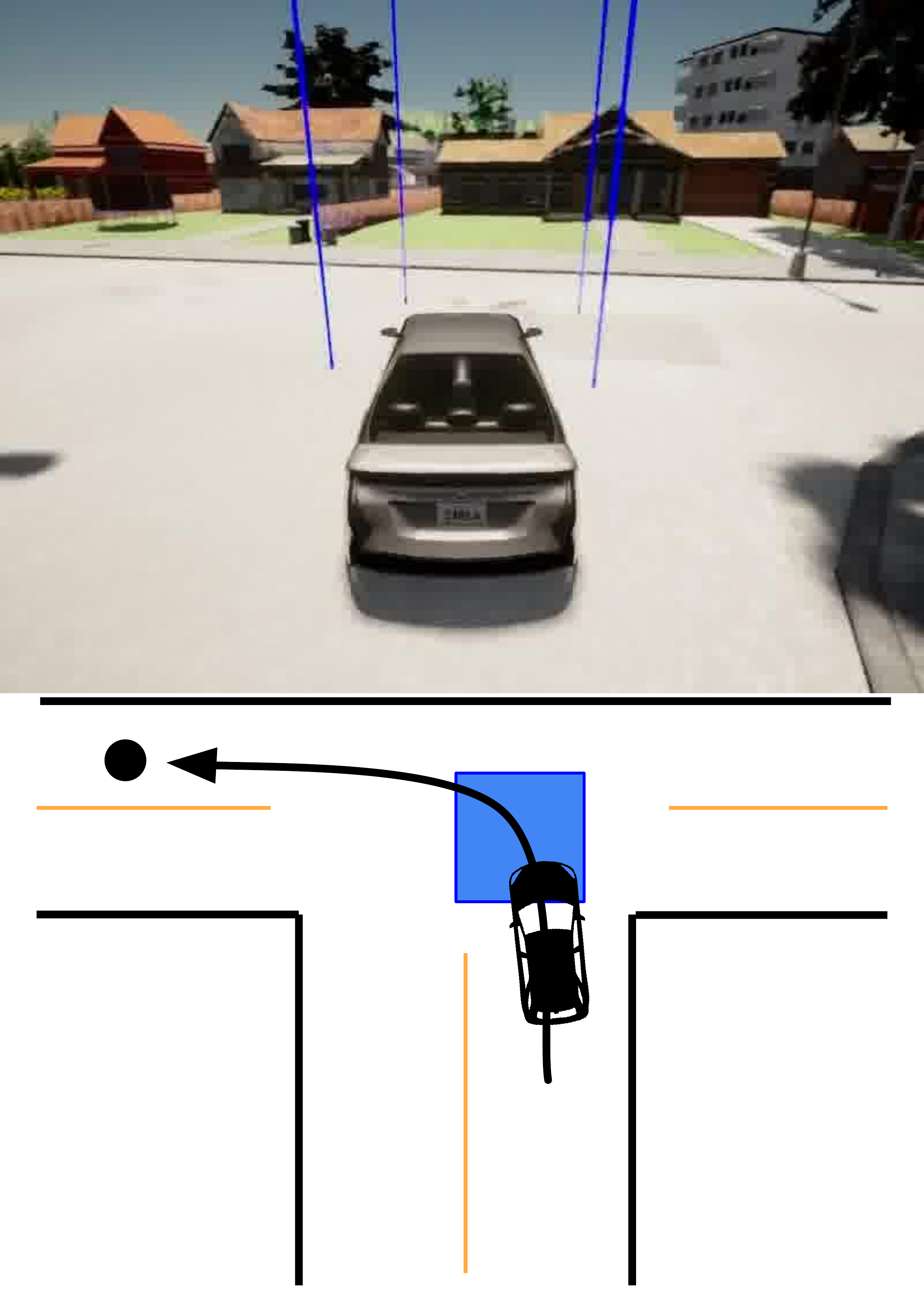}
  }
  \hspace{-0.8em}
  \hfill
  \subfigure[Frame 178]{
    \label{fig:us3}
    \includegraphics[width=0.11\textwidth]{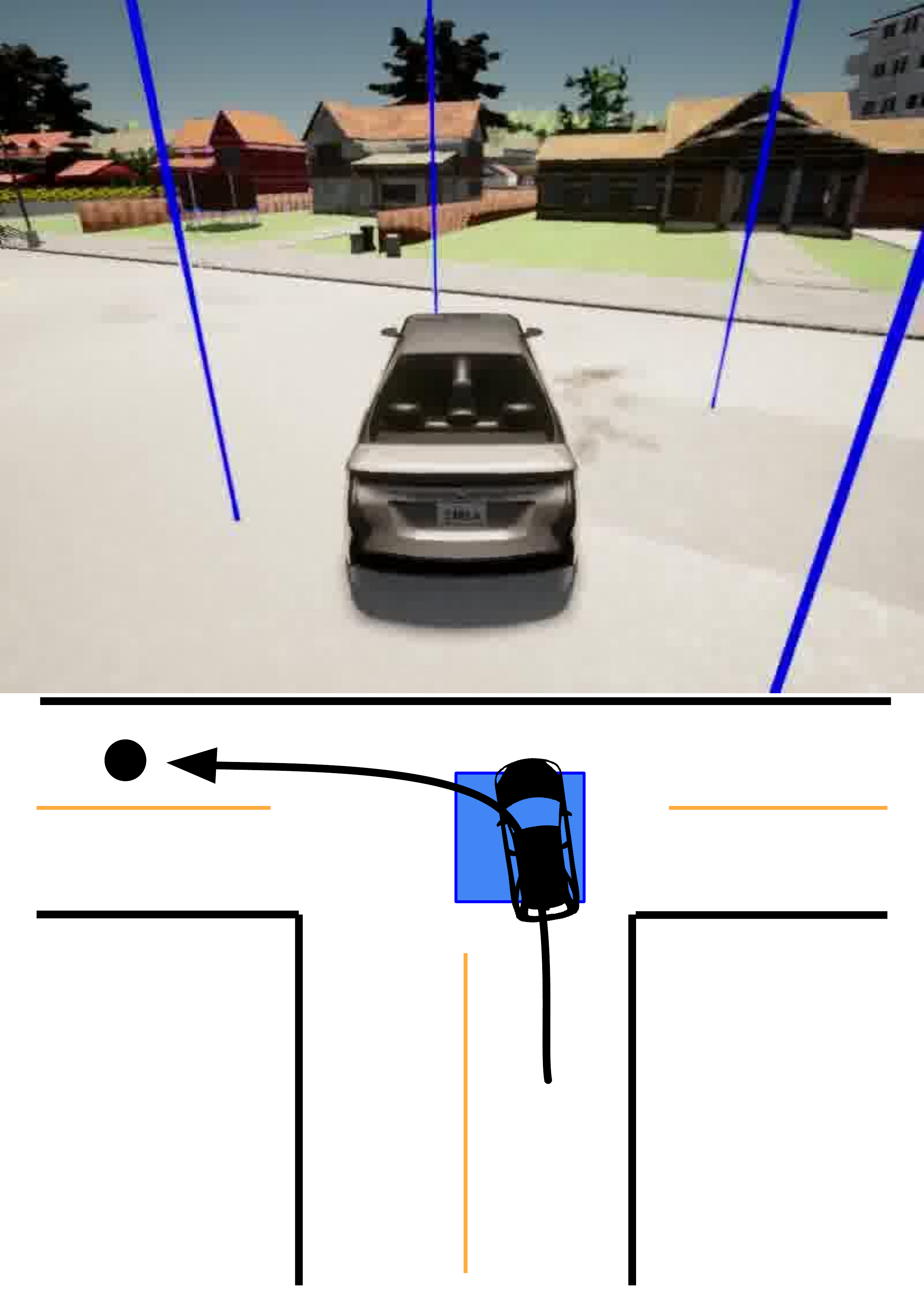}
  }
  \hspace{-0.8em}
  \hfill
  \subfigure[Frame 220]{
    \label{fig:us4}
    \includegraphics[width=0.11\textwidth]{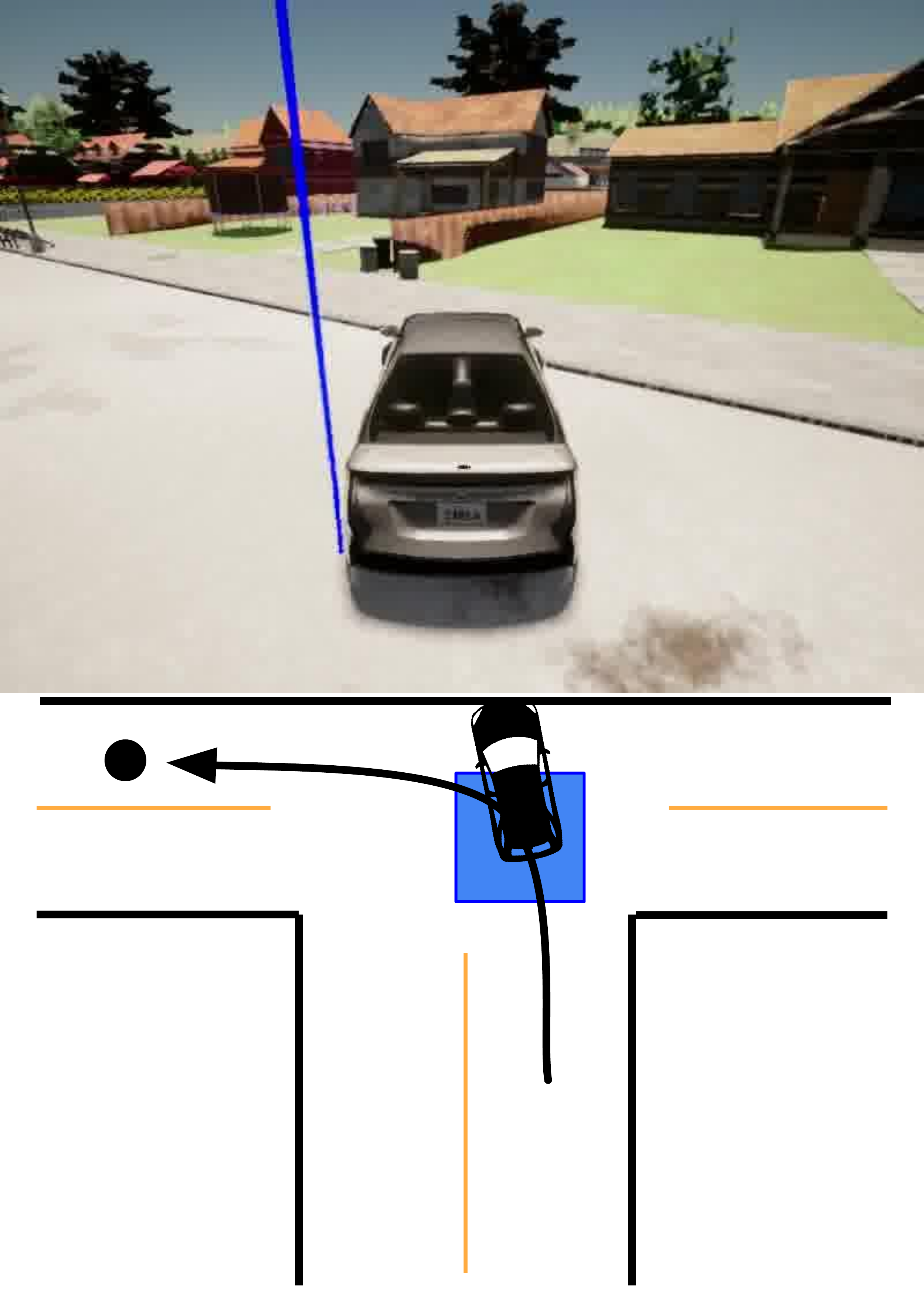}
  }
  \caption{
    Simulated scenario of the understeer experiment.
    At the frame 150, the vehicle enters the puddle (the blue box)
    while turning left to get to the destination.
    At the frame 178, the vehicle is sliding out to the right
    even though it tries to turn left,
    and at the frame 220, the vehicle exits the puddle right before invading
    the sidewalk.
  }
  \label{f:usfig}

\end{figure}

Figure~\ref{f:usfig}$\sim$\ref{f:osgraph} show the correctness of driving quality metrics.
They are omitted in \autoref{ss:eval-quality} due to the space.

\begin{figure}[h!]
  \resizebox{0.3\textwidth}{!}{
    \input{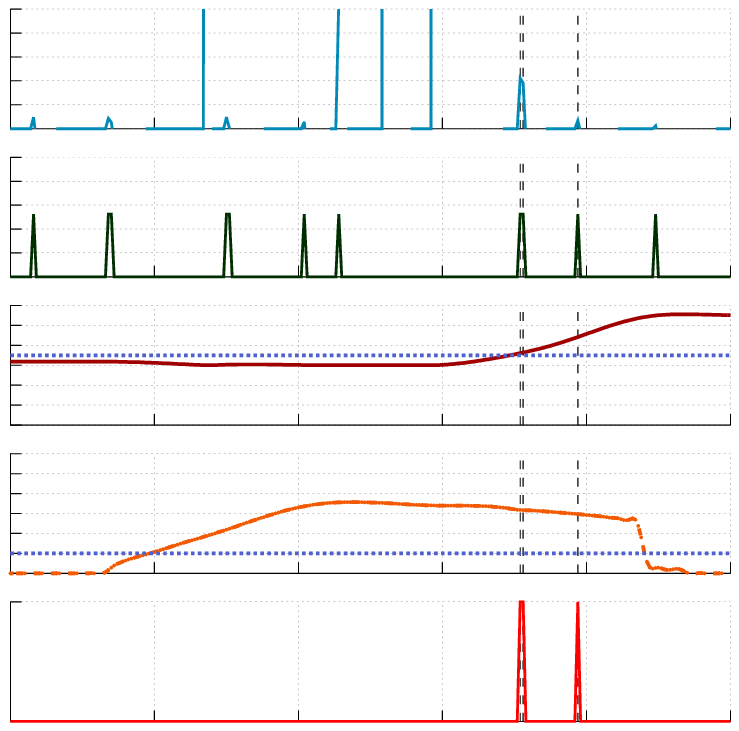}
  }
  \vspace*{-4em}
  \caption{
    Fractional drop ($FD$) of the lateral acceleration ($A_y$),
    understeer indicator ($USI$),
    steering wheel angle ($SWA$),
    longitudinal speed ($V_x$),
    and final understeer detection
    throughout the scenario shown in \autoref{f:usfig}.
    The blue dotted lines shown in the $SWA$ (10 deg) and $V_x$ (5 $km/h$) graphs
    are activation thresholds;
    $USI$ combined with $SWA$ and $V_x$ above thresholds
    lead to final understeer calls at the frames 177, 178, and 197
    (marked by dashed vertical lines).
  }
  \label{f:usgraph}
  \vspace*{-1em}
\end{figure}

\begin{figure}[h!]
  \subfigure[Frame 1]{
    \label{fig:os1}
    \includegraphics[width=0.11\textwidth]{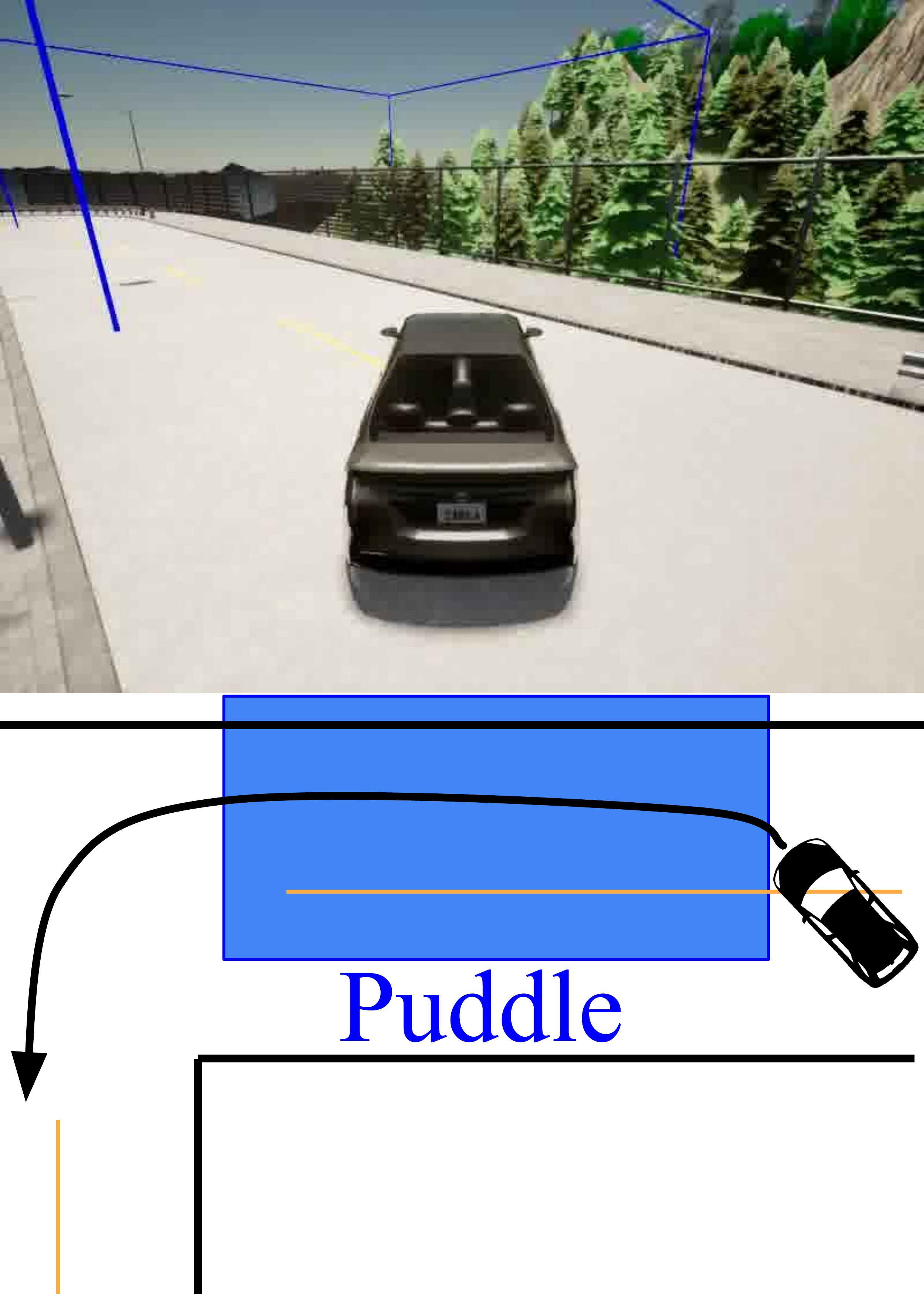}
  }
  \hspace{-0.8em}
  \hfill
  \subfigure[Frame 58]{
    \label{fig:os2}
    \includegraphics[width=0.11\textwidth]{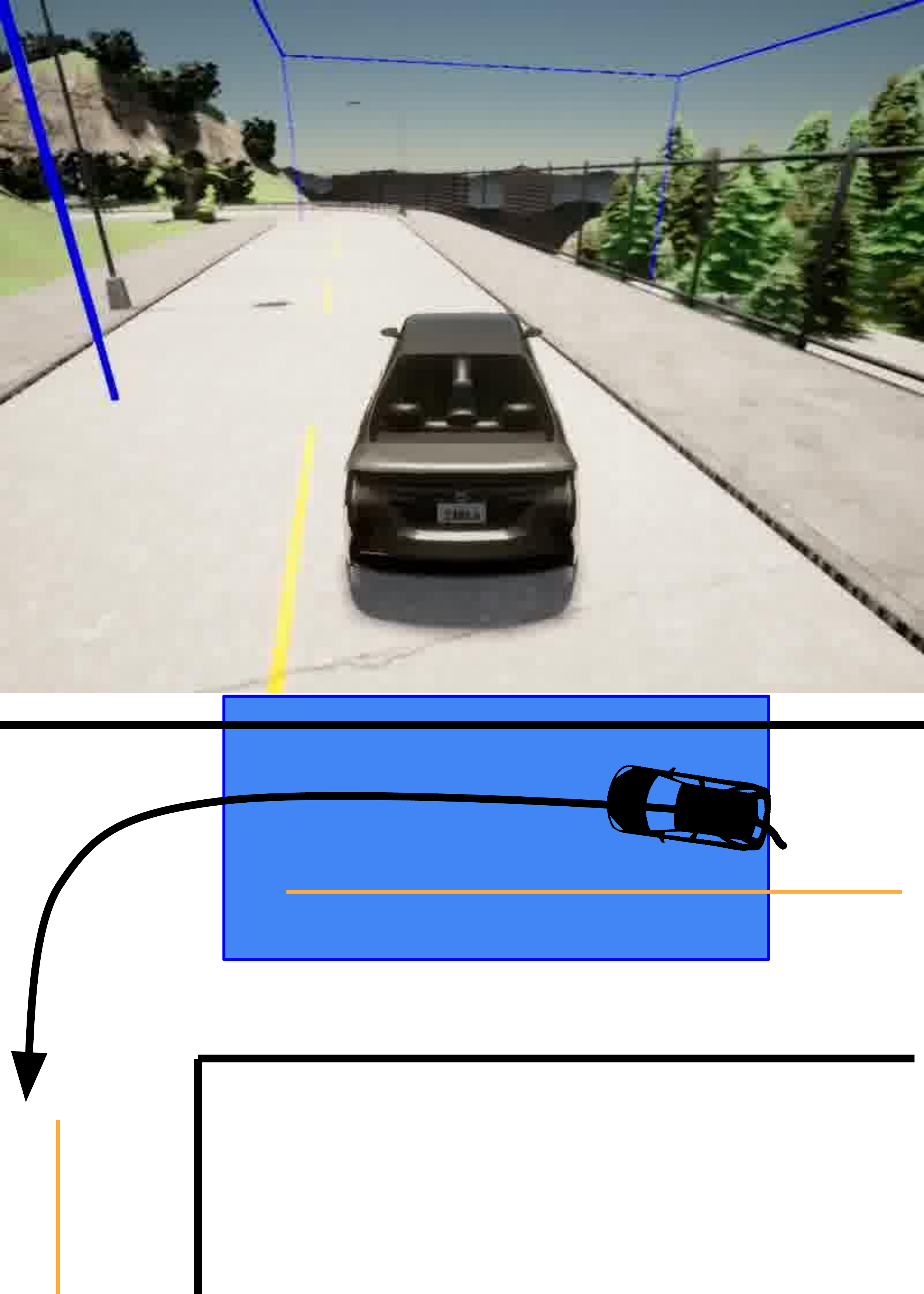}
  }
  \hspace{-0.8em}
  \hfill
  \subfigure[Frame 108]{
    \label{fig:os3}
    \includegraphics[width=0.11\textwidth]{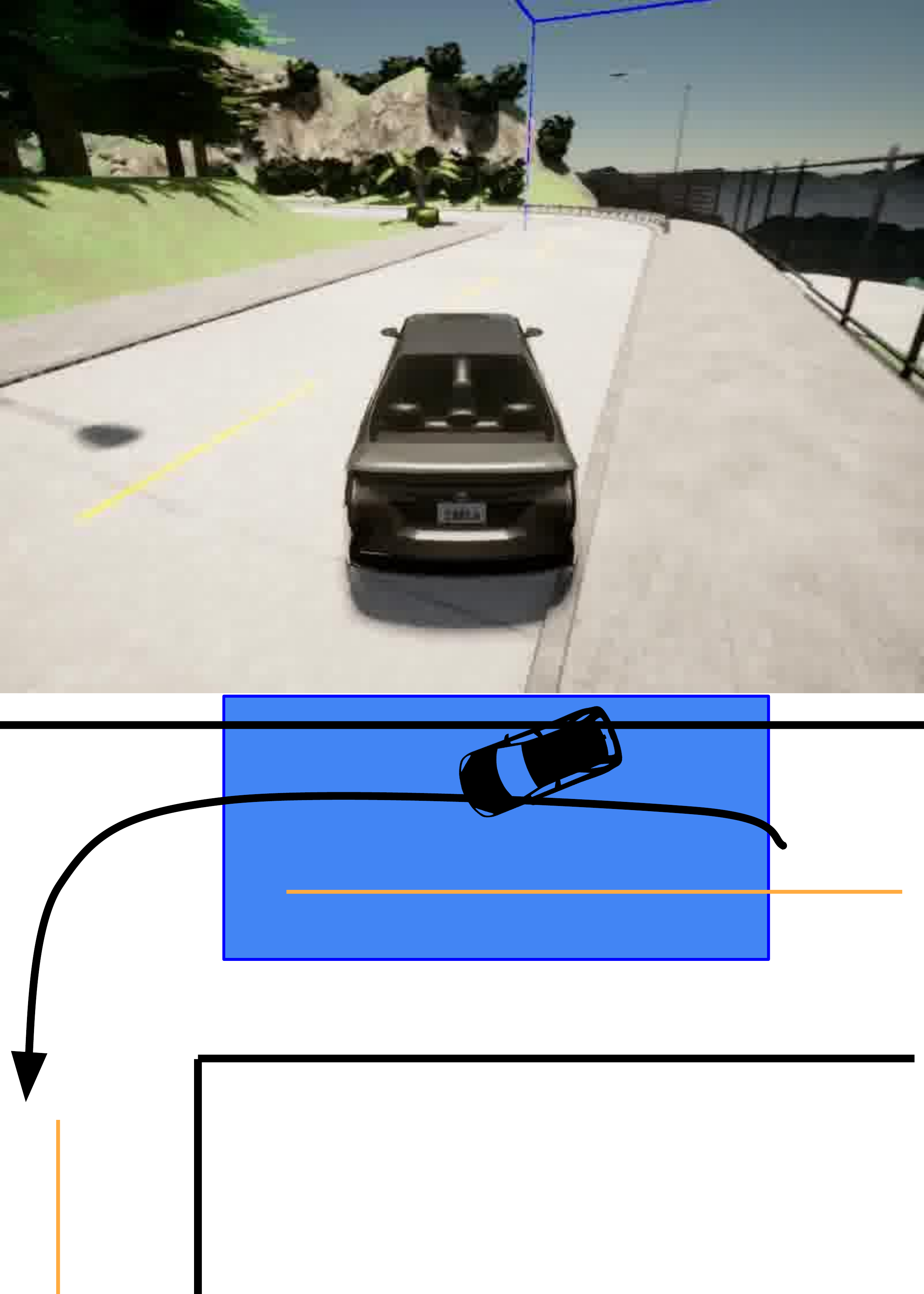}
  }
  \hspace{-0.8em}
  \hfill
  \subfigure[Frame 185]{
    \label{fig:os4}
    \includegraphics[width=0.11\textwidth]{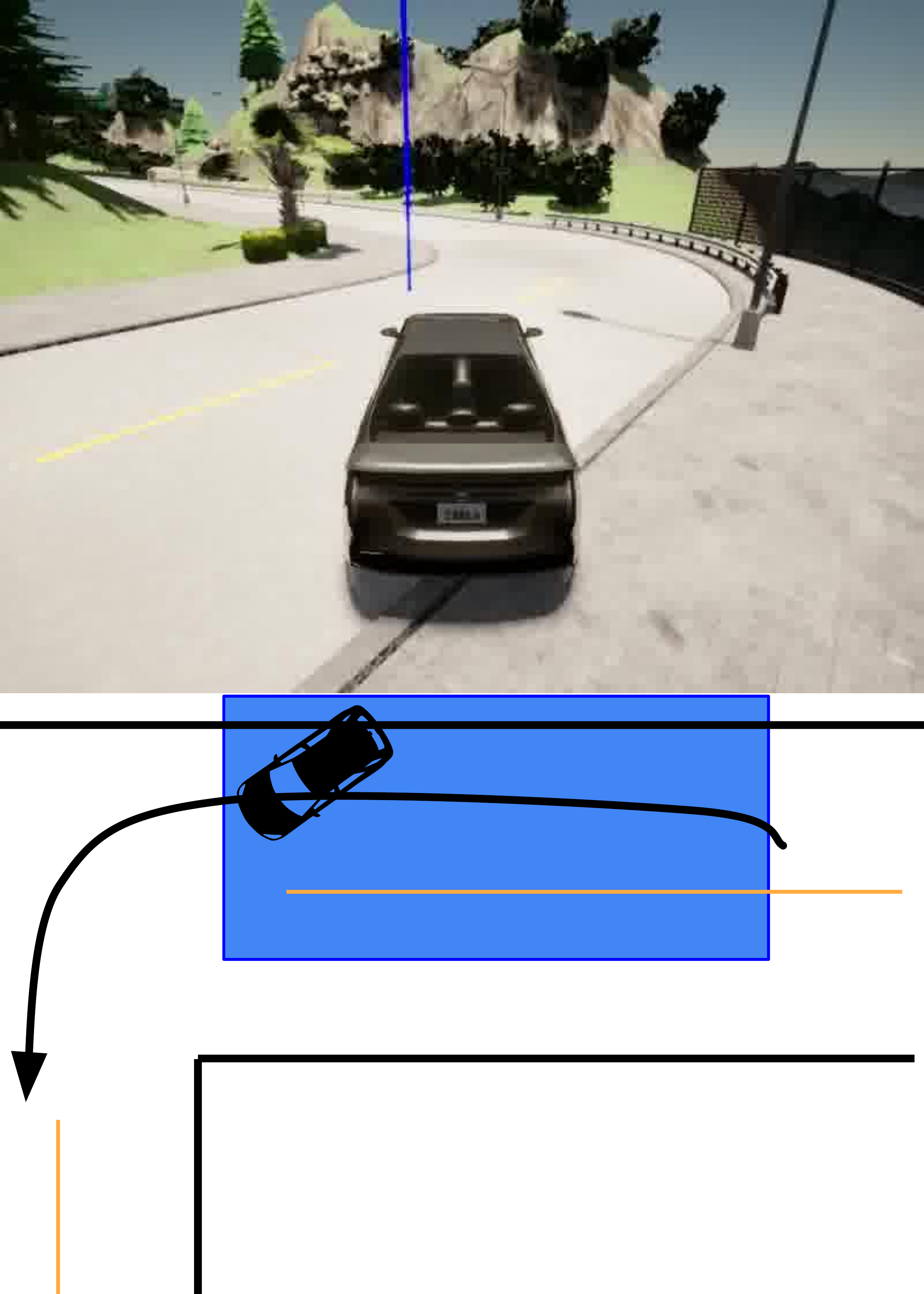}
  }
  \vspace*{-1em}
  \caption{
    Simulated scenario of the oversteer experiment.
    At the frame 58, the vehicle enters the puddle (the blue box)
    with a non-zero yaw speed ($V_x$).
    At the frame 108, the rear end of the vehicle rotates counter-clockwise
    even though the steering amount small.
    The rotation stops, but the vehicle continues to slide
    until it reaches the end of the puddle at the frame 185.
  }
  \label{f:osfig}
\end{figure}

\begin{figure}[h!]
  \resizebox{0.3\textwidth}{!}{
    \input{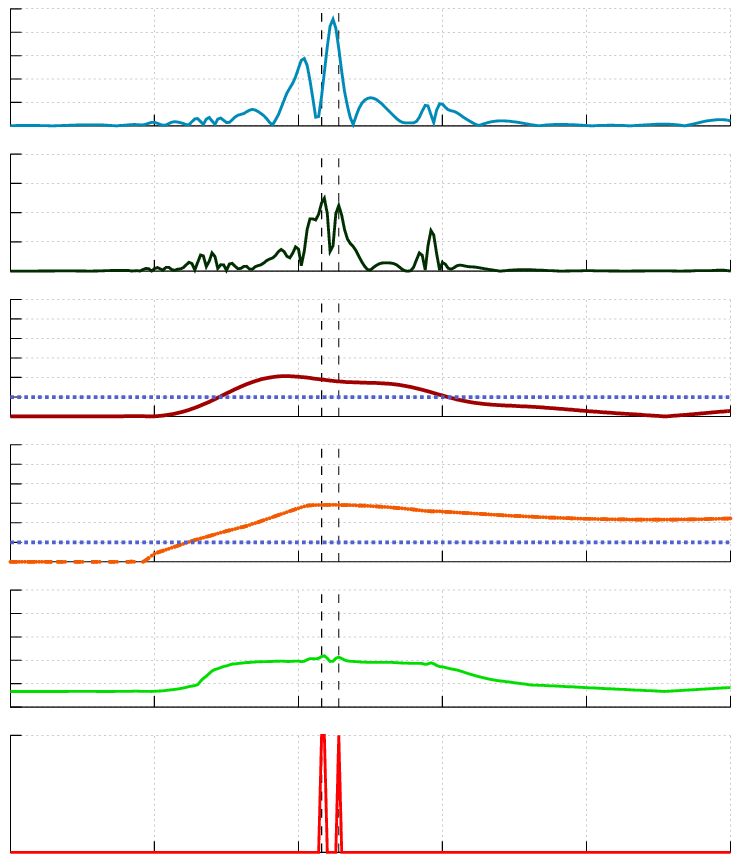}
  }
  \vspace*{-4.4em}
  \caption{
    Difference of lightly- and heavily-filtered
    steering wheel angles ($dSWA$),
    difference of filtered lateral acceleration ($dA_y$),
    yaw speed ($AV_z$),
    longitudinal speed ($V_x$),
    oversteer indicator ($OSI$),
    and final oversteer detected
    in the scenario shown in \autoref{f:osfig}.
    When the vehicle starts to oversteer at the frame 108
    as the rear-end loses grip and slips out,
    and when the state exacerbates at the frame 114,
    the oversteer events are detected (marked by dashed vertical lines).
  }
  \label{f:osgraph}
\end{figure}


%% file: data/coverage.tex
\begingroup
\scriptsize
  \makeatletter
  \providecommand\color[2][]{%
    \GenericError{(gnuplot) \space\space\space\@spaces}{%
      Package color not loaded in conjunction with
      terminal option `colourtext'%
    }{See the gnuplot documentation for explanation.%
    }{Either use 'blacktext' in gnuplot or load the package
      color.sty in LaTeX.}%
    \renewcommand\color[2][]{}%
  }%
  \providecommand\includegraphics[2][]{%
    \GenericError{(gnuplot) \space\space\space\@spaces}{%
      Package graphicx or graphics not loaded%
    }{See the gnuplot documentation for explanation.%
    }{The gnuplot epslatex terminal needs graphicx.sty or graphics.sty.}%
    \renewcommand\includegraphics[2][]{}%
  }%
  \providecommand\rotatebox[2]{#2}%
  \@ifundefined{ifGPcolor}{%
    \newif\ifGPcolor
    \GPcolortrue
  }{}%
  \@ifundefined{ifGPblacktext}{%
    \newif\ifGPblacktext
    \GPblacktextfalse
  }{}%
  \let\gplgaddtomacro\g@addto@macro
  \gdef\gplbacktext{}%
  \gdef\gplfronttext{}%
  \makeatother
  \ifGPblacktext
    \def\colorrgb#1{}%
    \def\colorgray#1{}%
  \else
    \ifGPcolor
      \def\colorrgb#1{\color[rgb]{#1}}%
      \def\colorgray#1{\color[gray]{#1}}%
      \expandafter\def\csname LTw\endcsname{\color{white}}%
      \expandafter\def\csname LTb\endcsname{\color{black}}%
      \expandafter\def\csname LTa\endcsname{\color{black}}%
      \expandafter\def\csname LT0\endcsname{\color[rgb]{1,0,0}}%
      \expandafter\def\csname LT1\endcsname{\color[rgb]{0,1,0}}%
      \expandafter\def\csname LT2\endcsname{\color[rgb]{0,0,1}}%
      \expandafter\def\csname LT3\endcsname{\color[rgb]{1,0,1}}%
      \expandafter\def\csname LT4\endcsname{\color[rgb]{0,1,1}}%
      \expandafter\def\csname LT5\endcsname{\color[rgb]{1,1,0}}%
      \expandafter\def\csname LT6\endcsname{\color[rgb]{0,0,0}}%
      \expandafter\def\csname LT7\endcsname{\color[rgb]{1,0.3,0}}%
      \expandafter\def\csname LT8\endcsname{\color[rgb]{0.5,0.5,0.5}}%
    \else
      \def\colorrgb#1{\color{black}}%
      \def\colorgray#1{\color[gray]{#1}}%
      \expandafter\def\csname LTw\endcsname{\color{white}}%
      \expandafter\def\csname LTb\endcsname{\color{black}}%
      \expandafter\def\csname LTa\endcsname{\color{black}}%
      \expandafter\def\csname LT0\endcsname{\color{black}}%
      \expandafter\def\csname LT1\endcsname{\color{black}}%
      \expandafter\def\csname LT2\endcsname{\color{black}}%
      \expandafter\def\csname LT3\endcsname{\color{black}}%
      \expandafter\def\csname LT4\endcsname{\color{black}}%
      \expandafter\def\csname LT5\endcsname{\color{black}}%
      \expandafter\def\csname LT6\endcsname{\color{black}}%
      \expandafter\def\csname LT7\endcsname{\color{black}}%
      \expandafter\def\csname LT8\endcsname{\color{black}}%
    \fi
  \fi
    \setlength{\unitlength}{0.0500bp}%
    \ifx\gptboxheight\undefined%
      \newlength{\gptboxheight}%
      \newlength{\gptboxwidth}%
      \newsavebox{\gptboxtext}%
    \fi%
    \setlength{\fboxrule}{0.5pt}%
    \setlength{\fboxsep}{1pt}%
    \definecolor{tbcol}{rgb}{1,1,1}%
\begin{picture}(4608.00,1728.00)%
    \gplgaddtomacro\gplbacktext{%
      \csname LTb\endcsname
      \put(372,384){\makebox(0,0)[r]{\strut{}$0$}}%
      \csname LTb\endcsname
      \put(372,690){\makebox(0,0)[r]{\strut{}$10$}}%
      \csname LTb\endcsname
      \put(372,996){\makebox(0,0)[r]{\strut{}$20$}}%
      \csname LTb\endcsname
      \put(372,1301){\makebox(0,0)[r]{\strut{}$30$}}%
      \csname LTb\endcsname
      \put(372,1607){\makebox(0,0)[r]{\strut{}$40$}}%
      \csname LTb\endcsname
      \put(444,264){\makebox(0,0){\strut{}$0$}}%
      \csname LTb\endcsname
      \put(773,264){\makebox(0,0){\strut{}$30$}}%
      \csname LTb\endcsname
      \put(1102,264){\makebox(0,0){\strut{}$60$}}%
      \csname LTb\endcsname
      \put(1431,264){\makebox(0,0){\strut{}$90$}}%
      \csname LTb\endcsname
      \put(1760,264){\makebox(0,0){\strut{}$120$}}%
      \csname LTb\endcsname
      \put(2089,264){\makebox(0,0){\strut{}$150$}}%
      \csname LTb\endcsname
      \put(2418,264){\makebox(0,0){\strut{}$180$}}%
      \csname LTb\endcsname
      \put(2746,264){\makebox(0,0){\strut{}$210$}}%
      \csname LTb\endcsname
      \put(3075,264){\makebox(0,0){\strut{}$240$}}%
      \csname LTb\endcsname
      \put(3404,264){\makebox(0,0){\strut{}$270$}}%
      \csname LTb\endcsname
      \put(3733,264){\makebox(0,0){\strut{}$300$}}%
      \csname LTb\endcsname
      \put(4062,264){\makebox(0,0){\strut{}$330$}}%
      \csname LTb\endcsname
      \put(4391,264){\makebox(0,0){\strut{}$360$}}%
    }%
    \gplgaddtomacro\gplfronttext{%
      \csname LTb\endcsname
      \put(114,995){\rotatebox{-270}{\makebox(0,0){\strut{}Coverage (\%)}}}%
      \put(2417,84){\makebox(0,0){\strut{}Fuzzing time (min)}}%
    }%
    \gplbacktext
    \put(0,0){\includegraphics[width={230.40bp},height={86.40bp}]{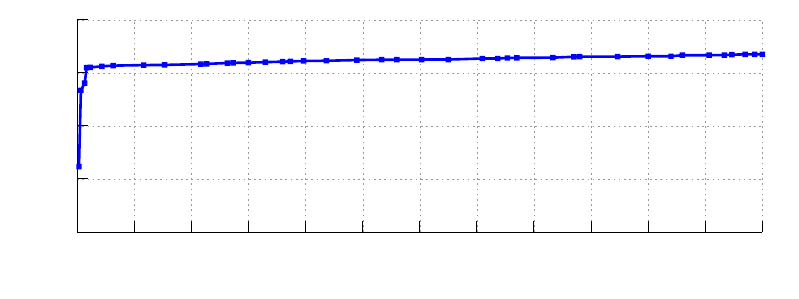}}%
    \gplfronttext
  \end{picture}%
\endgroup

%% file: data/understeer.tex
\begingroup
\scriptsize
  \makeatletter
  \providecommand\color[2][]{%
    \GenericError{(gnuplot) \space\space\space\@spaces}{%
      Package color not loaded in conjunction with
      terminal option `colourtext'%
    }{See the gnuplot documentation for explanation.%
    }{Either use 'blacktext' in gnuplot or load the package
      color.sty in LaTeX.}%
    \renewcommand\color[2][]{}%
  }%
  \providecommand\includegraphics[2][]{%
    \GenericError{(gnuplot) \space\space\space\@spaces}{%
      Package graphicx or graphics not loaded%
    }{See the gnuplot documentation for explanation.%
    }{The gnuplot epslatex terminal needs graphicx.sty or graphics.sty.}%
    \renewcommand\includegraphics[2][]{}%
  }%
  \providecommand\rotatebox[2]{#2}%
  \@ifundefined{ifGPcolor}{%
    \newif\ifGPcolor
    \GPcolortrue
  }{}%
  \@ifundefined{ifGPblacktext}{%
    \newif\ifGPblacktext
    \GPblacktextfalse
  }{}%
  \let\gplgaddtomacro\g@addto@macro
  \gdef\gplbacktext{}%
  \gdef\gplfronttext{}%
  \makeatother
  \ifGPblacktext
    \def\colorrgb#1{}%
    \def\colorgray#1{}%
  \else
    \ifGPcolor
      \def\colorrgb#1{\color[rgb]{#1}}%
      \def\colorgray#1{\color[gray]{#1}}%
      \expandafter\def\csname LTw\endcsname{\color{white}}%
      \expandafter\def\csname LTb\endcsname{\color{black}}%
      \expandafter\def\csname LTa\endcsname{\color{black}}%
      \expandafter\def\csname LT0\endcsname{\color[rgb]{1,0,0}}%
      \expandafter\def\csname LT1\endcsname{\color[rgb]{0,1,0}}%
      \expandafter\def\csname LT2\endcsname{\color[rgb]{0,0,1}}%
      \expandafter\def\csname LT3\endcsname{\color[rgb]{1,0,1}}%
      \expandafter\def\csname LT4\endcsname{\color[rgb]{0,1,1}}%
      \expandafter\def\csname LT5\endcsname{\color[rgb]{1,1,0}}%
      \expandafter\def\csname LT6\endcsname{\color[rgb]{0,0,0}}%
      \expandafter\def\csname LT7\endcsname{\color[rgb]{1,0.3,0}}%
      \expandafter\def\csname LT8\endcsname{\color[rgb]{0.5,0.5,0.5}}%
    \else
      \def\colorrgb#1{\color{black}}%
      \def\colorgray#1{\color[gray]{#1}}%
      \expandafter\def\csname LTw\endcsname{\color{white}}%
      \expandafter\def\csname LTb\endcsname{\color{black}}%
      \expandafter\def\csname LTa\endcsname{\color{black}}%
      \expandafter\def\csname LT0\endcsname{\color{black}}%
      \expandafter\def\csname LT1\endcsname{\color{black}}%
      \expandafter\def\csname LT2\endcsname{\color{black}}%
      \expandafter\def\csname LT3\endcsname{\color{black}}%
      \expandafter\def\csname LT4\endcsname{\color{black}}%
      \expandafter\def\csname LT5\endcsname{\color{black}}%
      \expandafter\def\csname LT6\endcsname{\color{black}}%
      \expandafter\def\csname LT7\endcsname{\color{black}}%
      \expandafter\def\csname LT8\endcsname{\color{black}}%
    \fi
  \fi
    \setlength{\unitlength}{0.0500bp}%
    \ifx\gptboxheight\undefined%
      \newlength{\gptboxheight}%
      \newlength{\gptboxwidth}%
      \newsavebox{\gptboxtext}%
    \fi%
    \setlength{\fboxrule}{0.5pt}%
    \setlength{\fboxsep}{1pt}%
\begin{picture}(4608.00,5472.00)%
    \gplgaddtomacro\gplbacktext{%
      \csname LTb\endcsname
      \put(388,4727){\makebox(0,0)[r]{\strut{}$0$}}%
      \csname LTb\endcsname
      \put(388,4865){\makebox(0,0)[r]{\strut{}$0.02$}}%
      \csname LTb\endcsname
      \put(388,5003){\makebox(0,0)[r]{\strut{}$0.04$}}%
      \csname LTb\endcsname
      \put(388,5140){\makebox(0,0)[r]{\strut{}$0.06$}}%
      \csname LTb\endcsname
      \put(388,5278){\makebox(0,0)[r]{\strut{}$0.08$}}%
      \csname LTb\endcsname
      \put(388,5416){\makebox(0,0)[r]{\strut{}$0.1$}}%
      \csname LTb\endcsname
      \put(460,4607){\makebox(0,0){\strut{}}}%
      \csname LTb\endcsname
      \put(1289,4607){\makebox(0,0){\strut{}}}%
      \csname LTb\endcsname
      \put(2119,4607){\makebox(0,0){\strut{}}}%
      \csname LTb\endcsname
      \put(2948,4607){\makebox(0,0){\strut{}}}%
      \csname LTb\endcsname
      \put(3778,4607){\makebox(0,0){\strut{}}}%
      \csname LTb\endcsname
      \put(4607,4607){\makebox(0,0){\strut{}}}%
    }%
    \gplgaddtomacro\gplfronttext{%
      \csname LTb\endcsname
      \put(-14,5071){\rotatebox{-270}{\makebox(0,0){\strut{}FD}}}%
    }%
    \gplgaddtomacro\gplbacktext{%
      \csname LTb\endcsname
      \put(388,3874){\makebox(0,0)[r]{\strut{}$0$}}%
      \csname LTb\endcsname
      \put(388,4012){\makebox(0,0)[r]{\strut{}$2$}}%
      \csname LTb\endcsname
      \put(388,4149){\makebox(0,0)[r]{\strut{}$4$}}%
      \csname LTb\endcsname
      \put(388,4287){\makebox(0,0)[r]{\strut{}$6$}}%
      \csname LTb\endcsname
      \put(388,4424){\makebox(0,0)[r]{\strut{}$8$}}%
      \csname LTb\endcsname
      \put(388,4562){\makebox(0,0)[r]{\strut{}$10$}}%
      \csname LTb\endcsname
      \put(460,3754){\makebox(0,0){\strut{}}}%
      \csname LTb\endcsname
      \put(1289,3754){\makebox(0,0){\strut{}}}%
      \csname LTb\endcsname
      \put(2119,3754){\makebox(0,0){\strut{}}}%
      \csname LTb\endcsname
      \put(2948,3754){\makebox(0,0){\strut{}}}%
      \csname LTb\endcsname
      \put(3778,3754){\makebox(0,0){\strut{}}}%
      \csname LTb\endcsname
      \put(4607,3754){\makebox(0,0){\strut{}}}%
    }%
    \gplgaddtomacro\gplfronttext{%
      \csname LTb\endcsname
      \put(130,4218){\rotatebox{-270}{\makebox(0,0){\strut{}USI}}}%
    }%
    \gplgaddtomacro\gplbacktext{%
      \csname LTb\endcsname
      \put(388,3020){\makebox(0,0)[r]{\strut{}$-60$}}%
      \csname LTb\endcsname
      \put(388,3135){\makebox(0,0)[r]{\strut{}$-40$}}%
      \csname LTb\endcsname
      \put(388,3250){\makebox(0,0)[r]{\strut{}$-20$}}%
      \csname LTb\endcsname
      \put(388,3365){\makebox(0,0)[r]{\strut{}$0$}}%
      \csname LTb\endcsname
      \put(388,3479){\makebox(0,0)[r]{\strut{}$20$}}%
      \csname LTb\endcsname
      \put(388,3594){\makebox(0,0)[r]{\strut{}$40$}}%
      \csname LTb\endcsname
      \put(388,3709){\makebox(0,0)[r]{\strut{}$60$}}%
      \csname LTb\endcsname
      \put(460,2900){\makebox(0,0){\strut{}}}%
      \csname LTb\endcsname
      \put(1289,2900){\makebox(0,0){\strut{}}}%
      \csname LTb\endcsname
      \put(2119,2900){\makebox(0,0){\strut{}}}%
      \csname LTb\endcsname
      \put(2948,2900){\makebox(0,0){\strut{}}}%
      \csname LTb\endcsname
      \put(3778,2900){\makebox(0,0){\strut{}}}%
      \csname LTb\endcsname
      \put(4607,2900){\makebox(0,0){\strut{}}}%
    }%
    \gplgaddtomacro\gplfronttext{%
      \csname LTb\endcsname
      \put(58,3364){\rotatebox{-270}{\makebox(0,0){\strut{}SWA ($deg$)}}}%
    }%
    \gplgaddtomacro\gplbacktext{%
      \csname LTb\endcsname
      \put(388,2166){\makebox(0,0)[r]{\strut{}$0$}}%
      \csname LTb\endcsname
      \put(388,2281){\makebox(0,0)[r]{\strut{}$5$}}%
      \csname LTb\endcsname
      \put(388,2396){\makebox(0,0)[r]{\strut{}$10$}}%
      \csname LTb\endcsname
      \put(388,2511){\makebox(0,0)[r]{\strut{}$15$}}%
      \csname LTb\endcsname
      \put(388,2625){\makebox(0,0)[r]{\strut{}$20$}}%
      \csname LTb\endcsname
      \put(388,2740){\makebox(0,0)[r]{\strut{}$25$}}%
      \csname LTb\endcsname
      \put(388,2855){\makebox(0,0)[r]{\strut{}$30$}}%
      \csname LTb\endcsname
      \put(460,2046){\makebox(0,0){\strut{}}}%
      \csname LTb\endcsname
      \put(1289,2046){\makebox(0,0){\strut{}}}%
      \csname LTb\endcsname
      \put(2119,2046){\makebox(0,0){\strut{}}}%
      \csname LTb\endcsname
      \put(2948,2046){\makebox(0,0){\strut{}}}%
      \csname LTb\endcsname
      \put(3778,2046){\makebox(0,0){\strut{}}}%
      \csname LTb\endcsname
      \put(4607,2046){\makebox(0,0){\strut{}}}%
    }%
    \gplgaddtomacro\gplfronttext{%
      \csname LTb\endcsname
      \put(130,2510){\rotatebox{-270}{\makebox(0,0){\strut{}$V_x$ ($km/h$)}}}%
    }%
    \gplgaddtomacro\gplbacktext{%
      \csname LTb\endcsname
      \put(388,1313){\makebox(0,0)[r]{\strut{}$0$}}%
      \csname LTb\endcsname
      \put(388,2002){\makebox(0,0)[r]{\strut{}$1$}}%
      \csname LTb\endcsname
      \put(460,1193){\makebox(0,0){\strut{}$0$}}%
      \csname LTb\endcsname
      \put(1289,1193){\makebox(0,0){\strut{}$50$}}%
      \csname LTb\endcsname
      \put(2119,1193){\makebox(0,0){\strut{}$100$}}%
      \csname LTb\endcsname
      \put(2948,1193){\makebox(0,0){\strut{}$150$}}%
      \csname LTb\endcsname
      \put(3778,1193){\makebox(0,0){\strut{}$200$}}%
      \csname LTb\endcsname
      \put(4607,1193){\makebox(0,0){\strut{}$250$}}%
    }%
    \gplgaddtomacro\gplfronttext{%
      \csname LTb\endcsname
      \put(202,1657){\rotatebox{-270}{\makebox(0,0){\strut{}Understeer}}}%
      \put(2533,1013){\makebox(0,0){\strut{}frame \#}}%
    }%
    \gplbacktext
    \put(0,0){\includegraphics{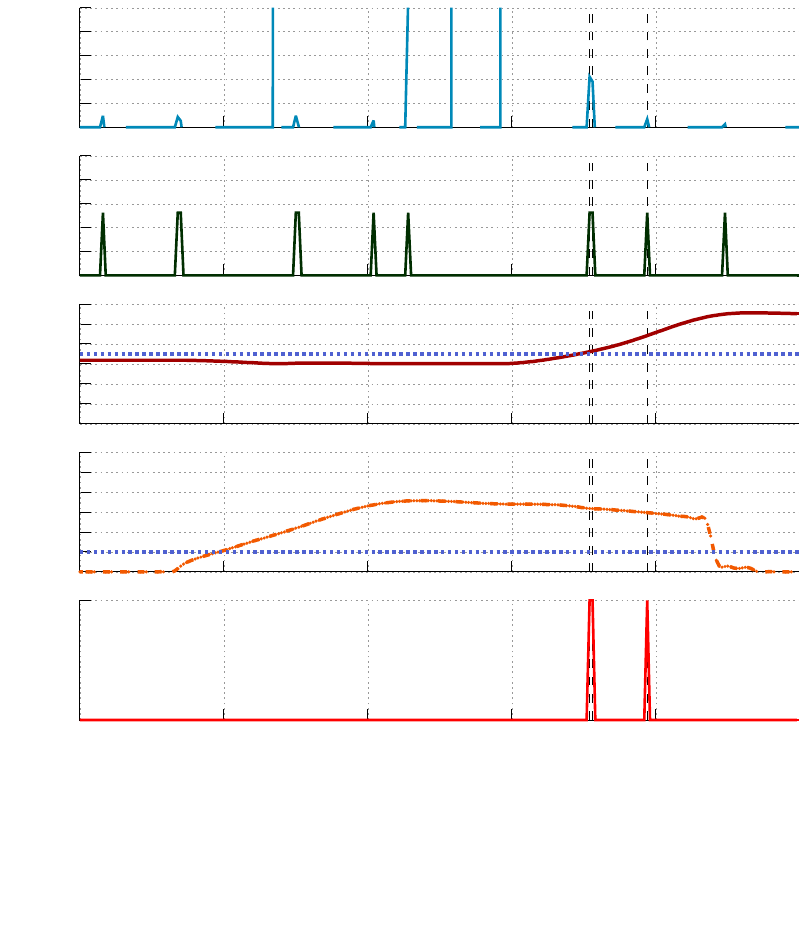}}%
    \gplfronttext
  \end{picture}%
\endgroup

%% file: data/oversteer.tex
\begingroup
\scriptsize
  \makeatletter
  \providecommand\color[2][]{%
    \GenericError{(gnuplot) \space\space\space\@spaces}{%
      Package color not loaded in conjunction with
      terminal option `colourtext'%
    }{See the gnuplot documentation for explanation.%
    }{Either use 'blacktext' in gnuplot or load the package
      color.sty in LaTeX.}%
    \renewcommand\color[2][]{}%
  }%
  \providecommand\includegraphics[2][]{%
    \GenericError{(gnuplot) \space\space\space\@spaces}{%
      Package graphicx or graphics not loaded%
    }{See the gnuplot documentation for explanation.%
    }{The gnuplot epslatex terminal needs graphicx.sty or graphics.sty.}%
    \renewcommand\includegraphics[2][]{}%
  }%
  \providecommand\rotatebox[2]{#2}%
  \@ifundefined{ifGPcolor}{%
    \newif\ifGPcolor
    \GPcolortrue
  }{}%
  \@ifundefined{ifGPblacktext}{%
    \newif\ifGPblacktext
    \GPblacktextfalse
  }{}%
  \let\gplgaddtomacro\g@addto@macro
  \gdef\gplbacktext{}%
  \gdef\gplfronttext{}%
  \makeatother
  \ifGPblacktext
    \def\colorrgb#1{}%
    \def\colorgray#1{}%
  \else
    \ifGPcolor
      \def\colorrgb#1{\color[rgb]{#1}}%
      \def\colorgray#1{\color[gray]{#1}}%
      \expandafter\def\csname LTw\endcsname{\color{white}}%
      \expandafter\def\csname LTb\endcsname{\color{black}}%
      \expandafter\def\csname LTa\endcsname{\color{black}}%
      \expandafter\def\csname LT0\endcsname{\color[rgb]{1,0,0}}%
      \expandafter\def\csname LT1\endcsname{\color[rgb]{0,1,0}}%
      \expandafter\def\csname LT2\endcsname{\color[rgb]{0,0,1}}%
      \expandafter\def\csname LT3\endcsname{\color[rgb]{1,0,1}}%
      \expandafter\def\csname LT4\endcsname{\color[rgb]{0,1,1}}%
      \expandafter\def\csname LT5\endcsname{\color[rgb]{1,1,0}}%
      \expandafter\def\csname LT6\endcsname{\color[rgb]{0,0,0}}%
      \expandafter\def\csname LT7\endcsname{\color[rgb]{1,0.3,0}}%
      \expandafter\def\csname LT8\endcsname{\color[rgb]{0.5,0.5,0.5}}%
    \else
      \def\colorrgb#1{\color{black}}%
      \def\colorgray#1{\color[gray]{#1}}%
      \expandafter\def\csname LTw\endcsname{\color{white}}%
      \expandafter\def\csname LTb\endcsname{\color{black}}%
      \expandafter\def\csname LTa\endcsname{\color{black}}%
      \expandafter\def\csname LT0\endcsname{\color{black}}%
      \expandafter\def\csname LT1\endcsname{\color{black}}%
      \expandafter\def\csname LT2\endcsname{\color{black}}%
      \expandafter\def\csname LT3\endcsname{\color{black}}%
      \expandafter\def\csname LT4\endcsname{\color{black}}%
      \expandafter\def\csname LT5\endcsname{\color{black}}%
      \expandafter\def\csname LT6\endcsname{\color{black}}%
      \expandafter\def\csname LT7\endcsname{\color{black}}%
      \expandafter\def\csname LT8\endcsname{\color{black}}%
    \fi
  \fi
    \setlength{\unitlength}{0.0500bp}%
    \ifx\gptboxheight\undefined%
      \newlength{\gptboxheight}%
      \newlength{\gptboxwidth}%
      \newsavebox{\gptboxtext}%
    \fi%
    \setlength{\fboxrule}{0.5pt}%
    \setlength{\fboxsep}{1pt}%
\begin{picture}(4608.00,6480.00)%
    \gplgaddtomacro\gplbacktext{%
      \csname LTb\endcsname
      \put(388,5740){\makebox(0,0)[r]{\strut{}$0$}}%
      \csname LTb\endcsname
      \put(388,5875){\makebox(0,0)[r]{\strut{}$0.2$}}%
      \csname LTb\endcsname
      \put(388,6010){\makebox(0,0)[r]{\strut{}$0.4$}}%
      \csname LTb\endcsname
      \put(388,6144){\makebox(0,0)[r]{\strut{}$0.6$}}%
      \csname LTb\endcsname
      \put(388,6279){\makebox(0,0)[r]{\strut{}$0.8$}}%
      \csname LTb\endcsname
      \put(388,6414){\makebox(0,0)[r]{\strut{}$1$}}%
      \csname LTb\endcsname
      \put(460,5620){\makebox(0,0){\strut{}}}%
      \csname LTb\endcsname
      \put(1289,5620){\makebox(0,0){\strut{}}}%
      \csname LTb\endcsname
      \put(2119,5620){\makebox(0,0){\strut{}}}%
      \csname LTb\endcsname
      \put(2948,5620){\makebox(0,0){\strut{}}}%
      \csname LTb\endcsname
      \put(3778,5620){\makebox(0,0){\strut{}}}%
      \csname LTb\endcsname
      \put(4607,5620){\makebox(0,0){\strut{}}}%
    }%
    \gplgaddtomacro\gplfronttext{%
      \csname LTb\endcsname
      \put(58,6077){\rotatebox{-270}{\makebox(0,0){\strut{}$dSWA$ ($deg$)}}}%
    }%
    \gplgaddtomacro\gplbacktext{%
      \csname LTb\endcsname
      \put(388,4903){\makebox(0,0)[r]{\strut{}$0$}}%
      \csname LTb\endcsname
      \put(388,5072){\makebox(0,0)[r]{\strut{}$0.05$}}%
      \csname LTb\endcsname
      \put(388,5240){\makebox(0,0)[r]{\strut{}$0.1$}}%
      \csname LTb\endcsname
      \put(388,5409){\makebox(0,0)[r]{\strut{}$0.15$}}%
      \csname LTb\endcsname
      \put(388,5577){\makebox(0,0)[r]{\strut{}$0.2$}}%
      \csname LTb\endcsname
      \put(460,4783){\makebox(0,0){\strut{}}}%
      \csname LTb\endcsname
      \put(1289,4783){\makebox(0,0){\strut{}}}%
      \csname LTb\endcsname
      \put(2119,4783){\makebox(0,0){\strut{}}}%
      \csname LTb\endcsname
      \put(2948,4783){\makebox(0,0){\strut{}}}%
      \csname LTb\endcsname
      \put(3778,4783){\makebox(0,0){\strut{}}}%
      \csname LTb\endcsname
      \put(4607,4783){\makebox(0,0){\strut{}}}%
    }%
    \gplgaddtomacro\gplfronttext{%
      \csname LTb\endcsname
      \put(-14,5240){\rotatebox{-270}{\makebox(0,0){\strut{}$dA_y$ ($gs$)}}}%
    }%
    \gplgaddtomacro\gplbacktext{%
      \csname LTb\endcsname
      \put(388,4066){\makebox(0,0)[r]{\strut{}$0$}}%
      \csname LTb\endcsname
      \put(388,4178){\makebox(0,0)[r]{\strut{}$10$}}%
      \csname LTb\endcsname
      \put(388,4291){\makebox(0,0)[r]{\strut{}$20$}}%
      \csname LTb\endcsname
      \put(388,4403){\makebox(0,0)[r]{\strut{}$30$}}%
      \csname LTb\endcsname
      \put(388,4515){\makebox(0,0)[r]{\strut{}$40$}}%
      \csname LTb\endcsname
      \put(388,4628){\makebox(0,0)[r]{\strut{}$50$}}%
      \csname LTb\endcsname
      \put(388,4740){\makebox(0,0)[r]{\strut{}$60$}}%
      \csname LTb\endcsname
      \put(460,3946){\makebox(0,0){\strut{}}}%
      \csname LTb\endcsname
      \put(1289,3946){\makebox(0,0){\strut{}}}%
      \csname LTb\endcsname
      \put(2119,3946){\makebox(0,0){\strut{}}}%
      \csname LTb\endcsname
      \put(2948,3946){\makebox(0,0){\strut{}}}%
      \csname LTb\endcsname
      \put(3778,3946){\makebox(0,0){\strut{}}}%
      \csname LTb\endcsname
      \put(4607,3946){\makebox(0,0){\strut{}}}%
    }%
    \gplgaddtomacro\gplfronttext{%
      \csname LTb\endcsname
      \put(130,4403){\rotatebox{-270}{\makebox(0,0){\strut{}$AV_z$ ($deg/s$)}}}%
    }%
    \gplgaddtomacro\gplbacktext{%
      \csname LTb\endcsname
      \put(388,3229){\makebox(0,0)[r]{\strut{}$0$}}%
      \csname LTb\endcsname
      \put(388,3341){\makebox(0,0)[r]{\strut{}$5$}}%
      \csname LTb\endcsname
      \put(388,3454){\makebox(0,0)[r]{\strut{}$10$}}%
      \csname LTb\endcsname
      \put(388,3566){\makebox(0,0)[r]{\strut{}$15$}}%
      \csname LTb\endcsname
      \put(388,3678){\makebox(0,0)[r]{\strut{}$20$}}%
      \csname LTb\endcsname
      \put(388,3791){\makebox(0,0)[r]{\strut{}$25$}}%
      \csname LTb\endcsname
      \put(388,3903){\makebox(0,0)[r]{\strut{}$30$}}%
      \csname LTb\endcsname
      \put(460,3109){\makebox(0,0){\strut{}}}%
      \csname LTb\endcsname
      \put(1289,3109){\makebox(0,0){\strut{}}}%
      \csname LTb\endcsname
      \put(2119,3109){\makebox(0,0){\strut{}}}%
      \csname LTb\endcsname
      \put(2948,3109){\makebox(0,0){\strut{}}}%
      \csname LTb\endcsname
      \put(3778,3109){\makebox(0,0){\strut{}}}%
      \csname LTb\endcsname
      \put(4607,3109){\makebox(0,0){\strut{}}}%
    }%
    \gplgaddtomacro\gplfronttext{%
      \csname LTb\endcsname
      \put(130,3566){\rotatebox{-270}{\makebox(0,0){\strut{}$V_x$ ($km/h$)}}}%
    }%
    \gplgaddtomacro\gplbacktext{%
      \csname LTb\endcsname
      \put(388,2392){\makebox(0,0)[r]{\strut{}$0$}}%
      \csname LTb\endcsname
      \put(388,2527){\makebox(0,0)[r]{\strut{}$2$}}%
      \csname LTb\endcsname
      \put(388,2662){\makebox(0,0)[r]{\strut{}$4$}}%
      \csname LTb\endcsname
      \put(388,2796){\makebox(0,0)[r]{\strut{}$6$}}%
      \csname LTb\endcsname
      \put(388,2931){\makebox(0,0)[r]{\strut{}$8$}}%
      \csname LTb\endcsname
      \put(388,3066){\makebox(0,0)[r]{\strut{}$10$}}%
      \csname LTb\endcsname
      \put(460,2272){\makebox(0,0){\strut{}}}%
      \csname LTb\endcsname
      \put(1289,2272){\makebox(0,0){\strut{}}}%
      \csname LTb\endcsname
      \put(2119,2272){\makebox(0,0){\strut{}}}%
      \csname LTb\endcsname
      \put(2948,2272){\makebox(0,0){\strut{}}}%
      \csname LTb\endcsname
      \put(3778,2272){\makebox(0,0){\strut{}}}%
      \csname LTb\endcsname
      \put(4607,2272){\makebox(0,0){\strut{}}}%
    }%
    \gplgaddtomacro\gplfronttext{%
      \csname LTb\endcsname
      \put(130,2729){\rotatebox{-270}{\makebox(0,0){\strut{}OSI}}}%
    }%
    \gplgaddtomacro\gplbacktext{%
      \csname LTb\endcsname
      \put(388,1555){\makebox(0,0)[r]{\strut{}$0$}}%
      \csname LTb\endcsname
      \put(388,2229){\makebox(0,0)[r]{\strut{}$1$}}%
      \csname LTb\endcsname
      \put(460,1435){\makebox(0,0){\strut{}$0$}}%
      \csname LTb\endcsname
      \put(1289,1435){\makebox(0,0){\strut{}$50$}}%
      \csname LTb\endcsname
      \put(2119,1435){\makebox(0,0){\strut{}$100$}}%
      \csname LTb\endcsname
      \put(2948,1435){\makebox(0,0){\strut{}$150$}}%
      \csname LTb\endcsname
      \put(3778,1435){\makebox(0,0){\strut{}$200$}}%
      \csname LTb\endcsname
      \put(4607,1435){\makebox(0,0){\strut{}$250$}}%
    }%
    \gplgaddtomacro\gplfronttext{%
      \csname LTb\endcsname
      \put(202,1892){\rotatebox{-270}{\makebox(0,0){\strut{}Oversteer}}}%
      \put(2533,1255){\makebox(0,0){\strut{}frame \#}}%
    }%
    \gplbacktext
    \put(0,0){\includegraphics{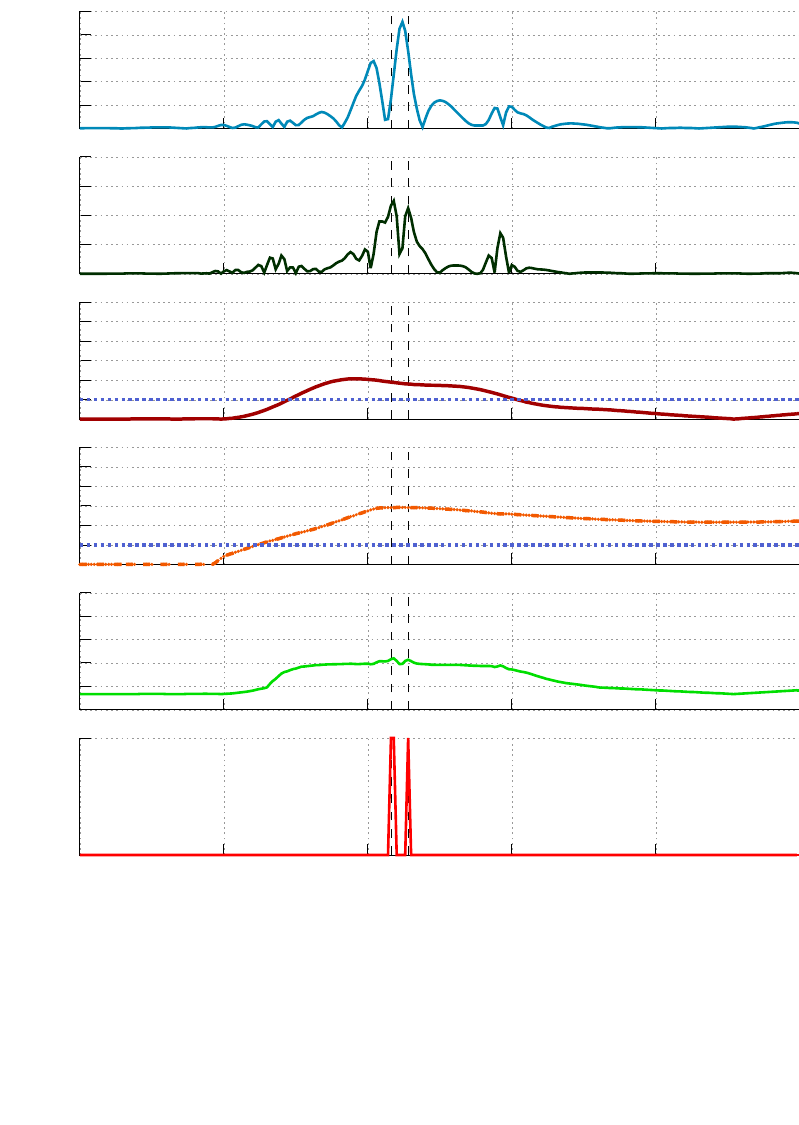}}%
    \gplfronttext
  \end{picture}%
\endgroup